\documentclass[twoside]{article}

\usepackage[preprint]{aistats2027}
\usepackage[round]{natbib}

\usepackage{macros}

\begin{document}

%

%

\twocolumn[

\aistatstitle{Learning Distributions from Samples and Qualities}

\aistatsauthor{ Marc Jourdan \And Nicolas Flammarion }

\aistatsaddress{ EPFL, Lausanne, Switzerland } ]

\begin{abstract}%
  We study parameter estimation when each sample from a continuous parametric distribution is accompanied by a binary quality label governed by the same unknown parameter. 
  By imposing hard feasibility constraints, deterministic labels can accelerate estimation from the classical \(n^{-1/2}\) rate of smooth quality-aware M-estimators to \(n^{-1}\). 
  Combining PAC guarantees with local error-to-parameter conversions, we establish this rate for both identifiable and finitely ambiguous quality classifiers, with likelihood information resolving the latter ambiguity. 
  When the quality model is known only up to an additive constant, we construct threshold-free pairwise preferences from oppositely labeled samples and retain the accelerated rate despite the unknown threshold. 
  We further show that pointwise qualities and pairwise preferences are incomparable: either can be strictly more informative than the other. 
  Our assumptions hold for Gaussian, Laplace, and exponential families under thresholded log-likelihood rewards. 
  Finally, a minimax lower bound establishes the optimal \(n^{-1}\) dependence on sample size, up to logarithmic and dimension-dependent factors, while experiments confirm the benefits of constraint-based estimation.
\end{abstract}%


\section{Introduction}
\label{sec:introduction}
Modern learning systems increasingly rely on two forms of feedback: pointwise quality labels, which evaluate individual observations, and pairwise preferences, which compare their relative quality. 
Such feedback may be provided by humans or automated evaluators and used to filter, reweight, or rank training data~\citep{zhou2023lima,wang2023self,ouyang2022training,rafailov2023direct}. 
In a controlled statistical setting, deterministic preferences have recently been shown to accelerate parameter estimation from the classical \(n^{-1/2}\) rate to \(n^{-1}\)~\citep{jourdan2025learning}. 
We ask whether pointwise quality feedback can yield the same acceleration and whether either form of feedback can be strictly more informative than the other.

We study a continuous parametric model in which the same unknown parameter \(\theta^\star\) governs both the sample distribution \(p_{\theta^\star}\) and a reward function \(r_{\theta^\star}\), whose sign determines sample quality. 
Each i.i.d.\ sample is accompanied by the resulting binary quality label. 
Our canonical example uses the thresholded log-likelihood reward \(r_\theta=\log p_\theta+\eta\), which labels a sample as good whenever its likelihood exceeds a fixed threshold. 
This formulation links quality feedback to the likelihood while capturing mechanisms used in machine learning~\citep{ouyang2022training,zhu2023principled}.

How quality labels are used is crucial. 
Smooth quality-aware M-estimators retain the classical \(n^{-1/2}\) rate~\citep{van2000asymptotic} and may even be inconsistent under deterministic feedback. 
By contrast, enforcing deterministic labels as hard constraints can yield the accelerated \(n^{-1}\) rate. 
We establish this rate by combining uniform error bounds for consistent classifiers~\citep{2946645.3007088} with an error-to-parameter conversion. 
Our analysis also covers finitely ambiguous models, where likelihood information selects the true parameter among the remaining candidates, and unknown thresholds, which we eliminate by constructing pairwise preferences from oppositely labeled samples. 
Pointwise qualities can be strictly more informative: for exponential distributions with thresholded log-likelihood rewards, they yield the accelerated rate even though pairwise log-likelihood differences provide no localization. 
A converse example shows that preferences can also be more informative, so neither feedback model uniformly dominates the other.
\paragraph{Gaussian Example.}
For a univariate Gaussian location family with thresholded log-likelihood quality, good samples constrain the unknown mean to intervals whose intersection is determined by observations closest to the quality boundaries. 
Because these observations approach the boundaries at rate \(1/n\), the consistent parameter set contracts at rate \(1/n\), compared with \(n^{-1/2}\) for the sample mean. 
The acceleration remains attainable without knowing the threshold by pairing oppositely labeled samples to form threshold-free preference constraints. 
Our analysis probes the generality of this phenomenon by identifying sufficient conditions under which it extends beyond this simple example.
Appendix~\ref{app:discussions} discusses practical connections and extensions beyond well-specified deterministic feedback.
\subsection{Contributions}
\label{ssec:contributions}
We show that enforcing deterministic quality labels as hard constraints can accelerate parametric estimation from the \(n^{-1/2}\) rate attained by standard smooth quality-aware M-estimators to nearly \(n^{-1}\).
\begin{enumerate}
    \item 
    We introduce estimators that enforce consistency with the observed quality labels.
    Combining PAC guarantees with local error-to-parameter conversions yields high-probability \(\widetilde{\mathcal O}(n^{-1})\) rates under finite-VC and geometric conditions on \((p_\theta,r_\theta)\).
    Our framework also accommodates finite ambiguity, which we resolve through constrained negative log-likelihood minimization.
    For thresholded log-likelihood rewards, we verify the required conditions for Gaussian, Laplace, and exponential families. 
    Thus, pointwise qualities can be strictly more informative than pairwise preferences.
    \item 
    When the quality threshold is unknown, pairing oppositely labeled samples yields threshold-free pairwise preferences.
    Despite the unknown additive threshold and the strong dependence among the resulting preference constraints, we retain fast rates, verifying the required conditions for Gaussian and Laplace families.
    \item 
    We establish a minimax \(\Omega(n^{-1})\) lower bound, showing that our upper bounds are optimal in their dependence on the sample size, up to logarithmic and dimension-dependent factors.
    Experiments support the predicted separation between smooth quality-aware M-estimators and constraint-based estimators.
\end{enumerate}
\subsection{Related Work}
\label{ssec:relatedWork}
\paragraph{Learning Parametric Distributions.}
Parametric estimation classically proceeds by maximizing the likelihood of the observed samples~\citep{lehmann2006theory,wasserman2013all}.
Under regularity conditions, the MLE is asymptotically normal and achieves the \(n^{-1/2}\) rate characterized by Fisher information and the Cramér--Rao bound~\citep{cramer1946mathematical,van2000asymptotic,ibragimov2013statistical}. 
Finite-sample upper bounds and information-theoretic lower bounds further characterize the difficulty of such problems~\citep{birge1993rates,spokoiny2012parametric,lecam1973convergence,fano1952class,assouad1983deux}.
We depart from the sample-only setting by observing a parameter-dependent quality labels.
The discontinuity introduced by deterministic labels permits faster rate.
\paragraph{Level-set Estimation and Binary Quality Feedback.}
A deterministic quality label specifies membership in the parameter-dependent super-level set $\{x:r_{\theta^\star}(x)\geq 0\}$.
This connects our model to density level-set and excess-mass estimation, as well as prediction-set construction~\citep{hartigan1987estimation,polonik1995measuring,tsybakov2009nonparametric,cadre2006kernel,rigollet2009optimal,lei2013distribution}.
Classical work seeks to recover the level set itself from unlabeled samples, with rates controlled by smoothness, margin, and geometric assumptions.
Our objective is different: the level-set memberships are observed, and we use them as side information for estimating the finite-dimensional parameter that determines both the distribution and the level set.
\paragraph{Preference-based Estimation.}
The closest work is \citet{jourdan2025learning}, which estimates a parametric distribution from samples and reward-based pairwise preferences.
We instead observe pointwise quality labels.
When the quality threshold is unknown, these labels can be converted into preferences by pairing good and bad samples, allowing threshold-invariant reward differences to be used.
However, our comparisons are constructed from a common set of observations and are therefore strongly dependent, whereas the comparisons in \citet{jourdan2025learning} are directly observed from independent sample pairs.
Moreover, our separation examples show that pointwise quality and pairwise preference feedback contain genuinely different information.
Our results complement the broader literature on Bradley--Terry models, ranking, reward learning, and preference-based optimization~\citep{bradley1952rank,hajek2014minimax,shah2016estimation,shah2018simple,zhu2023principled,ge2024learning,yao2025leveraging}.
\paragraph{PAC Learning and Version Spaces.}
Our analysis draws on realizable PAC learning~\citep{10.1145/1968.1972} and the characterization of distribution-free learnability through VC dimension~\citep{vapnik1971uniform,10.1145/76359.76371}. 
For a class of VC dimension \(d\), classical results bound the error uniformly over all classifiers in the version space by \(\mathcal O((d\log(n/d)+\log(1/\delta))/n)\). 
This differs by a logarithmic factor from the distribution-free minimax lower bound \(\Omega((d+\log(1/\delta))/n)\)~\citep{EHRENFEUCHT1989247}, motivating refined analyses of consistent learners. 
Sample-compression schemes instead control generalization through the number of observations needed to reconstruct a consistent classifier~\citep{floyd1995sample}. 
For consistent monotone compression rules,~\citet{2946645.3007088} remove the additional logarithmic factor and obtain the optimal \(1/n\) dependence. 
In our setting, classification is an intermediate device: geometric conditions convert these uniform classification-error bounds into parameter-localization guarantees.


\section{Problem statement}
\label{sec:problem_statement}
\paragraph{Parameter Estimation.}
Let $\Theta \subseteq \R^k$ be a set of parameters for a class of continuous probability distributions $\cF$ over $\cX \subseteq \R^d$.
Let $\theta^\star \in \Theta$ be an unknown parameter to estimate.
Our samples are drawn i.i.d. from $p_{\theta^\star}$.
As illustrative examples, we use three classes of distributions: $\cF_{\cN, \Sigma}$ for multivariate \emph{Gaussian} with known covariance $\Sigma$ and natural parameters $\Theta$; $\cF_{\lap, b}$ for \emph{Laplace} with known scale $b$ and natural parameters $\Theta$; $\cF_{\Exp}$ for \emph{exponential} and natural parameters $\Theta$.
\paragraph{Quality Feedback.}
To model sample quality, we use a parametric reward model $r_{\theta}:\cX\to\R$.
A sample is \emph{good} if its reward is positive and \emph{bad} if its reward is negative; equivalently, its quality label is $\sign(r_{\theta}(x))\in\{\pm1,0\}$.
As a concrete example used in our derivations, we take the reward to be a \emph{thresholded log-probability}, $r_{\theta}(x)\eqdef \log p_{\theta}(x)+\eta$, where $\eta\in\R$ is a threshold.
In this case, good samples are those whose model likelihood exceeds \(e^{-\eta}\), making the reward a natural measure of whether a sample lies in a typical region of the model.

We model the quality feedback as \emph{deterministic} labels, and refer to Section~\ref{ssec:QbasedMestimator} for a treatment of noisy feedback with \emph{stochastic} labels.
Let $q_{\theta}$ denote the joint density function of $(X,Z)$, i.e., for all $(x,z) \in \cX\times\{\pm1,0\}$,
\[
    q_{\theta}(x,z)\eqdef p_{\theta}(x) \indi{z=\sign(r_{\theta}(x))} \: .
\]
In our examples, $p_{\theta^\star}(r_{\theta^\star}(X) = 0) = 0$, so zero labels occur with probability zero; otherwise they are treated as uninformative boundary observations.
Our observations are drawn i.i.d. from $q_{\theta^\star}$, namely $(X_i, Z_i)_{i \in [n]} \sim q_{\theta^\star}^{\otimes n}$.

In practice, such feedback may be provided by a human or automated annotator indicating whether each sample meets a prescribed quality criterion, with deterministic labels corresponding to reproducible assessments and stochastic labels capturing variability or noise in this evaluation.
\paragraph{Goal: Estimators with a Fast Estimation Rate.}
Given sample--quality pairs, we seek estimators $\wh\theta_n$ achieving a \emph{fast estimation rate}, namely $\|\wh\theta_n-\theta^\star\|=\wt\cO(1/n)$ with high probability for all sufficiently large $n$.
Here, $\wt\cO(\cdot)$ suppresses factors in $\log(n)$ and in the dimension $k$.
\paragraph{Outline.}
Sections~\ref{sub:sample_only_mle} and~\ref{ssec:QbasedMestimator} show that sample-only and stochastic-quality M-estimators retain the classical \(n^{-1/2}\) rate, while the same smooth objective may be inconsistent under deterministic feedback.
Sections~\ref{sec:estimator4qualities} and~\ref{sec:reduction_to_preferences} instead exploit hard constraints to attain \(\widetilde{\mathcal O}(n^{-1})\) rates when the reward is known exactly or up to an additive constant.
Section~\ref{sec:lower_bound} establishes a matching \(\Omega(n^{-1})\) minimax dependence on \(n\), and Section~\ref{sec:experiments} provides empirical validation.
\subsection{Sample-only MLE}
\label{sub:sample_only_mle}
In the absence of quality observations, a natural baseline is to estimate $\theta^\star$ directly from the observations. 
The sample-only MLE (\ref{eq:somle}) is
\begin{align} \label{eq:somle}
    &\wh \theta^{\so}_{n} \in  \argmin \{ L^{\so}_{n}(\theta) \mid \theta \in \Theta \}  \tag{SO MLE} \\
    &\text{with} \quad L^{\so}_{n}(\theta) \eqdef -\sum\nolimits_{i \in [n]} \log p_{\theta}(X_i) \: . \nonumber 
\end{align}
\paragraph{Asymptotic Behavior of $\so$.}
Under standard regularity conditions~\citep{van2000asymptotic},~\ref{eq:somle} is asymptotically normal: 
$\sqrt{n}(\wh\theta^{\so}_{n}-\theta^\star) \rightsquigarrow_{n\to+\infty} \cN(0_k,\cI(p_{\theta^\star})^{-1})$, where $\cI(p_{\theta})\eqdef \bE_{p_{\theta}}[\nabla_{\theta}\log p_{\theta}(\nabla_{\theta}\log p_{\theta})\transpose]$ is the Fisher information matrix of $p_{\theta}$ and $\rightsquigarrow$ denotes convergence in distribution.
By the Cramér--Rao bound~\citep{rao1992information},~\ref{eq:somle} has optimal asymptotic covariance among unbiased sample-only estimators: any such estimator with asymptotic covariance $V$ satisfies $V\succeq \cI(p_{\theta^\star})^{-1}$, where $\succeq$ denotes the Loewner order on positive semidefinite matrices.

By ``standard regularity,'' we mean the usual identifiability and local regularity conditions ensuring that the population negative log-likelihood is uniquely minimized at $\theta^\star$, and that the MLE is consistent and asymptotically normal; see, e.g., \citet[Ch.~5]{van2000asymptotic}. 
These include, in particular, continuity of $\theta\mapsto\KL(p_{\theta^\star},p_\theta)$.
Classical smooth-likelihood theory applies to $\cF_{\cN,\Sigma}$ and $\cF_{\Exp}$. 
For $\cF_{\lap,b}$, whose MLE is a sample median, the same conclusions follow from nonsmooth M-estimation theory.
\subsection{Quality-based M-estimators}
\label{ssec:QbasedMestimator}
To model noisy quality feedback, we use the sigmoid link \(\sigma(x)\eqdef(1+e^{-x})^{-1}\) and define
\[
    \forall (x,z) \in \cX\times\{\pm1\}, \quad q_{\theta}^{\sto}(x,z)\eqdef p_{\theta}(x)\sigma(z r_{\theta}(x))\:.
\]
The corresponding negative log-likelihood augments \(L_n^{\so}\) with a logistic classification loss, yielding the stochastic quality MLE (\ref{eq:sqmle}) 
\begin{align} \label{eq:sqmle}
    &\wh \theta^{\textrm{SQ}}_{n} \in \argmin \{ L^{\textrm{SQ}}_{n}(\theta)\mid \theta \in \Theta \} \: \text{ with } \tag{SQ MLE}\\
    &L^{\textrm{SQ}}_{n} (\theta) \eqdef  L^{\so}_{n}(\theta) -  \sum\nolimits_{i \in [n]}  \log \sigma (Z_i r_{\theta}(X_i)) \: . \nonumber
\end{align}
We distinguish its behavior under stochastic and deterministic quality labels by writing the resulting estimators as
\(\sqe\) and \(\lle\), respectively.
\paragraph{Stochastic Qualities.}
Under standard regularity conditions~\citep{van2000asymptotic}, \(\sqe\) is asymptotically normal:  $\sqrt{n}(\wh \theta^{\sqe}_{n}  - \theta^\star ) \rightsquigarrow_{n\to+ \infty}  \cN(0_{k}, \cI(q_{\theta^\star}^{\sto})^{-1})$.
This covariance is optimal among unbiased estimators using stochastic qualities with the sigmoid link~\citep{rao1992information}.
Lemma~\ref{lem:sp_mle_asymptotic_variance} (Appendix~\ref{app:mestimator}) gives conditions under which \(\sqe\) improves asymptotically over the sample-only MLE.
\paragraph{Deterministic Qualities.}
Applying the same smooth objective to deterministic labels produces a misspecified M-estimator.
Its population objective is
\[
    M_{\det}(\theta) \eqdef \bE_{X\sim p_{\theta^\star}} \left[ \log q_{\theta}^{\sto} \bigl(X,\sign(r_{\theta^\star}(X))\bigr) \right] \: ,
\]
so, under standard regularity conditions, \(\wh\theta_n^{\lle}\) converges to a maximizer of \(M_{\det}\), which need not be \(\theta^\star\).
Thus, \(\lle\) may be inconsistent.
Even when it is consistent, it generally remains asymptotically normal and need not improve upon~\ref{eq:somle}.
Appendix~\ref{app:mestimator} gives Gaussian, Laplace, and exponential examples illustrating both phenomena.
\paragraph{Deterministic versus Stochastic Feedback.}
Under the smooth sigmoid model, regular estimators are asymptotically normal and retain the classical \(\Theta(n^{-1/2})\) rate.
By contrast, deterministic labels create a discontinuity at the quality boundary, which hard consistency constraints exploit to achieve faster estimation rates.


\section{Estimators for deterministic qualities}
\label{sec:estimator4qualities}
When the reward model is known exactly, we establish accelerated \(\widetilde{\cO}(1/n)\) rates for a broad class of parametric distributions. 
Pointwise quality labels can be strictly more informative than pairwise preferences: the rate acceleration persists even when preference feedback provides no localization.
\paragraph{Quality-based Estimators.}
We define $\cC_n$ as the set of parameters that perfectly classify all training examples (namely, minimize the empirical $0$--$1$ loss), i.e.,
\begin{align} \label{eq:qualityzero_one_loss}
    \cC_{n} &\eqdef \argmin_{\theta \in \Theta}  \sum\nolimits_{i \in [n]} \indi{ Z_i r_{\theta}(X_i) < 0} \\
    &= \left\{ \theta \in \Theta \mid \: \forall i \in [n], \: Z_{i} r_{\theta}(X_i) \ge 0 \right\} \: , \nonumber
\end{align}
which is non-empty as $\theta^\star \in \cC_{n}$.
Let $\cC_n^{+}$ denote the positive-only ($\po$) analogue, defined using only constraints from good samples $\cI_n^{+}\eqdef\{i\in[n]\mid Z_i=1\}$.

Any member of these feasibility sets can be used as an estimator of $\theta^\star$.
However, this discards some of the likelihood information contained in the samples themselves.
Therefore, we minimize the negative log-likelihood of the samples within these feasibility sets.
Specifically, we define the deterministic quality MLE (\ref{eq:dqle}) as
\begin{equation}  \label{eq:dqle} 
    \wh \theta^{\dpe}_{n} \in \argmin \{ L^{\so}_{n}(\theta) \mid \theta \in \cC_{n} \}  \: .\tag{DQ MLE}
\end{equation}
\paragraph{Positive-only Identifiability.}
Let $\cD^{+}(\theta^\star,\theta)\eqdef\{x\in\cX\mid r_{\theta^\star}(x)>0, r_\theta(x)<0\}$ be the set of truly good observations misclassified as bad by \(\theta\), and $p_{\theta^\star}(\cD^{+}(\theta^\star,\theta))$ be its associated disagreement probability.
Let $\cC^{+}(\theta^\star) \eqdef \{ \theta \in \Theta \mid p_{\theta^\star}(\cD^{+}(\theta^\star,\theta)) = 0 \} $ be the parameters whose induced classifier labels as good \(p_{\theta^\star}\)-almost every observation that is good under \(\theta^\star\).
A model is \emph{positive-only identifiable} at \(\theta^\star\) if \(\cC^{+}(\theta^\star)=\{\theta^\star\}\).
Deterministic positive-only quality labels alone distinguish \(\theta^\star\) from the other elements of \(\Theta\).

For thresholded log-likelihood rewards with $\cF_{\cN, \Sigma}$ or $\cF_{\lap, b}$, the model is positive-only identifiable.
Theorem~\ref{thm:specialized_classifier_to_parameter_identifiability} gives a \(\widetilde{\cO}(1/n)\) bound on $\sup_{\theta \in \cC_{n}^{+}}\|\theta - \theta^\star\|$ with high probability; thus fast convergence rates for any $\wh \theta_n\in \cC_{n}^{+}$.
\begin{theorem} \label{thm:specialized_classifier_to_parameter_identifiability}
    Let $\delta \in (0,1)$, $\eta_0 = \frac{1}{2}\log((2\pi)^d|\Sigma|)$ for $\cF_{\cN, \Sigma}$ and $\eta_0 = \log(2b)$ for $\cF_{\lap, b}$ with $d=1$.
    For thresholded log-likelihood rewards with \emph{known} threshold $\eta \in (\eta_0,+ \infty) $, there exist $\epsilon_{\cF,\eta}, \alpha_{\cF,\eta} >0$ s.t., for all $\theta^\star \in \R^d $ and all $n \ge  \cO\left(\frac{\log(1/\delta)}{\epsilon_{\cF,\eta} \alpha_{\cF,\eta}}\right)$, with probability $1-\delta$,
    \begin{align*}
        &\cF_{\lap,b} \text{ or } \cF_{\cN,\sigma^2} \: \text{:}\quad \sup_{\theta \in \cC_{n}^{+}}|\theta - \theta^\star| \le \frac{42 + 16 \ln \left(\frac{3}{\delta}\right)}{n \alpha_{\cF,\eta}} \: , \\
        &\cF_{\cN,\Sigma} \:\text{:}\quad  \sup_{\theta \in \cC_{n}^{+}}\|\theta - \theta^\star\|_{\Sigma} \le \frac{\cO\left( d \ln \left( \frac{n}{d} \right) + \ln \left(\frac{1}{\delta}\right)\right)}{n \alpha_{\cF,\eta}} \: ,
    \end{align*}
    where $\|x\|_{\Sigma} \eqdef \sqrt{x\transpose \Sigma x}$.
\end{theorem}
\paragraph{Computational Benefits of Positive-only Constraints.}
Although \(\mathcal C_n^+\) discards constraints from bad samples, Theorem~\ref{thm:specialized_classifier_to_parameter_identifiability} shows that any estimator in \(\mathcal C_n^+\) achieves the fast rate.
Moreover, if \(\theta\mapsto r_\theta(x)\) is quasi-concave, then \(\mathcal C_n^+\) is an intersection of convex sets---ellipsoids for Gaussian location families and intervals for the univariate Laplace family.
Thus, our guarantees can be attained through convex feasibility, or convex likelihood minimization, without solving the generally non-convex full-feedback problem.
See Appendix~\ref{app:ssec_implementation_details} for details.

\paragraph{Proof Intuition and Structural Conditions.}
The accelerated rate rests on three key features.
First, deterministic feedback creates a hard boundary: moving the parameter across this boundary changes the observed label discontinuously.
Second, every identifiable parameter direction must move the induced classifier.
Third, a displacement of size \(t\) must sweep out a region of probability mass of order \(t\).
A realizable PAC bound first controls the disagreement probability of every feasible classifier at rate \(\widetilde{\mathcal O}(1/n)\); the linear boundary-mass condition then converts this directly into \(\widetilde{\mathcal O}(1/n)\) parameter localization.
More generally, if the disagreement probability grows as \(\|\theta-\theta^\star\|^\kappa\), the same argument would yield a rate \(\widetilde{\cO}(n^{-1/\kappa})\).
Appendix~\ref{app:discussions} discusses how these conditions can be verified and weakened.

For \(\cF_{\lap,b}\) and \(\cF_{\cN,\sigma^2}\), monotone compression rules of size at most \(2\) remove the logarithmic
dependence on $n$~\citep[Theorem 3]{2946645.3007088}.
Section~\ref{sub:classifier_to_parameter_identifiability} formalizes these two ingredients for general reward and distribution models (Theorem~\ref{thm:classifier_to_parameter_identifiability}).
\paragraph{Finite Ambiguity.}
Let $\cD(\theta^\star,\theta)\eqdef\{x\in\cX\mid r_{\theta^\star}(x)r_\theta(x)<0\}$ be the set of observations on which $\theta$ and $\theta^\star$ induce opposite classifiers, and $p_{\theta^\star}(\cD(\theta^\star,\theta))$ be its associated disagreement probability.
Let $\cC(\theta^\star) \eqdef \{ \theta \in \Theta \mid p_{\theta^\star}(\cD(\theta^\star,\theta)) = 0 \}$ be the set of parameters whose induced classifier agrees with that of \(\theta^\star\) \(p_{\theta^\star}\)-almost surely.
A model is finitely ambiguous at \(\theta^\star\) if \(\cC(\theta^\star)\setminus\{\theta^\star\}\) is finite and nonempty.
Quality labels alone cannot distinguish these observationally equivalent parameters, so we use the sample likelihood to resolve the remaining ambiguity.

For thresholded log-likelihood rewards with $\cF_{\Exp}$, the model is ambiguous, with $|\cC(\theta^\star)|=2$.
Despite this ambiguity, Theorem~\ref{thm:specialized_classifier_to_parameter_ambiguity} shows a fast convergence rate for $\wh \theta^{\dpe}_{n}$.
\begin{theorem} \label{thm:specialized_classifier_to_parameter_ambiguity}
    Let $\delta \in (0,1)$. 
    For thresholded log-likelihood rewards with \emph{known} threshold $\eta \in \R$ and $\cF_{\Exp}$ with $\theta^\star \in (e^{-\eta},+\infty)\setminus \{e^{1-\eta}\}$, there exist $\epsilon_{\star,\eta}, \alpha_{\star,\eta} >0$ such that, for all $n \ge \cO\left(\log(1/\delta)\right)$, with probability $1-\delta$, 
    \begin{align*}
        &|\wh \theta^{\dpe}_{n}  - \theta^\star| \le \frac{42 + 16 \ln \left(6 / \delta\right)}{n \alpha_{\star,\eta} } \: .
    \end{align*}
\end{theorem}
The quality constraints first localize \(\cC_n\) around the finite set \(\cC(\theta^\star)\) at rate \(\cO(1/n)\).
Although the sample-only MLE converges at the slower \(\cO(n^{-1/2})\) rate, the sample likelihood identifies the component containing \(\theta^\star\), preserving the fast localization rate.
Section~\ref{sub:classifier_to_parameter_ambiguity} extends this argument to finitely ambiguous models under general geometric and likelihood-concentration conditions (Theorem~\ref{thm:classifier_to_parameter_ambiguity}).

Theorem~\ref{thm:specialized_classifier_to_parameter_ambiguity} excludes two regimes.
Parameters in \((0,e^{-\eta}]\) all induce the same all-negative classifier, while at \(\theta^\star=e^{1-\eta}\) the good region is locally insensitive to parameter perturbations.
A uniform-in-$\theta^\star$ version of Theorem~\ref{thm:specialized_classifier_to_parameter_ambiguity} therefore requires staying away from these regimes.
\paragraph{Pointwise Qualities versus Pairwise Preferences.}
For Gaussian and Laplace location families, Theorem~\ref{thm:specialized_classifier_to_parameter_identifiability}
matches the fast rate obtained from pairwise preferences by~\citet{jourdan2025learning}.
For the exponential family, however, Theorem~\ref{thm:specialized_classifier_to_parameter_ambiguity} shows that pointwise qualities yield fast estimation even though the corresponding preferences provide no localization.
Therefore, pointwise qualities can be more informative than pairwise preferences.

Conversely, pairwise preferences can be strictly more informative than pointwise qualities. 
Let \(\eta_0\) denote the critical threshold introduced in Theorem~\ref{thm:specialized_classifier_to_parameter_identifiability} for thresholded log-likelihood rewards under \(\cF_{\cN,\Sigma}\) and \(\cF_{\lap,b}\). 
When \(\eta<\eta_0\), all pointwise labels are negative and provide no information beyond the samples.
Pairwise preferences nevertheless remain informative because the additive threshold cancels in reward differences, allowing fast estimation rates~\citep{jourdan2025learning}.

A similar separation can arise when all pointwise labels are positive. 
Consider a Gaussian location family truncated to \([-1,1]\), with \(\theta\in[-1/2,1/2]\), and choose the threshold large enough so that all pointwise quality labels are positive. 
These labels are again uninformative, whereas each pairwise preference reveals on which side of a random midpoint \(\theta^\star\) lies. 
Since these midpoints have positive density around \(\theta^\star\), pairwise preferences yield an \(\mathcal O(1/n)\)
localization rate. 
In Appendix~\ref{sub:pointwise_qualities_versus_pairwise_preferences}, we provide a less degenerate separation in which both pointwise labels occur with positive probability but remain uninformative.
\paragraph{Unknown Threshold.}
Suppose that the true reward is \(r_{\theta^\star}=\log p_{\theta^\star}+\eta^\star\), while the constraints are constructed using a candidate threshold \(\eta\).
When \(\eta\neq\eta^\star\), the full constraints eventually exclude \(\theta^\star\), whereas the positive-only constraints become either too restrictive or too permissive, depending on the direction of misspecification.
Section~\ref{sec:reduction_to_preferences} circumvents this issue by constructing threshold-free pairwise preferences from oppositely labeled samples.
\subsection{Positive-only Identifiability}
\label{sub:classifier_to_parameter_identifiability}
We propose suitable geometric assumptions on \(r_\theta\) and \(p_\theta\) to obtain fast convergence rates for any $\wh \theta_n\in \cC_{n}^{+}$.
\paragraph{Uniform PAC Bound.}
The VC~\citep{vapnik1971uniform} dimension of a set $\cA$ of subsets of $\cX$ is 
\[
    \vc(\cA) \eqdef \sup\left\{ |S| \mid  S \subseteq \cX,\; \{S \cap A : A \in \cA\}=2^S \right\} \: .
\]
The VC dimension of a set $\cH$ of classifiers is defined as $\vc(\cH) \eqdef \vc (\{\{x : h(x)=+1\} : h \in \cH \})$.
Let $\cH^{+}_{\star}\eqdef \{h_{\theta}^{+} \mid \theta \in \Theta\}$ denote the class of classifiers restricted to the positive region induced by $\theta^\star$, namely
\[
    \forall x \in \cX, \: h_{\theta}^{+}(x) \eqdef \begin{cases}
    -1 & \text{if } r_{\theta^\star}(x) > 0 \text{ and } r_\theta(x) < 0\\
    +1 &\text{otherwise}
    \end{cases} \: .
\]
The setting is realizable as $h_{\theta^\star}^{+}\in \cH^{+}_{\star}$, and $h_{\theta^\star}^{+} \equiv 1$.
\begin{assumption} \label{ass:finite_VC_identifiability}
    $d^{+}_{\star} \eqdef \mathrm{vc}(\cH^{+}_{\star})$ is finite.
\end{assumption}
In the terminology of realizable PAC learning, $\cC_n^{+}$ is the version space of parameters consistent with its observed labels, while $p_{\theta^\star}(\cD^{+}(\theta^\star,\theta))$ is the classification error of $h_{\theta}^{+}$ relative to \(h_{\theta^\star}^{+}\).
Namely, we have
\begin{align*}
    &\cC_n^{+} = \{\theta \mid \forall i \in [n], \: h_{\theta}^{+}(X_i) = h_{\theta^\star}^{+}(X_i) \} \: , \\
    &p_{\theta^\star}(\cD^{+}(\theta^\star,\theta)) = p_{\theta^\star}( \{x \mid h_{\theta}^{+}(x) \ne h_{\theta^\star}^{+}(x)\} ) \: .
\end{align*}
For consistent learning rules, the uniform control on the classification error has been extensively studied in the realizable case.
Under Assumption~\ref{ass:finite_VC_identifiability},~\citet[Theorem 11]{2946645.3007088} yields, with probability at least $1-\delta$,
\[
    \sup_{\theta \in \cC_{n}^{+}} p_{\theta^\star}(\cD^{+}(\theta^\star,\theta)) \le \frac{\cO \left( d^{+}_{\star} \ln \left( n/d^{+}_{\star} \right) + \ln \left(1/\delta\right)\right)}{n} \: .
\]
\paragraph{Local Error-to-parameter Conversion.}
Assumption~\ref{ass:conversion_error_to_parameter_identifiability} turns the control of the disagreement probability into a bound on the parameter error. 
In particular, it implies that the model is positive-only identifiable at $\theta^\star$.
\begin{assumption} \label{ass:conversion_error_to_parameter_identifiability}
    There exist $\alpha^{+}_{\star}, \varepsilon^{+}_{\star} > 0$ such that
    \[
        \forall \theta\in \Theta, \quad p_{\theta^\star}(\cD^+(\theta^\star,\theta)) \geq \alpha^{+}_{\star} \min\left\{ \left\|\theta-\theta^\star\right\|, \varepsilon^{+}_{\star} \right\} \: .
    \]
\end{assumption}
\paragraph{Fast Estimation Rate.}
Under these assumptions, Theorem~\ref{thm:classifier_to_parameter_identifiability} gives a \(\widetilde{\cO}(1/n)\) bound on $\sup_{\theta \in \cC_{n}^{+}}\|\theta - \theta^\star\|$ with high probability; thus fast convergence rates on $\cC_{n}^{+}$.
\begin{theorem} \label{thm:classifier_to_parameter_identifiability}
    Let $\delta \in (0,1)$.
    Let $\theta^\star \in \Theta$ such that Assumptions~\ref{ass:finite_VC_identifiability} and~\ref{ass:conversion_error_to_parameter_identifiability} hold true.
    For all $n \ge  \cO\left(\frac{\log(1/\delta)}{\epsilon^{+}_{\star} \alpha^{+}_{\star}}\right)$, with probability at least $1-\delta$, 
    \begin{align*}
        \sup_{\theta \in \cC_{n}^{+}}\|\theta - \theta^\star\| \le \frac{\cO \left( d^{+}_{\star} \ln \left( n/d^{+}_{\star} \right) + \ln \left(1/\delta\right)\right)}{n \alpha^{+}_{\star}} \: .
    \end{align*}
\end{theorem}
\subsection{Finite Ambiguity}
\label{sub:classifier_to_parameter_ambiguity}
We now allow the quality classifier to identify only a finite set of observationally equivalent parameters. 
Unlike in Section~\ref{sub:classifier_to_parameter_identifiability}, the quality constraints may therefore localize \(\theta^\star\) only up to \(\cC(\theta^\star)\). 
The sample likelihood will resolve this remaining ambiguity.
Let $\cH_{\star}\eqdef \{h_{\theta} \mid \theta \in \Theta\}$ be the class of classifiers induced by $\theta^\star$, namely $h_{\theta}(x) = -1$ if $r_{\theta^\star}(x) r_\theta(x) < 0$, and $h_{\theta}(x) = +1$ otherwise.

Then, \(\cC_n\) is the corresponding version space and the classification error associated with \(\theta\) is
\(p_{\theta^\star}(\cD(\theta^\star,\theta))\).
\begin{assumption} \label{ass:finite_VC_ambiguity}
    $d _{\star} \eqdef \mathrm{vc}(\cH _{\star})$ is finite.
\end{assumption}
\begin{assumption} \label{ass:conversion_error_to_parameter_ambiguity}
    $\cC(\theta^\star)$ is finite. There exist $\alpha_{\star}, \varepsilon_{\star} > 0$ s.t.
    \[
        \forall \theta\in \Theta, \quad p_{\theta^\star}(\cD(\theta^\star,\theta)) \geq \alpha_{\star} \min\left\{ \mathrm{dist}(\theta,\cC(\theta^\star)), \varepsilon_{\star} \right\} \: ,
    \]
    where $\mathrm{dist}(\theta,\cC(\theta^\star)) \eqdef \inf_{\lambda \in \cC(\theta^\star)}\|\theta - \lambda\|$.
\end{assumption}
Assumptions~\ref{ass:finite_VC_ambiguity} and~\ref{ass:conversion_error_to_parameter_ambiguity}, together with the same PAC argument as in Section~\ref{sub:classifier_to_parameter_identifiability}, localize \(\cC_n\) within \(\widetilde{\cO}(1/n)\) of \(\cC(\theta^\star)\).
To select the component containing \(\theta^\star\), we impose the following standard uniform concentration condition on the sample negative log-likelihood (Assumption~\ref{ass:uniform_bernstein_bound}).
\begin{assumption} \label{ass:uniform_bernstein_bound}
    For every compact $K\subset\Theta$, there exist $C_K,D_K\ge 0$ such that, for all $(n,\delta)\in\N \times (0,1)$, with probability at least $1-\delta$,
    \begin{align*}
        & \sup_{\theta\in K} \left| (L_n^{\so}(\theta)-L_n^{\so}(\theta^\star))/n - \KL(p_{\theta^\star},p_\theta)\right|\\
        & \quad \le C_K\sqrt{\frac{\log(2/\delta)}{n}} + D_K\frac{\log(2/\delta)}{n} \:.
    \end{align*}
\end{assumption}
Because \(\cC(\theta^\star)\) is finite and \(\KL(p_{\theta^\star},p_\theta)>0\) for \(\theta\ne\theta^\star\), Assumption~\ref{ass:uniform_bernstein_bound} ensures that the constrained likelihood selects the component containing \(\theta^\star\). 
Therefore, $\wh \theta^{\dpe}_{n}$ achieves the same fast rate as in the identifiable case (Theorem~\ref{thm:classifier_to_parameter_ambiguity}).
\begin{theorem} \label{thm:classifier_to_parameter_ambiguity}
    Let $\delta \in (0,1)$.
    Let $\theta^\star \in \Theta$ such that Assumptions~\ref{ass:finite_VC_ambiguity},~\ref{ass:conversion_error_to_parameter_ambiguity} and~\ref{ass:uniform_bernstein_bound} hold true.
    For all $n \ge  \cO\left(\log(1/\delta)\right)$, with probability at least $1-\delta$, 
    \begin{align*}
        \|\wh \theta^{\dpe}_{n} - \theta^\star\| \le \frac{\cO \left( d_{\star} \ln \left( n/d_{\star} \right) + \ln \left(1/\delta\right)\right)}{n \alpha_{\star}} \: .
    \end{align*}
\end{theorem}


\section{Threshold-free estimators}
\label{sec:reduction_to_preferences}
When the reward model is known only up to an additive constant, we construct pairwise reward-based preferences from deterministic binary quality labels.
Despite this misspecification, we recover an accelerated \(\widetilde{\cO}(1/n)\) rate.
This reduction leads to the preference-based estimators of~\citet{jourdan2025learning}, but requires a new analysis because the threshold is unknown and the preferences are correlated.
\paragraph{Preference-based Estimators.}
Define the reward-based preference model $\ell_{\theta}(x,y) \eqdef r_{\theta}(x)-r_{\theta}(y)$.
A sample \(x\) is preferred to \(y\) when \(\ell_{\theta}(x,y)>0\).
Because additive constants cancel in pairwise differences, \(\ell_\theta\) remains known even when \(r_\theta\) is known only up to such a constant.

We construct preferences by pairing every good sample with every bad one.
Let \(\cI_n^-\eqdef\{i\in[n]\mid Z_i=-1\}\).
Then, $\ell_{\theta^\star}(X_i,X_j)>0$ for all $(i,j)\in\cI_n^+\times\cI_n^-$.
This motivates the feasible set
\begin{align} \label{eq:prefzero_one_loss}
    &\cC_{n}^{\all} \eqdef \left\{ \theta \mid \forall (i,j)\in \cI_{n}^{+} \times \cI_{n}^{-} , \:  \ell_{\theta}(X_i,X_j) \ge 0 \right\} \: .
\end{align} 
It is nonempty because \(\theta^\star\in\cC_n^{\all}\), and requires knowledge of \(r_\theta\) only up to an additive constant.
\paragraph{Threshold Misspecification.}
For ease of exposition, consider thresholded log-likelihood rewards with an unknown threshold.
Eliminating the threshold has different effects across distribution families.
For \(\cF_{\Exp}\), the resulting preferences provide no localization, since \(\cC_n^{\all}=\Theta\).
By contrast, for \(\cF_{\cN,\Sigma}\) and \(\cF_{\lap,b}\), Theorem~\ref{thm:specialized_threshold_misspecification} shows that \(\cC_n^{\all}\) contracts around \(\theta^\star\) at rate \(\widetilde{\cO}(1/n)\).
Thus, any estimator \(\widehat\theta_n\in\cC_n^{\all}\) achieves the fast rate despite the unknown threshold.
\begin{theorem} \label{thm:specialized_threshold_misspecification}
    Let $\delta \in (0,1)$ and $\eta_0$ as in Theorem~\ref{thm:specialized_classifier_to_parameter_identifiability}.
    For thresholded log-likelihood rewards with \emph{unknown} threshold $\eta \in (\eta_0,+ \infty) $, there exist $\epsilon_{\cF,\eta}^{\all}, \alpha_{\cF,\eta}^{\all} >0$ s.t., for all $\theta^\star \in \R^d $ and all $n \ge  \cO\left(\frac{\log(1/\delta)}{\epsilon_{\cF,\eta}^{\all} \alpha_{\cF,\eta}^{\all}}\right)$, with probability $1-\delta$,
    \begin{align*}
        &\cF_{\lap,b} \text{ or } \cF_{\cN,\sigma^2} \: \text{:}\quad \sup_{\theta \in \cC_{n}^{\all}}|\theta - \theta^\star| \le \frac{84 + 16 \ln \left(\frac{3}{\delta}\right)}{n \alpha_{\cF,\eta}^{\all}} \: , \\
        &\cF_{\cN,\Sigma} \:\text{:}\quad  \sup_{\theta \in \cC_{n}^{\all}}\|\theta - \theta^\star\|_{\Sigma} \le \frac{\cO\left( d \ln \left( \frac{n}{d} \right) + \ln \left(\frac{1}{\delta}\right)\right)}{n \alpha_{\cF,\eta}^{\all}} \: .
    \end{align*}
\end{theorem}
The proof follows the PAC and error-to-parameter strategy of Theorem~\ref{thm:specialized_classifier_to_parameter_identifiability}, but controls the minimum classification error over the unknown
threshold. 
For \(\cF_{\lap,b}\) and \(\cF_{\cN,\sigma^2}\), monotone compression rules of size at most \(4\) remove the
logarithmic dependence on \(n\).
Under suitable geometric conditions on \(r_\theta\) and \(p_\theta\), Theorem~\ref{thm:classifier_to_parameter_misspecification} (Appendix~\ref{sub:threshold_misspecification}) extends this result beyond thresholded log-likelihood rewards to reward models known up to an additive constant.

\paragraph{Comparison with~\citet{jourdan2025learning}.}
For thresholded log-likelihood rewards under \(\cF_{\cN,\Sigma}\) or \(\cF_{\lap,b}\), Theorem~\ref{thm:specialized_threshold_misspecification} recovers the fast rate of~\citet{jourdan2025learning} despite the unknown
threshold.
Both approaches use the same feasible set \(\cC_n^{\all}\), but construct its constraints from different observations.
In~\citet{jourdan2025learning}, preferences between sample pairs are observed directly; here, they are inferred by pairing every positively labeled sample with every negatively labeled one.
Accordingly, their localization is driven by comparisons between samples close to \(\theta^\star\), whereas ours is driven by cross-boundary pairs with opposite quality labels.

This distinction requires a different analysis.
The preference constraints in~\citet{jourdan2025learning} are independent, whereas the quadratic number of constraints defining \(\cC_n^{\all}\) share observations and are strongly dependent.
We control \(\cC_n^{\all}\) through a PAC-learning argument for consistent classifiers and then convert classification error into parameter error.
This yields the sharp \(\widetilde{\cO}(1/n)\) rate for univariate Gaussian and Laplace families while extending beyond their specific geometry.
Our local error-to-parameter condition, Assumption~\ref{ass:conversion_error_to_parameter_misspecification}, plays a role analogous to~\citet[Assumption~4.7]{jourdan2025learning}, but is formulated through classification error and applies beyond their specific order-statistic geometry.


\section{Minimax lower bound}
\label{sec:lower_bound}
We provide a minimax lower bound $\Omega(1/n)$, showing that our rate is optimal up to factors in $\log(n)$ and in the dimension $k$.
The proof relies on Assouad's lemma~\citep{assouad1983deux,tsybakov2009nonparametric}.
It provides lower bounds via the total variation ($\TV$) distance, which can be controlled with the squared Hellinger ($\mH$) distance as
\[
    \forall(\theta, \lambda) \in \Theta^2, \quad \TV(q_{\theta}^{\otimes n}, q_{\lambda}^{\otimes n}) \le \sqrt{2 n \mH(q_{\theta} , q_{\lambda})} \: .
\]
Assumption~\ref{ass:Hellinger_control} ensures that nearby parameters induce sufficiently similar sample--quality distributions, thereby providing the hard-to-distinguish alternatives underlying our minimax lower bound (Theorem~\ref{thm:lower_bound}). 
Although restrictive, it holds for \(\cF_{\cN,\Sigma}\), \(\cF_{\lap,b}\), and \(\cF_{\Exp}\) with thresholded log-likelihood rewards.
\begin{assumption} \label{ass:Hellinger_control} 
    There exists a set $\Theta_{1} \subseteq \Theta$ with non-empty interior and positive constants $\gamma_{\cF}(k)$, $\beta_{\cF}(k)$, and $c_1$ such that, for all $(\theta^\star,\theta) \in \Theta_{1} \times \Theta$ with $\| \theta - \theta^\star\| \in (0,\gamma_{\cF}(k))$, 
    \[
     \mH(q_{\theta^\star},q_{\theta})  \le \beta_{\cF}(k) \|\theta^\star -  \theta\| + c_1 \|\theta^\star - \theta\|^2 \: .
    \]
\end{assumption}
\begin{theorem} \label{thm:lower_bound}
    Suppose Assumption~\ref{ass:Hellinger_control} holds. 
    Then,
    \begin{align*}
        &\inf_{\widehat \theta_n : (\cX \times \{\pm 1, 0\})^{n} \to \Theta} \sup_{\theta^\star\in\Theta_1} \bE_{q_{\theta^\star}^{\otimes n}} \left[\|\widehat\theta_n-\theta^\star\|\right] \\
        &\qquad \ge \Omega\left( \min \left\{ \beta_{\cF}(k)^{-1}\sqrt{k}/n, \sqrt{ k /n}  \right\} \right) \: .
    \end{align*}
\end{theorem}
For thresholded log-likelihood rewards, the estimators based on $\cC_n$ (resp. $\cC_n^{\all}$) achieve the minimax optimal rate---up to logarithmic and dimension-dependent constants---for $\cF_{\cN, \Sigma}$ or $\cF_{\lap, b}$ by Theorem~\ref{thm:specialized_classifier_to_parameter_identifiability} (resp. Theorem~\ref{thm:specialized_threshold_misspecification}); and for $\cF_{\Exp}$ by Theorem~\ref{thm:specialized_classifier_to_parameter_ambiguity} when the threshold is known.
\paragraph{Lower-bound Intuition.}
The same boundary phenomenon driving the upper bound also explains the minimax scale.
A perturbation of size \(t\) changes labels on a region of probability mass \(\mathcal O(t)\), making the squared Hellinger distance linear rather than quadratic in \(t\).
Thus, distinguishability requires \(nt\asymp1\), yielding the \(n^{-1}\) minimax scale.
\paragraph{Dimensionality Gap.}
In higher dimensions, there is a dimensionality gap between our upper and lower bounds.
We illustrate this gap with thresholded log-likelihood rewards for $\cF_{\cN, I_d}$.
To avoid class imbalance, we consider the special case of $2\eta(d) \eqdef d\log(2\pi) + F^{-1}_{\chi^2(d)}(1/2)$, satisfying that $p_{\theta^\star}(r_{\theta^\star}>0)=1/2$.
Let $\alpha_{I_d,\eta(d)}$ as in Lemma~\ref{lem:gaussian_conversion_error_to_parameter_identifiability} and $\beta_{I_d,\eta(d)}$ as in Lemma~\ref{lem:gaussian_hellinger_control} be the explicit constants for which Theorems~\ref{thm:specialized_classifier_to_parameter_identifiability} and~\ref{thm:lower_bound} hold.
Using Stirling's formula, one can show that $\beta_{I_d,\eta(d)}/\alpha_{I_d,\eta(d)}\to_{d\to+\infty}2$ (see Appendix~\ref{sub:dimensionality_gap_between_upper_and_lower_bounds}).
Since the lower bound scales as $\Omega\left(\frac{\sqrt{d}}{n\beta_{I_d,\eta(d)}}\right)$ and the upper bound scales as $\wt O\left(\frac{d}{n\alpha_{I_d,\eta(d)}}\right)$, there is an asymptotic multiplicative gap of $\sqrt{d}$. 
Closing this gap requires either sharper upper/lower bounds or more efficient estimators that better exploit deterministic quality feedback.


\section{Experiments}
\label{sec:experiments}
In this section, we compare the empirical performance of the proposed estimators for Gaussian distributions with thresholded log-likelihood.
We defer experiments on other distributions to Appendix~\ref{app:ssec_other_distributions}: Laplace distributions with known scale or known center, Rayleigh, Exponential, and Pareto distributions (Figures~\ref{fig:OtherDistFctnErrorSupp} and~\ref{fig:OtherDistFctnErrorBisSupp}).
We consider a uniformly drawn natural parameter $\theta^\star \sim \cU([1,2]^d)$ and covariance $\Sigma=I_d$.
We compute the estimation error $\|\wh\theta_n-\theta^\star\|_2$ for $n$ up to $10^4$.
For various choices of $d$, we repeat this process over $N_{\text{runs}}$ independent instances.
The error bars show the variability of the averaged error $\|\widehat\theta_n-\theta^\star\|_2$, with symmetric and asymmetric $2$-standard-error bars computed on the original and $\log_{10}$ error scales, respectively.
\begin{figure*}[t]
    \centering
    \includegraphics[width=0.32\linewidth]{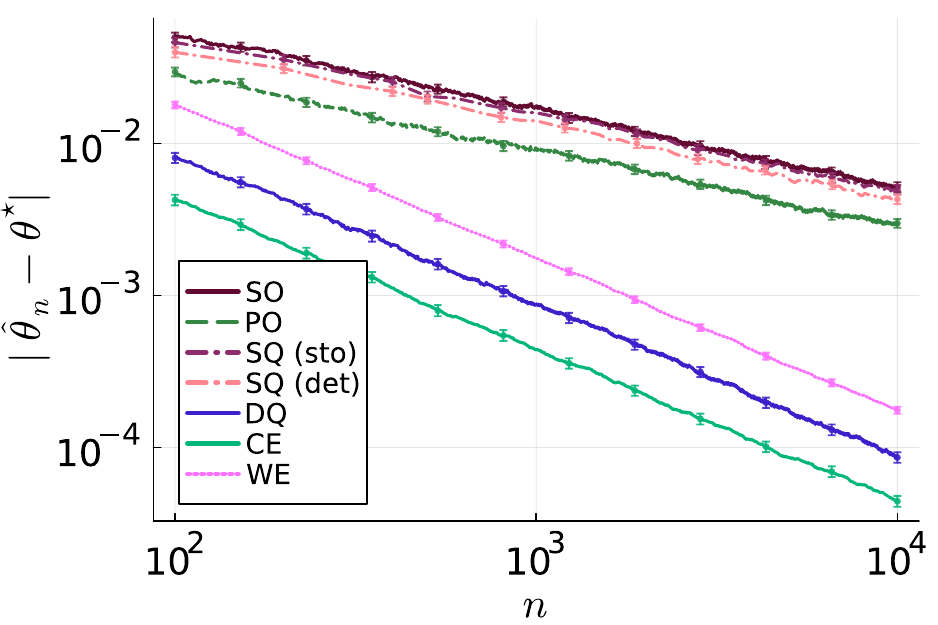}   
    \includegraphics[width=0.32\linewidth]{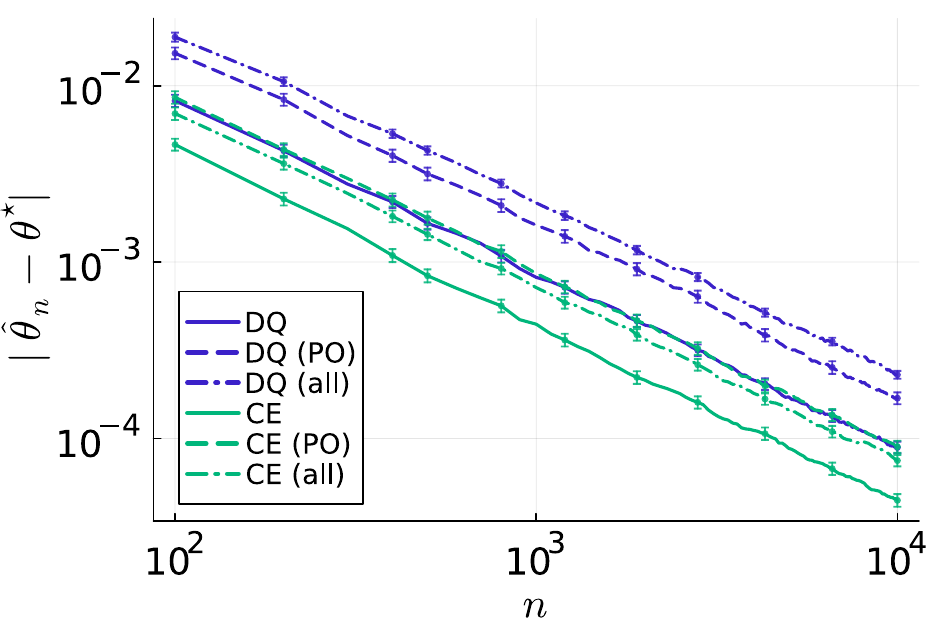}
    \includegraphics[width=0.32\linewidth]{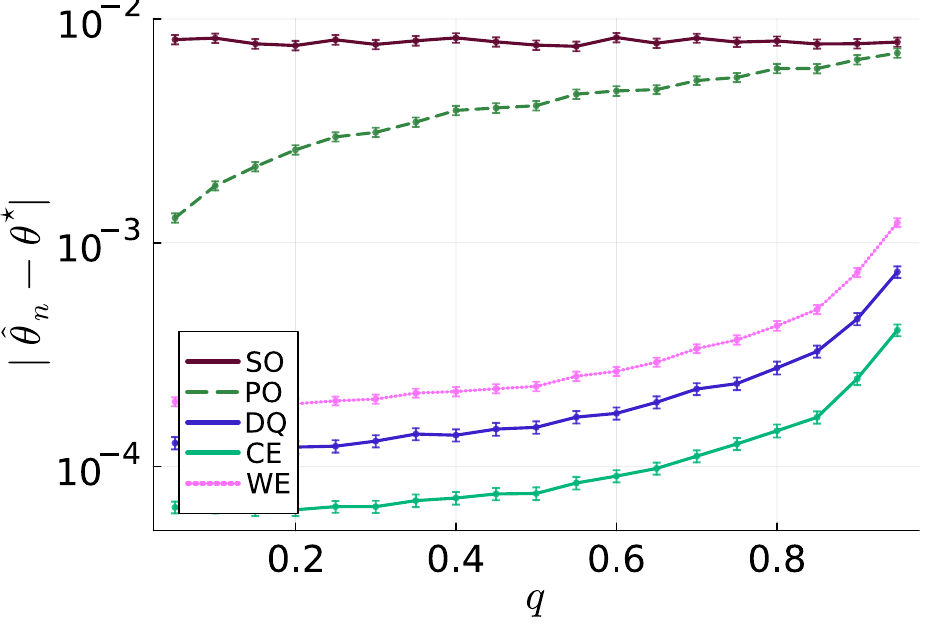} 
    \caption{Estimation errors of $\cN(\theta^\star,1)$ with $\theta^\star \sim \cU([1,2])$ and $N_{\text{runs}} = 10^3$ at (a-b) $\eta(1/2) \approx 1.15$ and (c) $n=10^4$ where $c_{\eta(q)}^2 \eqdef F^{-1}_{\chi^2(1)}(q)$. Error bars are $\pm 2$ standard errors.}
    \label{fig:Gaussian1DVariousError}
\end{figure*}
\begin{figure*}[t]
    \centering
    \includegraphics[width=0.48\linewidth]{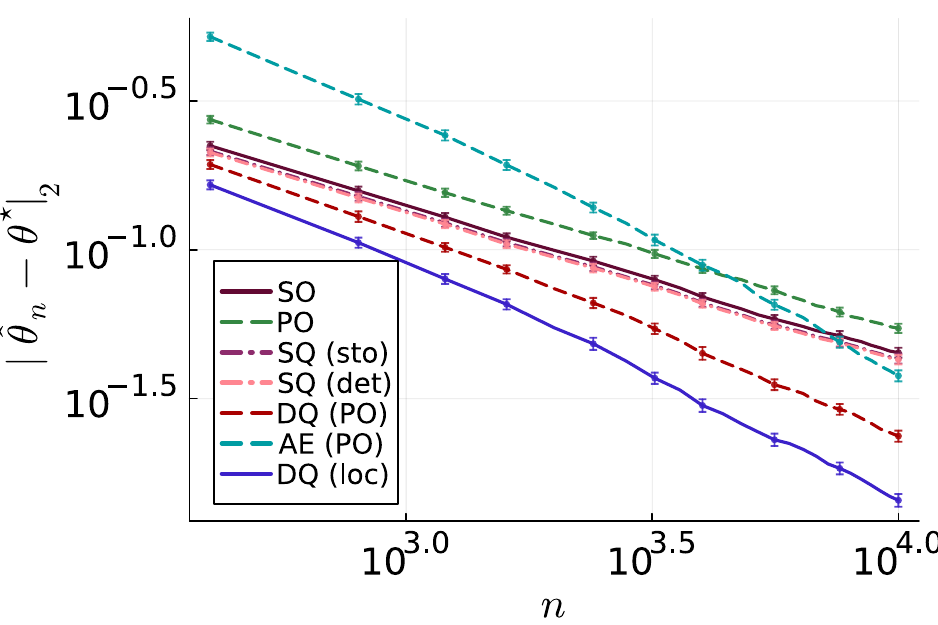} 
    \includegraphics[width=0.48\linewidth]{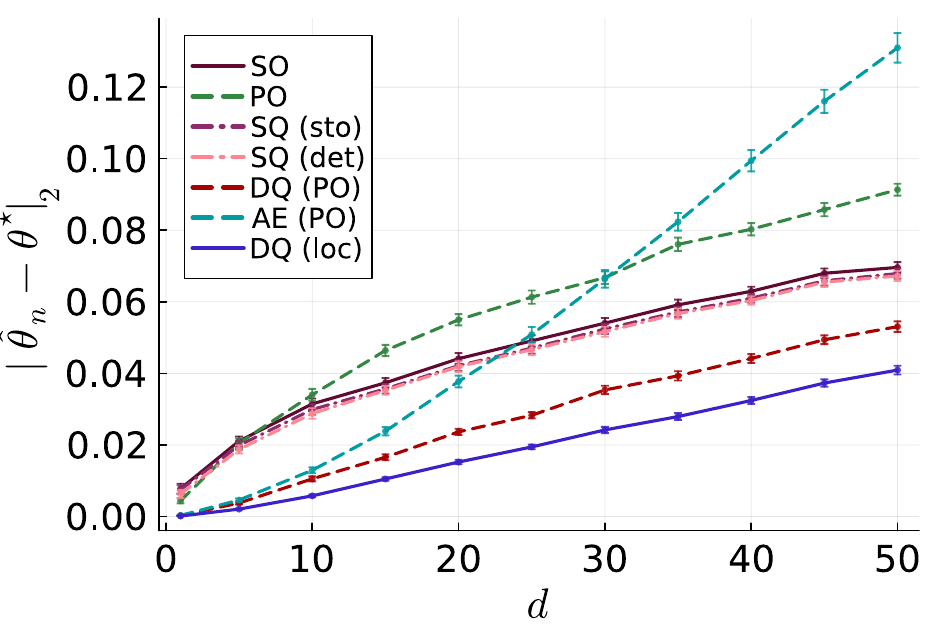} 
    \caption{Estimation errors of $\cN(\theta^\star,I_d)$ with $\theta^\star \sim \cU([1,2]^{d})$ and $N_{\text{runs}} = 10^2$ at (a) $d=20$ and (b) $n=10^4$ where $c_{\eta(d)}^2 \eqdef F^{-1}_{\chi^2(d)}(1/2)$. Error bars are $\pm 2$ standard errors.}
    \label{fig:GaussianImpactDonError}
\end{figure*}
\paragraph{Computational Cost.}
We have $\wh\theta^{\so}_{n}=\frac{1}{n}\sum_{i\in[n]}X_i$, and we define $\wh\theta^{\po}_{n} = |\cI_{n}^{+}|^{-1}\sum_{i\in[n]}\indi{Z_i=1}X_i$.
\(\sqe\) (resp. \(\lle\)) is defined in Eq.~\eqref{eq:sqmle} for stochastic (resp. deterministic) labels; it is an unconstrained convex minimization problem.
Let $c_{\eta}\eqdef \sqrt{2\eta-d\log(2\pi)}$.
To mitigate class imbalance, we use $\eta(d)$ such that $c_{\eta(d)}^2 \eqdef F^{-1}_{\chi^2(d)}(1/2)$, thus $p_{\theta^\star}(r_{\theta^\star}>0)=1/2$.
Then, 
\begin{align*}
    &\wh\theta^{\dpe}_{n} \in \argmin_{\theta\in\cC_n}\|\theta-\wh\theta^{\so}_{n}\|_2^2\: ,\\
    &\cC_n = \{\theta \mid \forall i\in[n],\: Z_i(c_{\eta}^2-\|X_i-\theta\|_2^2)\ge 0\} \: .
\end{align*}
While $\cC_n$ can be non-convex for $d\ge2$, it is an interval $\cC_n=[L_n,U_n]$ that can be efficiently implemented for $d=1$.
For $d=1$, we define the center estimator $\wh\theta^{\cee}_{n}\eqdef (L_n+U_n)/2$ and the \emph{oracle} worst-case estimator ($\we$) by $\wh\theta^{\we}_{n}\eqdef\argmax_{\theta\in \{L_n,U_n\}} |\theta-\theta^\star|$.
For $d=1$, $\cC_{n}^{+}$ and $\cC_{n}^{\all}$ are also intervals with efficient implementation.
Likewise, the $\po$ (resp. $\all$) estimators are defined based on $\cC_n^{+} = \bigcap_{i\in\cI_{n}^{+}} B_{2}(X_i,c_{\eta})$ (resp. $\cC_{n}^{\all}$) instead of $\cC_n$.
Solving $\dpe$ ($\po$) is a convex minimization problem over the convex set $\cC_{n}^{+}$.
Solving $\dpe$ ($\all$) is a convex minimization problem over the polytope $\cC_{n}^{\all}$ defined by at most a quadratic number of linear constraints.
For $d\ge2$, $\dpe$ (loc) is a local numerical solution as $\cC_n$ is non-convex, and $\anye$ ($\po$) is an arbitrary estimator ($\anye$) $\wh\theta_{n}$ in the convex set $\cC_{n}^{+}$.
See Appendix~\ref{app:ssec_implementation_details} for additional implementation details.\looseness=-1
\paragraph{Dependence on the Sample Size.}
Figure~\ref{fig:Gaussian1DVariousError}(a-b) empirically confirms the difference in estimation rates between the M-estimators ($\so$, $\sqe$, and $\lle$) and $\po$, which obtain the $\cO(1/\sqrt{n})$ rate, and our estimators based on quality constraints, which achieve the $\cO(1/n)$ rate.
Figure~\ref{fig:GaussianImpactDonError} shows that the slight advantage of $\po$ over $\so$ for $d=1$ disappears as $d$ increases.
As expected, $\sqe$ and $\lle$ slightly outperform $\so$.
Moreover, estimators using $\cC_n$ consistently outperform their counterparts using $\cC_n^{+}$ or $\cC_{n}^{\all}$.
\paragraph{Dependence on the Threshold.}
We study the effects of class imbalance by choosing thresholds $\eta(q)$ such that $c_{\eta(q)}^2 \eqdef F^{-1}_{\chi^2(1)}(q)$ so that $\bP_{\cN(\theta^\star,1)}(r_{\theta^\star}>0)=q$.
Figure~\ref{fig:Gaussian1DVariousError}(c) shows that the findings above remain robust under class imbalance.
As the constraints become less informative for larger $\eta$, performance degrades as $q\to1$.
For larger $d$, Figure~\ref{fig:GaussianLargeDErrorSupp} in Appendix~\ref{app:ssec_other_estimators} further illustrates the impact of class imbalance under alternative threshold choices.
\paragraph{Dependence on the Dimension.}
Figure~\ref{fig:GaussianImpactDonError}(b) reinforces the empirical observations above.
At a fixed sample size, the performance of $\anye$ ($\po$) deteriorates rapidly as $d$ increases, and $\dpe$ achieves the best estimation error.
Overall, $\dpe$ consistently outperforms the other estimators across all regimes.


\section{Conclusion}
\label{sec:conclusion}
We have shown that deterministic binary quality feedback can accelerate parameter estimation from the classical \(n^{-1/2}\) rate to \(\widetilde{\mathcal O}(n^{-1})\) when incorporated as hard constraints. 
Our framework covers identifiable models, including Gaussian and Laplace families, as well as finitely ambiguous models such as the exponential family, where likelihood information resolves the remaining ambiguity. 
When the reward is known only up to an additive constant, we retain the accelerated rate by constructing threshold-free pairwise preferences from oppositely labeled samples. 
Our examples further show that pointwise qualities and pairwise preferences are generally incomparable: either can localize the parameter when the other cannot. 
Finally, our minimax lower bound proves that the \(1/n\) rate is optimal, for fixed dimension, up to logarithmic and dimension-dependent factors, while our experiments confirm the benefits of combining likelihood information with deterministic quality constraints.

Several questions remain open. 
First, it would be valuable to identify minimal structural conditions on \((p_\theta,r_\theta)\) that determine whether quality feedback is more informative than preference feedback, less informative, or statistically equivalent. 
Second, understanding how the accelerated \(1/n\) rate degrades as deterministic labels become noisy would help bridge the deterministic and smooth stochastic regimes. 
Finally, closing the gap between our upper and lower bounds in higher dimensions may require sharper geometric analyses or estimators exploiting the constraint structure.

\subsection*{AI use statement}
We used generative AI tools for editing, including grammar, spelling, and word-choice suggestions.
We also used these tools to accelerate the initial implementation of experimental code and to refine constants in our initial proofs through integral computations.
All AI-assisted mathematical derivations, experimental code, and textual edits were independently reviewed and verified by the authors.
We take responsibility for the final content of this work, including all text, claims, and artifacts produced with the aid of generative AI.
\subsubsection*{Acknowledgements}
This work was supported by the Swiss National Science Foundation (grant number 212111) and by an unrestricted gift from Google.
We thank reviewers of an earlier version of this paper for insightful discussions and valuable feedback.

\bibliographystyle{plainnat}
\bibliography{main}

\begin{thebibliography}{55}
\providecommand{\natexlab}[1]{#1}
\providecommand{\url}[1]{\texttt{#1}}
\expandafter\ifx\csname urlstyle\endcsname\relax
  \providecommand{\doi}[1]{doi: #1}\else
  \providecommand{\doi}{doi: \begingroup \urlstyle{rm}\Url}\fi

\bibitem[Assouad(1983)]{assouad1983deux}
Patrice Assouad.
\newblock Deux remarques sur l'estimation.
\newblock \emph{Comptes rendus des s{\'e}ances de l'Acad{\'e}mie des sciences.
  S{\'e}rie 1, Math{\'e}matique}, 296\penalty0 (23):\penalty0 1021--1024, 1983.

\bibitem[Badrinath et~al.(2025)Badrinath, Agarwal, and Xu]{badrinathunified}
Anirudhan Badrinath, Prabhat Agarwal, and Jiajing Xu.
\newblock Unified preference optimization: Language model alignment beyond the
  preference frontier.
\newblock \emph{Transactions on Machine Learning Research}, 2025.

\bibitem[Bezanson et~al.(2017)Bezanson, Edelman, Karpinski, and
  Shah]{Julia-2017}
Jeff Bezanson, Alan Edelman, Stefan Karpinski, and Viral~B Shah.
\newblock {Julia}: A fresh approach to numerical computing.
\newblock \emph{SIAM {R}eview}, 59\penalty0 (1):\penalty0 65--98, 2017.

\bibitem[Birg{\'e} and Massart(1993)]{birge1993rates}
Lucien Birg{\'e} and Pascal Massart.
\newblock Rates of convergence for minimum contrast estimators.
\newblock \emph{Probability Theory and Related Fields}, 97:\penalty0 113--150,
  1993.

\bibitem[Blumer et~al.(1989)Blumer, Ehrenfeucht, Haussler, and
  Warmuth]{10.1145/76359.76371}
Anselm Blumer, Andrzej Ehrenfeucht, David Haussler, and Manfred~K Warmuth.
\newblock Learnability and the {Vapnik--Chervonenkis} dimension.
\newblock \emph{Journal of the ACM}, 36\penalty0 (4):\penalty0 929--965, 1989.

\bibitem[Bradley and Terry(1952)]{bradley1952rank}
Ralph~Allan Bradley and Milton~E Terry.
\newblock Rank analysis of incomplete block designs: I. the method of paired
  comparisons.
\newblock \emph{Biometrika}, 39\penalty0 (3/4):\penalty0 324--345, 1952.

\bibitem[Cadre(2006)]{cadre2006kernel}
Benoît Cadre.
\newblock Kernel estimation of density level sets.
\newblock \emph{Journal of multivariate analysis}, 97\penalty0 (4):\penalty0
  999--1023, 2006.

\bibitem[Cai et~al.(2023)Cai, Song, Jiang, Teng, Gu, and Zhang]{cai2023ulma}
Tianchi Cai, Xierui Song, Jiyan Jiang, Fei Teng, Jinjie Gu, and Guannan Zhang.
\newblock {ULMA}: Unified language model alignment with human demonstration and
  point-wise preference.
\newblock \emph{arXiv preprint arXiv:2312.02554}, 2023.

\bibitem[Cram\'er(1946)]{cramer1946mathematical}
Harald Cram\'er.
\newblock \emph{Mathematical {M}ethods of {S}tatistics}, volume~9 of
  \emph{Princeton Mathematical Series}.
\newblock Princeton University Press, Princeton, NJ, 1946.

\bibitem[Ehrenfeucht et~al.(1989)Ehrenfeucht, Haussler, Kearns, and
  Valiant]{EHRENFEUCHT1989247}
Andrzej Ehrenfeucht, David Haussler, Michael Kearns, and Leslie Valiant.
\newblock A general lower bound on the number of examples needed for learning.
\newblock \emph{Information and Computation}, 82\penalty0 (3):\penalty0
  247--261, 1989.

\bibitem[Ethayarajh et~al.(2024)Ethayarajh, Xu, Muennighoff, Jurafsky, and
  Kiela]{ethayarajh2024kto}
Kawin Ethayarajh, Winnie Xu, Niklas Muennighoff, Dan Jurafsky, and Douwe Kiela.
\newblock Model alignment as prospect theoretic optimization.
\newblock In \emph{Proceedings of the 41st International Conference on Machine
  Learning}, ICML'24, 2024.

\bibitem[Fano(1952)]{fano1952class}
RM~Fano.
\newblock Class notes for transmission of information.
\newblock In \emph{Course 6.574}. MIT Cambridge, MA, 1952.

\bibitem[Floyd and Warmuth(1995)]{floyd1995sample}
Sally Floyd and Manfred Warmuth.
\newblock Sample compression, learnability, and the vapnik-chervonenkis
  dimension.
\newblock \emph{Machine learning}, 21\penalty0 (3):\penalty0 269--304, 1995.

\bibitem[Gao et~al.(2020)Gao, Biderman, Black, Golding, Hoppe, Foster, Phang,
  He, Thite, Nabeshima, et~al.]{gao2020pile}
Leo Gao, Stella Biderman, Sid Black, Laurence Golding, Travis Hoppe, Charles
  Foster, Jason Phang, Horace He, Anish Thite, Noa Nabeshima, et~al.
\newblock The pile: An {800GB} dataset of diverse text for language modeling.
\newblock \emph{arXiv preprint arXiv:2101.00027}, 2020.

\bibitem[Ge et~al.(2024)Ge, Juba, and Vorobeychik]{ge2024learning}
Luise Ge, Brendan Juba, and Yevgeniy Vorobeychik.
\newblock Learning linear utility functions from pairwise comparison queries.
\newblock \emph{arXiv preprint arXiv:2405.02612}, 2024.

\bibitem[Hajek et~al.(2014)Hajek, Oh, and Xu]{hajek2014minimax}
Bruce Hajek, Sewoong Oh, and Jiaming Xu.
\newblock Minimax-optimal inference from partial rankings.
\newblock \emph{Advances in Neural Information Processing Systems}, 27, 2014.

\bibitem[Hanneke(2016)]{2946645.3007088}
Steve Hanneke.
\newblock Refined error bounds for several learning algorithms.
\newblock \emph{Journal of Machine Learning Research}, 17\penalty0
  (135):\penalty0 1--55, 2016.

\bibitem[Hartigan(1987)]{hartigan1987estimation}
John~A Hartigan.
\newblock Estimation of a convex density contour in two dimensions.
\newblock \emph{Journal of the American Statistical Association}, 82\penalty0
  (397):\penalty0 267--270, 1987.

\bibitem[Ibragimov and Has'minskii(1981)]{ibragimov2013statistical}
Ildar~A. Ibragimov and Rafail~Z. Has'minskii.
\newblock \emph{Statistical Estimation: Asymptotic Theory}, volume~16 of
  \emph{Stochastic Modelling and Applied Probability}.
\newblock Springer, 1981.

\bibitem[Jiang et~al.(2025)Jiang, Chen, Bai, He, Li, Yang, Zhao, Nie, and
  Zhang]{10.1145/3773279}
Ruili Jiang, Kehai Chen, Xuefeng Bai, Zhixuan He, Juntao Li, Muyun Yang, Tiejun
  Zhao, Liqiang Nie, and Min Zhang.
\newblock A survey on human preference learning for aligning large language
  models.
\newblock \emph{ACM Comput. Surv.}, 58\penalty0 (6), 2025.

\bibitem[Jourdan et~al.(2025)Jourdan, Y\"{u}ce, and
  Flammarion]{jourdan2025learning}
Marc Jourdan, Gizem Y\"{u}ce, and Nicolas Flammarion.
\newblock Learning parametric distributions from samples and preferences.
\newblock In \emph{Proceedings of the 42nd International Conference on Machine
  Learning}, volume 267 of \emph{Proceedings of Machine Learning Research},
  pages 28418--28445. PMLR, 2025.

\bibitem[Le~Cam(1973)]{lecam1973convergence}
Lucien Le~Cam.
\newblock Convergence of estimates under dimensionality restrictions.
\newblock \emph{The Annals of Statistics}, pages 38--53, 1973.

\bibitem[Lee et~al.(2022)Lee, Ippolito, Nystrom, Zhang, Eck, Callison-Burch,
  and Carlini]{lee2022deduplicating}
Katherine Lee, Daphne Ippolito, Andrew Nystrom, Chiyuan Zhang, Douglas Eck,
  Chris Callison-Burch, and Nicholas Carlini.
\newblock Deduplicating training data makes language models better.
\newblock In \emph{Proceedings of the 60th Annual Meeting of the Association
  for Computational Linguistics (Volume 1: Long Papers)}, pages 8424--8445,
  2022.

\bibitem[Lehmann and Casella(1998)]{lehmann2006theory}
Erich~Leo Lehmann and George Casella.
\newblock \emph{Theory of point estimation}.
\newblock Springer, 1998.

\bibitem[Lei et~al.(2013)Lei, Robins, and Wasserman]{lei2013distribution}
Jing Lei, James Robins, and Larry Wasserman.
\newblock Distribution-free prediction sets.
\newblock \emph{Journal of the American Statistical Association}, 108\penalty0
  (501):\penalty0 278--287, 2013.

\bibitem[Lin et~al.(2024)Lin, Gou, Gong, Liu, Shen, Xu, Lin, Yang, Jiao, Duan,
  et~al.]{lin2024not}
Zhenghao Lin, Zhibin Gou, Yeyun Gong, Xiao Liu, Yelong Shen, Ruochen Xu, Chen
  Lin, Yujiu Yang, Jian Jiao, Nan Duan, et~al.
\newblock Not all tokens are what you need for pretraining.
\newblock \emph{Advances in Neural Information Processing Systems},
  37:\penalty0 29029--29063, 2024.

\bibitem[Lubin et~al.(2023)Lubin, Dowson, Garcia, Huchette, Legat, and
  Vielma]{lubin2023jump}
Miles Lubin, Oscar Dowson, Joaquim~Dias Garcia, Joey Huchette, Beno{\^\i}t
  Legat, and Juan~Pablo Vielma.
\newblock {JuMP} 1.0: Recent improvements to a modeling language for
  mathematical optimization.
\newblock \emph{Mathematical Programming Computation}, 15\penalty0
  (3):\penalty0 581--589, 2023.

\bibitem[Meng et~al.(2024)Meng, Xia, and Chen]{meng2024simpo}
Yu~Meng, Mengzhou Xia, and Danqi Chen.
\newblock {SimPO}: Simple preference optimization with a reference-free reward.
\newblock \emph{Advances in Neural Information Processing Systems},
  37:\penalty0 124198--124235, 2024.

\bibitem[Mogensen and Riseth(2018)]{Mogensen2018}
Patrick~K. Mogensen and Asbjørn~N. Riseth.
\newblock {Optim}: A mathematical optimization package for {Julia}.
\newblock \emph{Journal of Open Source Software}, 3\penalty0 (24):\penalty0
  615, 2018.

\bibitem[Ouyang et~al.(2022)Ouyang, Wu, Jiang, Almeida, Wainwright, Mishkin,
  Zhang, Agarwal, Slama, Ray, et~al.]{ouyang2022training}
Long Ouyang, Jeffrey Wu, Xu~Jiang, Diogo Almeida, Carroll Wainwright, Pamela
  Mishkin, Chong Zhang, Sandhini Agarwal, Katarina Slama, Alex Ray, et~al.
\newblock Training language models to follow instructions with human feedback.
\newblock \emph{Advances in neural information processing systems},
  35:\penalty0 27730--27744, 2022.

\bibitem[Penedo et~al.(2023)Penedo, Malartic, Hesslow, Cojocaru, Alobeidli,
  Cappelli, Pannier, Almazrouei, and Launay]{penedo2023refinedweb}
Guilherme Penedo, Quentin Malartic, Daniel Hesslow, Ruxandra Cojocaru, Hamza
  Alobeidli, Alessandro Cappelli, Baptiste Pannier, Ebtesam Almazrouei, and
  Julien Launay.
\newblock The refinedweb dataset for {Falcon LLM}: outperforming curated
  corpora with web data only.
\newblock In \emph{Proceedings of the 37th International Conference on Neural
  Information Processing Systems}. Curran Associates Inc., 2023.

\bibitem[Penedo et~al.(2024)Penedo, Kydl{\'\i}{\v{c}}ek, Lozhkov, Mitchell,
  Raffel, Von~Werra, Wolf, et~al.]{penedo2024fineweb}
Guilherme Penedo, Hynek Kydl{\'\i}{\v{c}}ek, Anton Lozhkov, Margaret Mitchell,
  Colin Raffel, Leandro Von~Werra, Thomas Wolf, et~al.
\newblock The {FineWeb} datasets: Decanting the web for the finest text data at
  scale.
\newblock \emph{Advances in Neural Information Processing Systems},
  37:\penalty0 30811--30849, 2024.

\bibitem[Polonik(1995)]{polonik1995measuring}
Wolfgang Polonik.
\newblock Measuring mass concentrations and estimating density contour
  clusters-an excess mass approach.
\newblock \emph{The Annals of Statistics}, pages 855--881, 1995.

\bibitem[Rafailov et~al.(2023)Rafailov, Sharma, Mitchell, Manning, Ermon, and
  Finn]{rafailov2023direct}
Rafael Rafailov, Archit Sharma, Eric Mitchell, Christopher~D Manning, Stefano
  Ermon, and Chelsea Finn.
\newblock Direct preference optimization: Your language model is secretly a
  reward model.
\newblock \emph{Advances in Neural Information Processing Systems},
  36:\penalty0 53728--53741, 2023.

\bibitem[Raffel et~al.(2020)Raffel, Shazeer, Roberts, Lee, Narang, Matena,
  Zhou, Li, and Liu]{raffel2020exploring}
Colin Raffel, Noam Shazeer, Adam Roberts, Katherine Lee, Sharan Narang, Michael
  Matena, Yanqi Zhou, Wei Li, and Peter~J Liu.
\newblock Exploring the limits of transfer learning with a unified text-to-text
  transformer.
\newblock \emph{Journal of machine learning research}, 21\penalty0
  (140):\penalty0 1--67, 2020.

\bibitem[Rao(1945)]{rao1992information}
C~Radhakrishna Rao.
\newblock Information and the accuracy attainable in the estimation of
  statistical parameters.
\newblock \emph{Bull. Calcutta Math. Soc}, 37\penalty0 (3):\penalty0 81--91,
  1945.

\bibitem[Rigollet and Vert(2009)]{rigollet2009optimal}
Philippe Rigollet and R{\'e}gis Vert.
\newblock Optimal rates for plug-in estimators of density level sets.
\newblock \emph{Bernoulli}, pages 1154--1178, 2009.

\bibitem[Schulman et~al.(2017)Schulman, Wolski, Dhariwal, Radford, and
  Klimov]{schulman2017proximal}
John Schulman, Filip Wolski, Prafulla Dhariwal, Alec Radford, and Oleg Klimov.
\newblock Proximal policy optimization algorithms.
\newblock \emph{arXiv preprint arXiv:1707.06347}, 2017.

\bibitem[Shah and Wainwright(2018)]{shah2018simple}
Nihar~B Shah and Martin~J Wainwright.
\newblock Simple, robust and optimal ranking from pairwise comparisons.
\newblock \emph{Journal of machine learning research}, 18\penalty0
  (199):\penalty0 1--38, 2018.

\bibitem[Shah et~al.(2016)Shah, Balakrishnan, Bradley, Parekh, Ramchandran, and
  Wainwright]{shah2016estimation}
Nihar~B Shah, Sivaraman Balakrishnan, Joseph Bradley, Abhay Parekh, Kannan
  Ramchandran, and Martin~J Wainwright.
\newblock Estimation from pairwise comparisons: Sharp minimax bounds with
  topology dependence.
\newblock \emph{Journal of Machine Learning Research}, 17\penalty0
  (58):\penalty0 1--47, 2016.

\bibitem[Spokoiny(2012)]{spokoiny2012parametric}
Vladimir Spokoiny.
\newblock {Parametric estimation. Finite sample theory}.
\newblock \emph{The Annals of Statistics}, 40\penalty0 (6):\penalty0 2877 --
  2909, 2012.

\bibitem[Tang et~al.(2024)Tang, Guo, Zheng, Calandriello, Munos, Rowland,
  Richemond, Valko, Avila~Pires, and Piot]{tang2024generalized}
Yunhao Tang, Zhaohan~Daniel Guo, Zeyu Zheng, Daniele Calandriello, Remi Munos,
  Mark Rowland, Pierre~Harvey Richemond, Michal Valko, Bernardo Avila~Pires,
  and Bilal Piot.
\newblock Generalized preference optimization: A unified approach to offline
  alignment.
\newblock In \emph{Proceedings of the 41st International Conference on Machine
  Learning}, volume 235 of \emph{Proceedings of Machine Learning Research},
  pages 47725--47742. PMLR, 2024.

\bibitem[Tsybakov(2009)]{tsybakov2009nonparametric}
Alexandre~B Tsybakov.
\newblock \emph{Introduction to nonparametric estimation}, volume~11.
\newblock Springer, 2009.

\bibitem[Valiant(1984)]{10.1145/1968.1972}
L.~G. Valiant.
\newblock A theory of the learnable.
\newblock \emph{Communications of the ACM}, 27\penalty0 (11):\penalty0
  1134--1142, 1984.

\bibitem[van~der Vaart(1998)]{van2000asymptotic}
A.~W. van~der Vaart.
\newblock \emph{Asymptotic Statistics}.
\newblock Cambridge Series in Statistical and Probabilistic Mathematics.
  Cambridge University Press, 1998.

\bibitem[Vapnik and Chervonenkis(1971)]{vapnik1971uniform}
Vladimir~N. Vapnik and Alexey~Ya. Chervonenkis.
\newblock On the uniform convergence of relative frequencies of events to their
  probabilities.
\newblock \emph{Theory of Probability \& Its Applications}, 16\penalty0
  (2):\penalty0 264--280, 1971.

\bibitem[W{\"a}chter and Biegler(2006)]{wachter2006implementation}
Andreas W{\"a}chter and Lorenz~T Biegler.
\newblock On the implementation of an interior-point filter line-search
  algorithm for large-scale nonlinear programming.
\newblock \emph{Mathematical programming}, 106:\penalty0 25--57, 2006.

\bibitem[Wang et~al.(2023)Wang, Kordi, Mishra, Liu, Smith, Khashabi, and
  Hajishirzi]{wang2023self}
Yizhong Wang, Yeganeh Kordi, Swaroop Mishra, Alisa Liu, Noah~A Smith, Daniel
  Khashabi, and Hannaneh Hajishirzi.
\newblock Self-instruct: Aligning language models with self-generated
  instructions.
\newblock In \emph{Proceedings of the 61st annual meeting of the association
  for computational linguistics (volume 1: long papers)}, pages 13484--13508,
  2023.

\bibitem[Wasserman(2004)]{wasserman2013all}
Larry Wasserman.
\newblock \emph{All of statistics: a concise course in statistical inference},
  volume~26.
\newblock Springer, 2004.

\bibitem[Xie et~al.(2023)Xie, Pham, Dong, Du, Liu, Lu, Liang, Le, Ma, and
  Yu]{xie2023doremi}
Sang~Michael Xie, Hieu Pham, Xuanyi Dong, Nan Du, Hanxiao Liu, Yifeng Lu,
  Percy~S Liang, Quoc~V Le, Tengyu Ma, and Adams~Wei Yu.
\newblock {DoReMi}: Optimizing data mixtures speeds up language model
  pretraining.
\newblock \emph{Advances in Neural Information Processing Systems},
  36:\penalty0 69798--69818, 2023.

\bibitem[Yao et~al.(2025)Yao, He, and Gastpar]{yao2025leveraging}
Yunzhen Yao, Lie He, and Michael Gastpar.
\newblock Leveraging sparsity for sample-efficient preference learning: A
  theoretical perspective.
\newblock In \emph{International Conference on Machine Learning}, pages
  71739--71764. PMLR, 2025.

\bibitem[Yu et~al.(2024)Yu, Das, and Xiong]{yu2024mates}
Zichun Yu, Spandan Das, and Chenyan Xiong.
\newblock {MATES}: Model-aware data selection for efficient pretraining with
  data influence models.
\newblock \emph{Advances in Neural Information Processing Systems},
  37:\penalty0 108735--108759, 2024.

\bibitem[Zhou et~al.(2023)Zhou, Liu, Xu, Iyer, Sun, Mao, Ma, Efrat, Yu, Yu,
  et~al.]{zhou2023lima}
Chunting Zhou, Pengfei Liu, Puxin Xu, Srinivasan Iyer, Jiao Sun, Yuning Mao,
  Xuezhe Ma, Avia Efrat, Ping Yu, Lili Yu, et~al.
\newblock {LIMA}: Less is more for alignment.
\newblock \emph{Advances in Neural Information Processing Systems},
  36:\penalty0 55006--55021, 2023.

\bibitem[Zhu et~al.(2023)Zhu, Jordan, and Jiao]{zhu2023principled}
Banghua Zhu, Michael Jordan, and Jiantao Jiao.
\newblock Principled reinforcement learning with human feedback from pairwise
  or k-wise comparisons.
\newblock In \emph{International Conference on Machine Learning}, pages
  43037--43067. PMLR, 2023.

\bibitem[Ziegler et~al.(2019)Ziegler, Stiennon, Wu, Brown, Radford, Amodei,
  Christiano, and Irving]{ziegler2019fine}
Daniel~M Ziegler, Nisan Stiennon, Jeffrey Wu, Tom~B Brown, Alec Radford, Dario
  Amodei, Paul Christiano, and Geoffrey Irving.
\newblock Fine-tuning language models from human preferences.
\newblock \emph{arXiv preprint arXiv:1909.08593}, 2019.

\end{thebibliography}

\section*{Checklist}

\begin{enumerate}

  \item For all models and algorithms presented, check if you include:
  \begin{enumerate}
    \item A clear description of the mathematical setting, assumptions, algorithm, and/or model. Yes. See Sections~\ref{sec:problem_statement},~\ref{sec:estimator4qualities},~\ref{sec:reduction_to_preferences},~\ref{sec:lower_bound} and~\ref{sec:experiments}.
    \item An analysis of the properties and complexity (time, space, sample size) of any algorithm. Yes. See Sections~\ref{sec:estimator4qualities} and~\ref{sec:experiments}, as well as Appendix~\ref{app:ssec_implementation_details}.
    \item (Optional) Anonymized source code, with specification of all dependencies, including external libraries. Yes. See Appendix~\ref{app:ssec_implementation_details} and the code provided in the supplementary material.
  \end{enumerate}

  \item For any theoretical claim, check if you include:
  \begin{enumerate}
    \item Statements of the full set of assumptions of all theoretical results. Yes. See Sections~\ref{sec:estimator4qualities},~\ref{sec:reduction_to_preferences} and~\ref{sec:lower_bound}, as well as Appendices~\ref{app:discussions},~\ref{app:qualities_estimators},~\ref{app:reduction_to_preferences},~\ref{app:lower_bound} and~\ref{app:mestimator}.
    \item Complete proofs of all theoretical results. Yes. See Sections~\ref{sec:estimator4qualities},~\ref{sec:reduction_to_preferences} and~\ref{sec:lower_bound}, as well as Appendices~\ref{app:discussions},~\ref{app:qualities_estimators},~\ref{app:reduction_to_preferences},~\ref{app:lower_bound} and~\ref{app:mestimator}.
    \item Clear explanations of any assumptions. Yes. See Sections~\ref{sec:estimator4qualities},~\ref{sec:reduction_to_preferences} and~\ref{sec:lower_bound}, as well as Appendices~\ref{app:discussions},~\ref{app:qualities_estimators},~\ref{app:reduction_to_preferences},~\ref{app:lower_bound} and~\ref{app:mestimator}.
  \end{enumerate}

  \item For all figures and tables that present empirical results, check if you include:
  \begin{enumerate}
    \item The code, data, and instructions needed to reproduce the main experimental results (either in the supplemental material or as a URL). Yes. See Appendix~\ref{app:ssec_implementation_details} and the code provided in the supplementary material.
    \item All the training details (e.g., data splits, hyperparameters, how they were chosen). Yes. See details in Section~\ref{sec:experiments} and Appendix~\ref{app:ssec_implementation_details}.
    \item A clear definition of the specific measure or statistics and error bars (e.g., with respect to the random seed after running experiments multiple times). Yes. See details in Section~\ref{sec:experiments}.
    \item A description of the computing infrastructure used. (e.g., type of GPUs, internal cluster, or cloud provider). Yes. See details in Appendix~\ref{app:ssec_implementation_details}.
  \end{enumerate}

  \item If you are using existing assets (e.g., code, data, models) or curating/releasing new assets, check if you include:
  \begin{enumerate}
    \item Citations of the creator If your work uses existing assets. Not Applicable.
    \item The license information of the assets, if applicable. Not Applicable.
    \item New assets either in the supplemental material or as a URL, if applicable. Not Applicable.
    \item Information about consent from data providers/curators. Not Applicable.
    \item Discussion of sensible content if applicable, e.g., personally identifiable information or offensive content. Not Applicable.
  \end{enumerate}

  \item If you used crowdsourcing or conducted research with human subjects, check if you include:
  \begin{enumerate}
    \item The full text of instructions given to participants and screenshots. Not Applicable.
    \item Descriptions of potential participant risks, with links to Institutional Review Board (IRB) approvals if applicable. Not Applicable.
    \item The estimated hourly wage paid to participants and the total amount spent on participant compensation. Not Applicable.
  \end{enumerate}

\end{enumerate}

\clearpage
\appendix
\thispagestyle{empty}

\onecolumn


\section{Outline}
\label{app:outline}
The appendices are organized as follows:
\begin{itemize}
    \item In Appendix~\ref{app:discussions}, we provide additional related work, detailed discussions on our geometric assumptions, sources of misspecification, and the dimensionality gap between our upper and lower bounds.
    \item Appendix~\ref{app:qualities_estimators} contains the proofs of the results presented in Section~\ref{sec:estimator4qualities}: Theorems~\ref{thm:specialized_classifier_to_parameter_identifiability},~\ref{thm:specialized_classifier_to_parameter_ambiguity},~\ref{thm:classifier_to_parameter_identifiability}, and~\ref{thm:classifier_to_parameter_ambiguity}.
    For thresholded log-likelihood rewards, Assumption~\ref{ass:conversion_error_to_parameter_identifiability} is shown for $\cF_{\cN, \Sigma}$ (Lemma~\ref{lem:gaussian_conversion_error_to_parameter_identifiability}) and $\cF_{\lap, b}$ (Lemma~\ref{lem:laplace_conversion_error_to_parameter_identifiability}); and Assumption~\ref{ass:conversion_error_to_parameter_ambiguity} is shown for $\cF_{\Exp}$ (Lemma~\ref{lem:exponential_conversion_error_to_parameter_ambiguity}). 
    We also detail several examples where preference feedback is more informative than quality feedback.
    \item Appendix~\ref{app:reduction_to_preferences} contains the proof of Theorem~\ref{thm:specialized_threshold_misspecification} presented in Section~\ref{sec:reduction_to_preferences}, and its extension, Theorem~\ref{thm:classifier_to_parameter_misspecification} in Appendix~\ref{sub:threshold_misspecification}, under suitable assumptions on \(r_\theta\) and \(p_\theta\) (Assumptions~\ref{ass:finite_VC_misspecification},~\ref{ass:conversion_error_to_parameter_misspecification} and~\ref{ass:closure_strict_consistency_misspecification}).
    \item Appendix~\ref{app:lower_bound} contains the proofs of the results presented in Section~\ref{sec:lower_bound}: Theorem~\ref{thm:lower_bound}, and Assumption~\ref{ass:Hellinger_control} for thresholded log-likelihood rewards with $\cF_{\cN, \Sigma}$ (Lemma~\ref{lem:gaussian_hellinger_control}), $\cF_{\lap, b}$ (Lemma~\ref{lem:laplace_hellinger_control}) and $\cF_{\Exp}$ (Lemma~\ref{lem:exponential_hellinger_control}).
    \item In Appendix~\ref{app:mestimator}, we discuss the asymptotic behavior of $\sqe$ and $\lle$ from Section~\ref{ssec:QbasedMestimator} in greater depth.
    \item In Appendix~\ref{app:experiments}, we provide implementation details for reproducibility, extended discussions on the computational cost of our estimators, and supplementary experiments to support our theoretical findings. The code is available as supplementary material.
\end{itemize}


\section{Extended Discussions}
\label{app:discussions}
We provide extended discussions on additional related work (Appendix~\ref{app:ssec_related_work}), our geometric assumptions (Appendix~\ref{app:ssec_weakening_verifying_assumptions}), sources of misspecification (Appendix~\ref{app:ssec_misspecification}), and the dimensionality gap between our upper and lower bounds (Appendix~\ref{sub:dimensionality_gap_between_upper_and_lower_bounds}).
\subsection{Additional Related Work}
\label{app:ssec_related_work}
\paragraph{Quality-aware Data Selection in Pretraining.}
Large-scale pretraining pipelines rely on curated, filtered, and deduplicated corpora such as C4, the Pile, RefinedWeb, and FineWeb~\citep{raffel2020exploring,gao2020pile,lee2022deduplicating,penedo2023refinedweb,penedo2024fineweb}. 
Beyond static corpus construction, adaptive methods such as DoReMi, Rho-1, and MATES select or reweight data using domain, reference-model, or influence-based signals~\citep{xie2023doremi,lin2024not,yu2024mates}. 
These works demonstrate the empirical importance of quality signals throughout the training pipeline. 
\paragraph{Fine-tuning and Alignment with Preference/Quality Data.}
Language-model post-training increasingly relies on feedback beyond next-token prediction. 
RLHF learns a reward model from human preferences and then optimizes the policy, e.g., PPO~\citep{ziegler2019fine,schulman2017proximal,ouyang2022training}, whereas DPO and related direct-alignment methods replace the reinforcement-learning step with supervised preference objectives~\citep{rafailov2023direct,meng2024simpo,tang2024generalized}. 
Although most alignment methods use pairwise preferences, recent approaches also exploit pointwise or binary quality signals: ULMA combines demonstrations with pointwise preferences~\citep{cai2023ulma}, KTO trains on desirable/undesirable labels~\citep{ethayarajh2024kto}, and UPO incorporates auxiliary objectives that are not naturally expressed as pairwise comparisons~\citep{badrinathunified}; see~\citet{10.1145/3773279} for a survey. 
These empirical works motivate the study of quality feedback. 
Our contribution is theoretical: we isolate a parametric statistical setting in which quality labels can be analyzed directly, and we characterize when such labels improve estimation rates rather than merely constant factors.
\subsection{Geometric Assumptions}
\label{app:ssec_weakening_verifying_assumptions}
\paragraph{Finite VC Dimension.}
Assumption~\ref{ass:finite_VC_misspecification} implies both Assumptions~\ref{ass:finite_VC_ambiguity} and~\ref{ass:finite_VC_identifiability}; more precisely, $d_\star^{+}\leq d_\star\leq d^{\all}$.
Indeed, \(h_\theta^{+}\) is obtained from \(h_\theta\) by restricting its disagreement region to the fixed set \(\{r_{\theta^\star}>0\}\), which cannot increase VC dimension.
Moreover, \(h_\theta\) is the disagreement classifier between the fixed classifier \(h_{\theta^\star,0}^{\all}\) and \(\{h_{\theta,0}^{\all}:\theta\in\Theta\}\subseteq\cH^{\all}\).
Taking disagreement with a fixed classifier merely flips a sample-dependent fixed subset of labels and therefore preserves VC dimension.

Finite VC dimension is a standard distribution-free complexity condition. 
It can usually be verified by identifying the geometry of the thresholded reward regions \(\{x:r_\theta(x)>0\}\), or of their pairwise analogues, and invoking standard VC bounds for the resulting class. 
In particular, half-spaces, intervals, balls, and, more generally, semi-algebraic sets with uniformly bounded description complexity have finite VC dimension.
This directly covers our Gaussian, Laplace, and exponential examples.
Finite-dimensional parametrization alone is not sufficient: a pathological parametrization may encode arbitrarily complex decision regions and hence have infinite VC dimension.

The VC assumptions can also be weakened. 
Since our proofs only use a uniform bound on the disagreement class under \(p_{\theta^\star}\), they may be replaced by an appropriate distribution-dependent uniform-convergence condition, for example one based on covering numbers, bracketing entropy, or a localized Rademacher complexity.
Such replacements can be useful when the class has infinite global VC dimension but substantially smaller complexity in a neighborhood of \(\theta^\star\).
\paragraph{Local Error-to-parameter Conversion.}
Assumption~\ref{ass:conversion_error_to_parameter_identifiability} implies Assumption~\ref{ass:conversion_error_to_parameter_ambiguity}.
Indeed, \(\cD^+(\theta^\star,\theta)\subseteq\cD(\theta^\star,\theta)\), and the former assumption implies
\(\cC(\theta^\star)=\{\theta^\star\}\). 
Thus the latter holds with \(\alpha_\star=\alpha_\star^+\) and \(\varepsilon_\star=\varepsilon_\star^+\).
The converse need not hold: full feedback can distinguish parameters that agree with \(\theta^\star\) throughout its positive region.

These assumptions require the disagreement probability to grow at least linearly with the distance to the relevant equivalence class in a neighborhood of \(\theta^\star\). 
The truncation by \(\varepsilon_\star^+\) or \(\varepsilon_\star\) only requires a uniform positive separation outside this neighborhood. 
A useful way to verify them is through a local boundary argument. 
Suppose that the zero level set \(\{x:r_{\theta^\star}(x)=0\}\) is a regular hypersurface, that \(p_{\theta^\star}\) is bounded below in a neighborhood of this boundary, and that the boundary moves transversally in every identifiable parameter direction. 
Then a perturbation of size \(\|\theta-\theta^\star\|\) displaces the reward boundary by the same order. 
If \(p_{\theta^\star}\) is bounded below near that boundary, the region lying between the original and perturbed boundaries therefore has probability mass proportional to \(\|\theta-\theta^\star\|\). 
This yields the desired local linear bound. 
Uniform separation outside this local neighborhood can then often be obtained by continuity and compactness, after identifying parameters belonging to the same finite ambiguity class.

This perspective also clarifies the possible failures: the feedback may be insensitive to some parameter directions; symmetries may produce a non-trivial equivalence class; the boundary may have zero density; or its displacement may vanish to first order. 
The latter case need not preclude consistency, but generally yields a weaker modulus. 
More generally, the linear lower bound can be replaced by
\[
    p_{\theta^\star}(\cD(\theta^\star,\theta)) \geq c\,\mathrm{dist}(\theta,\cC(\theta^\star))^\kappa
\]
locally, for some \(\kappa>0\), and analogously for positive-only
feedback. 
The same proof strategy then yields a parameter rate of order \(\widetilde{\mathcal O}(n^{-1/\kappa})\); the present assumptions correspond to the regular, transversal case \(\kappa=1\).

The profiled error condition (Assumption~\ref{ass:conversion_error_to_parameter_misspecification}) for unknown thresholds is stronger still. 
Since $\mathrm{er}_{\mathrm{prof}}(\theta^\star,\theta) \leq p_{\theta^\star}(\cD(\theta^\star,\theta))$, a lower bound on the profiled error implies the corresponding full-feedback conversion bound, but not conversely. 
It rules out parameter changes that can be compensated by shifting the threshold: it therefore requires the \emph{shape} of the reward level sets, and not merely one thresholded positive region, to identify the parameter.
\paragraph{Likelihood Concentration.}
Assumption~\ref{ass:uniform_bernstein_bound} is a local uniform law of large numbers with Bernstein-type tails for the negative log-likelihood. 
It is standard for regular parametric models. 
On each compact \(K\subset\Theta\), it can typically be verified by combining continuity (or local Lipschitz continuity) of \(\theta\mapsto-\log p_\theta(x)\) with an integrable envelope for the class \(\{-\log p_\theta+\log p_{\theta^\star}:\theta\in K\}\), and suitable uniform variance or exponential-moment bounds. 
Standard empirical process arguments then yield the displayed \(\sqrt{\log(1/\delta)/n}+\log(1/\delta)/n\) control. These conditions hold for the Gaussian, Laplace, and exponential families considered in this paper.

Its role is only to select, among the finitely many feedback-compatible components, the component containing \(\theta^\star\). 
In particular, the proof does not rely on the specific Bernstein rate: any high-probability uniform deviation bound that converges to zero on the relevant compact set is sufficient for eventual component selection. 
The assumption can therefore be weakened to a uniform law of large numbers, possibly with an unspecified deterministic modulus \(a_n(\delta)\to0\). 
One may also replace the global compact-set condition by concentration on a random or deterministic neighborhood known to contain the feedback version space with high probability.
\paragraph{Closure of Strict Consistency.}
Assumption~\ref{ass:closure_strict_consistency_misspecification} is a geometric regularity condition: $\cC_n^{\all}=\overline{\cC_{n,>}^{\all}}$ almost surely.
It excludes isolated weakly feasible parameters that satisfy one or more sample constraints only through exact equalities and cannot be approximated by strictly feasible parameters. 
Its sole role is to transfer the PAC bound established for \(\cC_{n,>}^{\all}\) to the weak version space \(\cC_n^{\all}\).

A direct way to verify the condition is to show that every weakly feasible \(\theta\) can be moved by an arbitrarily small amount toward a strictly feasible point while preserving all weak inequalities. 
For example, if the sample constraints are affine in \(\theta\), the weak and strict feasible sets are respectively an intersection of closed and open half-spaces. 
If \(\theta^\star\) satisfies all constraints strictly, then, for every \(\theta\in\cC_n^{\all}\) and \(t\in(0,1]\), the point $(1-t)\theta+t\theta^\star$ is strictly feasible and converges to \(\theta\) as \(t\downarrow0\).
This verifies the assumption for Gaussian log-likelihood rewards.
More generally, the same argument applies whenever the feasible set is star-shaped with respect to a strictly feasible point, or whenever a suitable constraint qualification guarantees that weak feasibility can be perturbed into strict feasibility.

The condition can be weakened in several equivalent ways. 
It is enough that every weakly feasible parameter have the same classifier, or the same relevant error, as a limit of strictly feasible parameters. 
Alternatively, one may formulate the PAC guarantee directly for the weak constraint class, using a fixed convention for ties. 
This avoids the closure assumption altogether, at the cost of controlling classifiers arising from equality cases. 
In settings where ties occur with probability zero, it also suffices to establish the closure property outside a null event.
\paragraph{Hellinger Control for the Lower Bound.}
Assumption~\ref{ass:Hellinger_control} ensures that two nearby parameters induce sufficiently similar sample--quality distributions.
It supplies the hard-to-distinguish local alternatives used in the minimax lower bound of Theorem~\ref{thm:lower_bound}. 
The linear term is specific to deterministic quality feedback: even when the underlying statistical model is quadratically regular in Hellinger distance, moving a deterministic decision boundary generally changes the labels on a region of probability mass proportional to the boundary displacement.

A useful route for verifying the assumption is to decompose the distance between \(q_{\theta^\star}\) and \(q_\theta\) into two effects: the change in the marginal sample distribution \(p_{\theta^\star}\) versus \(p_\theta\), and the change in the thresholded reward region. 
For regular parametric families, the first is typically quadratic, $\mH(p_{\theta^\star},p_\theta) \leq C\|\theta-\theta^\star\|^2$ locally. 
For the deterministic labels, it then remains to bound the probability mass of the region on which the two quality rules differ.
A regular-boundary argument---using a bounded density near \(\{r_{\theta^\star}=0\}\) and a locally Lipschitz, transversal motion of this boundary---typically yields a linear bound. 
This is the mechanism used in the Gaussian, Laplace, and exponential examples.

The assumption may be weakened by replacing the displayed linear--quadratic modulus with any local modulus
\[
    \mH(q_{\theta^\star},q_\theta)
    \leq \omega(\|\theta-\theta^\star\|),
    \qquad \omega(t)\downarrow0.
\]
The resulting lower bound is then determined by the inverse modulus at scale \(1/n\). 
For example, if \(\omega(t)\lesssim t^\kappa\), the same two-point argument yields a local minimax lower bound of order \(n^{-1/\kappa}\). 
Thus the current assumption captures the regular deterministic-boundary regime \(\kappa=1\), which leads to the fast \(n^{-1}\) lower-bound scale.
It is also enough for the assumption to hold on a nonempty open subset of parameters, as stated: minimax lower bounds are local, so no global control over the entire parameter space is needed.
\subsection{Sources of Misspecification}
\label{app:ssec_misspecification}
\paragraph{Parameter Space.}
When $\theta^\star \notin \Theta$, separability within $\Theta$ is not guaranteed by deterministic qualities alone. 
The definition of \ref{eq:dqle} should be modified to combine the cross-entropy loss and the classification $0$-$1$ loss, i.e.,
\begin{equation} \label{eq:dqle_misspecified}
    \wh \theta^{\dpe}_{n} \in \argmin_{\theta \in \Theta} \left\{ L^{\so}_{n}(\theta) + \lambda \sum_{i \in [n]} \indi{ Z_i r_{\theta}(X_i) < 0}  \right\} \: ,
\end{equation}
where $\lambda > 0$ is a regularization between those two losses.
Without separability, solving Eq.~\eqref{eq:dqle_misspecified} can be NP-hard. 
Under sufficient regularity, $\wh \theta^{\dpe}_{n}$ converges to $\theta_{0} \in \argmin_{\theta \in \Theta} \{\KL(\theta^\star,\theta) + \lambda m(\theta^\star,\theta)\}$ where $m(\theta^\star,\theta) \eqdef p_{\theta^\star} \left( \cD(\theta^\star, \theta) \right)$ and $\theta_0 \ne \theta^\star$. 
Computing $\theta_{0}$ might be challenging since $\theta \mapsto m(\theta^\star,\theta)$ can be non-convex. 
To obtain tractable and robust estimators, we need a tractable surrogate or relaxation for this optimization problem. 
Since $\theta_0$ lies on the boundary of $\Theta$, proving an accelerated rate requires controlling the maximal deviation from $\theta_0$ along directions that point toward the interior of $\Theta$. 
Parts of our analysis may still apply, yet sharper technical arguments are needed.
\paragraph{Parametric Models.}
The true distribution $p^\star$ of the observations may not belong to the model class $\mathcal F$, that is, $p^\star \notin \mathcal F$. 
Such misspecification arises when $\mathcal F$ fails to capture the true data-generating structure, for instance because the truth lies in another parametric or non-parametric class of distributions.
In this setting, $L^{\so}_{n}(\theta)$ can be interpreted as a quasi-log-likelihood. 
We refer to the estimator defined in Eq.~\eqref{eq:somle} for this quasi-log-likelihood as the $\so$ quasi-MLE. 
Under suitable regularity conditions, the $\so$ quasi-MLE converges to $\theta_{0} \in \argmin_{\theta \in \Theta} \KL(p^{\star}, p_{\theta})$ where, in general, $p^\star \neq p_{\theta_0}$ even though $p_{\theta_0} \in \mathcal F$.

Similarly, the true reward $r^\star$ associated with $p^\star$ may not belong to the class of reward models
In the absence of separability induced by well-specified deterministic qualities, we define the $\dpe$ quasi-MLE as in Eq.~\eqref{eq:dqle_misspecified}. 
Under suitable regularity conditions, the $\dpe$ quasi-MLE converges to $\theta_{0} \in \argmin_{\theta \in \Theta} \{\KL(p^{\star}, p_{\theta}) + \lambda p^\star \left( \cD(\theta) \right)\}$ where $\cD(\theta) \eqdef \{x \in \cX \mid r^{\star}(x) r_{\theta}(x) < 0\}$.
\paragraph{Brittleness of Exact Interpolation.}
Exact feasibility is sensitive to persistent errors in the feedback model.
For example, suppose that the labels are generated using the threshold \(\eta^\star\), whereas the constraints use \(\eta\), and let
\[
    m(\eta,\eta^\star) \eqdef p_{\theta^\star}\!\left( \sign(\log p_{\theta^\star}(X)+\eta)\ne\sign(\log p_{\theta^\star}(X)+\eta^\star) \right) \: .
\]
Then the true parameter satisfies all full-feedback constraints with probability \((1-m(\eta,\eta^\star))^n\). 
Whenever \(m(\eta,\eta^\star)>0\), this probability converges exponentially fast to zero. 
Under the regular-boundary conditions discussed in Appendix~\ref{app:ssec_weakening_verifying_assumptions}, one typically has \(m(\eta,\eta^\star)\asymp|\eta-\eta^\star|\) locally. 
Similarly, under independent label flips with probability \(\rho\), the probability that \(\theta^\star\) interpolates all observed labels is \((1-\rho)^n\).
Consequently, any fixed nonzero threshold error or flip probability invalidates exact interpolation asymptotically; for interpolation to hold with probability tending to one, one needs, respectively, \(n|\eta_n-\eta^\star|\to0\) or \(n\rho_n\to0\), up to the corresponding boundary-mass constants. 
The all-pairs construction $\cC_{n}^{\all}$ removes a common additive threshold error, when the resulting preferences remain identifying, but does not by itself protect against corrupted labels, which may contaminate many constructed comparisons.
\paragraph{Intermediate Noise Regimes.}
The failure of exact interpolation should be distinguished from the loss of the \(1/n\) statistical rate. 
In particular, a fixed symmetric flip probability \(\rho<1/2\) destroys interpolation but leaves a discontinuity of size \(1-2\rho\) in the conditional label distribution across the quality boundary. 
Robust estimators that allow label violations may therefore retain change-point-type \(1/n\) rates, with constants deteriorating as \(\rho\) approaches \(1/2\). 
This differs fundamentally from the smooth sigmoid model of Section~\ref{ssec:QbasedMestimator}: there, the conditional label distribution varies smoothly with the parameter, the experiment is locally regular, and the classical \(n^{-1/2}\) rate is recovered. 
More generally, the transition is governed by the local modulus of separation between the induced sample--label distributions, rather than by the mere presence of noise. 
For instance, if the label discontinuity has magnitude \(a_n\) and the local squared Hellinger distance behaves heuristically as
\[
    H^2(q_{\theta^\star},q_\theta)
    \asymp
    a_n^2\|\theta-\theta^\star\|
    + \|\theta-\theta^\star\|^2,
\]
the corresponding distinguishability scale solves \(n(a_n^2t+t^2)\asymp1\): it interpolates between \(1/(na_n^2)\) and \(1/\sqrt n\), with a transition around \(a_n\asymp n^{-1/4}\). 
Establishing matching guarantees for robust, slackened constraints under general corruption mechanisms is an interesting direction for future work.

\subsection{Dimensionality Gap between Upper and Lower Bounds}
\label{sub:dimensionality_gap_between_upper_and_lower_bounds}

We detail the dimensional dependence discussed in Section~\ref{sec:lower_bound} for the Gaussian location family
\(\cF_{\cN,I_d}\) with thresholded log-likelihood rewards.
Let $m_d \eqdef F_{\chi^2(d)}^{-1}(1/2)$ and $2\eta(d) \eqdef d\log(2\pi)+m_d $.
Since
\[
    -2\log p_{\theta^\star}(X)-d\log(2\pi) = \|X-\theta^\star\|_2^2 \sim \chi^2(d) \: ,
\]
this choice yields balanced quality labels: $p_{\theta^\star}\bigl(r_{\theta^\star}(X)>0\bigr) = 1/2$.

We compare the constants \(\alpha_{I_d,\eta(d)}\) and \(\beta_{I_d,\eta(d)}\) appearing respectively in the upper bound of Theorem~\ref{thm:specialized_classifier_to_parameter_identifiability} and the lower bound of Theorem~\ref{thm:lower_bound}.
Throughout this subsection, we consider \(d\ge 3\), which is sufficient for the asymptotic comparison.

\paragraph{Asymptotic Behavior of the Upper-bound Constant.}

By Lemma~\ref{lem:gaussian_conversion_error_to_parameter_identifiability}, for $d\ge 2$,
\[
    \alpha_{I_d,\eta(d)} = \kappa_{d-1}s_d (2\pi)^{-d/2} e^{-m_d/2} m_d^{(d-1)/2} \: ,
\]
where $\kappa_{d-1} = \frac{\pi^{(d-1)/2}}{\Gamma\!\left(\frac{d+1}{2}\right)}$ and $s_d = \left(1-\frac{1}{16d^2}\right)^{(d-1)/2}$. 
Clearly, \(s_d=1+\mathcal O(d^{-1})\).
Set $z_d\eqdef\frac{d+1}{2}$ and $x_d\eqdef\frac{m_d}{2}$.
The standard expansion of the median of a chi-squared distribution gives
\[
    m_d=d-\frac23+\mathcal O(d^{-1}),    \qquad    x_d=z_d-\frac56+\mathcal O(d^{-1}) \: .
\]
Rearranging the expression above yields
\[
    \alpha_{I_d,\eta(d)}= \frac{s_d}{\sqrt{2\pi}} \frac{x_d^{z_d-1}e^{-x_d}}{\Gamma(z_d)} \: .
\]
Since \(x_d=z_d+\mathcal O(1)\), Stirling's formula gives
\[
    \frac{x_d^{z_d-1}e^{-x_d}}{\Gamma(z_d)} = \frac{1+\mathcal O(d^{-1})}{\sqrt{2\pi z_d}} \: .
\]
Consequently,
\[
    \alpha_{I_d,\eta(d)} = \frac{1+\mathcal O(d^{-1})}{2\pi\sqrt{z_d}} = \frac{1+o(1)}{\sqrt{2}\,\pi\sqrt d} \: .
\]

\paragraph{Asymptotic Behavior of the Lower-bound Constant.}
By Lemma~\ref{lem:gaussian_hellinger_control}, for \(d\ge3\),
\[
    \beta_{I_d,\eta(d)}
    =
    \min\left\{
        \frac{2\kappa_{d-1}m_d^{(d-1)/2}}{(2\pi)^{d/2}},
        \;
        2\sqrt{\frac{2}{\pi}}\,M_{d-1}
    \right\} \: ,
\]
where
\[
    M_{d-1}
    =
    \frac{(d-3)^{(d-3)/2}e^{-(d-3)/2}}
         {2^{(d-1)/2}\Gamma\!\left(\frac{d-1}{2}\right)}
\]
is the maximum of the density of a \(\chi^2(d-1)\) random variable.
The first term in the minimum can be related directly to the upper-bound constant:
\[
    \frac{2\kappa_{d-1}m_d^{(d-1)/2}}{(2\pi)^{d/2}}
    =
    \frac{2e^{m_d/2}}{s_d}\,
    \alpha_{I_d,\eta(d)}.
\]
Using the above and \(m_d=d+\mathcal O(1)\), this term grows exponentially in \(d\). 
It therefore cannot determine the minimum for all sufficiently large \(d\).

To analyze the second term, let \(w_d=(d-1)/2\). Then,
\[
    M_{d-1} = \frac{(w_d-1)^{w_d-1}e^{-(w_d-1)}}{2\Gamma(w_d)} \: .
\]
Another application of Stirling's formula gives
\[
    M_{d-1}
    =
    \frac{1+o(1)}{2\sqrt{2\pi w_d}}
    =
    \frac{1+o(1)}{2\sqrt{\pi d}}.
\]
Hence, for all sufficiently large \(d\),
\[
    \beta_{I_d,\eta(d)}
    =
    2\sqrt{\frac{2}{\pi}}\,M_{d-1}
    =
    \frac{\sqrt{2}+o(1)}{\pi\sqrt d} \: .
\]
Combining both results, we obtain
\[
    \frac{\beta_{I_d,\eta(d)}}{\alpha_{I_d,\eta(d)}} \to_{d\to +\infty}  2 \: .
\]
Therefore, the ratio between the upper and lower bounds (Theorems~\ref{thm:specialized_classifier_to_parameter_identifiability} and~\ref{thm:lower_bound}) is \(\widetilde{\mathcal O}(\sqrt d)\).

\paragraph{Different Statistical Objects.}
These two bounds concern different statistical objects. 
The upper bound controls the most distant point of \(\cC_n^+\) and consequently applies to every estimator selected from this feasible set. 
Its \(d\)-dependence arises from uniformly controlling the version space, whose associated classifier class has VC dimension of order \(d\).
By contrast, the lower bound concerns the risk of the best possible estimator and uses an Assouad construction that aggregates coordinatewise perturbations into a \(\sqrt d\) estimation loss.

The present results therefore do not determine whether the gap is due to a loose minimax lower bound, to the uniform control of the entire feasible set, or to a genuine suboptimality of arbitrary feasible estimators. 
In particular, a carefully selected estimator, such as the projected-likelihood $\dpe$ on $\cC_{n}$ or $\cC_{n}^{+}$, may have better dimensional dependence even when the diameter of \(\cC_n^+\) remains large. 
This interpretation is consistent with the experiments, where projected-likelihood estimators substantially outperform arbitrary feasible estimators. 
Closing the gap likely requires either an estimator-specific upper-bound analysis or a lower bound tailored to the diameter of the feasible set.


\section{Proofs of Section~\ref{sec:estimator4qualities}}
\label{app:qualities_estimators}

In this appendix, we prove the results presented in Section~\ref{sec:estimator4qualities}: Theorem~\ref{thm:specialized_classifier_to_parameter_identifiability} in Appendix~\ref{sub:proof_of_theorem_ref_thm_classifier_to_parameter_identifiability}, Theorem~\ref{thm:specialized_classifier_to_parameter_ambiguity} in Appendix~\ref{sub:proof_of_theorem_ref_thm_classifier_to_parameter_ambiguity}, Theorem~\ref{thm:classifier_to_parameter_identifiability} in Appendix~\ref{sub:proof_of_theorem_ref_thm_specialized_classifier_to_parameter_identifiability}, and Theorem~\ref{thm:classifier_to_parameter_ambiguity} in Appendix~\ref{sub:proof_of_theorem_ref_thm_specialized_classifier_to_parameter_ambiguity}.
Appendix~\ref{sub:local_error_to_parameter_conversion} contains the proof of the local error-to-parameter conversion for thresholded log-likelihood rewards: Assumption~\ref{ass:conversion_error_to_parameter_identifiability} holds for $\cF_{\cN, \Sigma}$ (Lemma~\ref{lem:gaussian_conversion_error_to_parameter_identifiability}) and $\cF_{\lap, b}$ (Lemma~\ref{lem:laplace_conversion_error_to_parameter_identifiability}), and Assumption~\ref{ass:conversion_error_to_parameter_ambiguity} holds for $\cF_{\Exp}$ (Lemma~\ref{lem:exponential_conversion_error_to_parameter_ambiguity}). 
Appendix~\ref{app:sub_exp_technical_lemmas} contains useful technical results for thresholded log-likelihood rewards with $\cF_{\Exp}$.
Appendix~\ref{sub:pointwise_qualities_versus_pairwise_preferences} contains several examples where preference feedback is more informative than quality feedback.

\subsection{Proof of Theorem~\ref{thm:classifier_to_parameter_identifiability}}
\label{sub:proof_of_theorem_ref_thm_classifier_to_parameter_identifiability}

Let $\delta \in (0,1)$.
Let $\theta^\star \in \Theta$ such that Assumptions~\ref{ass:finite_VC_identifiability} and~\ref{ass:conversion_error_to_parameter_identifiability} hold true for $d^{+}_{\star} \eqdef \mathrm{vc}(\cH^{+}_{\star})$ and $\alpha^{+}_{\star}, \varepsilon^{+}_{\star} > 0$.
Using these assumptions and notation from Section~\ref{sec:estimator4qualities},~\citet[Theorem 11]{2946645.3007088} yields, with probability at least $1-\delta$,
\begin{align*}
    \alpha^{+}_{\star} \min\left\{ \sup_{\theta \in \cC_{n}^{+}} \left\|\theta-\theta^\star\right\|, \varepsilon^{+}_{\star} \right\} \le \sup_{\theta \in \cC_{n}^{+}} p_{\theta^\star}(\cD^{+}(\theta^\star,\theta)) \le \frac{8}{n}\left( d^{+}_{\star} \ln \left( \frac{49 e n}{d^{+}_{\star}} + 37\right) + 8 \ln \left(\frac{6}{\delta}\right)\right) \: .
\end{align*}
Let us define
\[
    n^{+}_{\star}(\delta) \eqdef 1+\left\lceil \max\left\{ d_\star^+,\, \frac{128}{\varepsilon_\star^+ \alpha^{+}_{\star}}\ln\frac{6}{\delta},\, \frac{32d_\star^+}{\varepsilon_\star^+ \alpha^{+}_{\star}} \ln\left( e+\frac{32(49e+37)}{\varepsilon_\star^+ \alpha^{+}_{\star}} \right) \right\} \right\rceil  =_{\delta \to 0}  \cO\left(\frac{\log(1/\delta)}{\epsilon^{+}_{\star} \alpha^{+}_{\star}}\right) \: .
\]
For all $n \ge n^{+}_{\star}(\delta)$, we have
\[
    \alpha^{+}_{\star}  \varepsilon^{+}_{\star} >  \frac{8}{n}\left( d^{+}_{\star} \ln \left( \frac{49 e n}{d^{+}_{\star}} + 37\right) + 8 \ln \left(\frac{6}{\delta}\right)\right) \: .
\]
Therefore, for all $n \ge n^{+}_{\star}(\delta)$, we have, with probability at least $1-\delta$,
\begin{align*}
     \sup_{\theta \in \cC_{n}^{+}} \left\|\theta-\theta^\star\right\| \le \frac{8}{\alpha^{+}_{\star} n}\left( d^{+}_{\star} \ln \left( \frac{49 e n}{d^{+}_{\star}} + 37\right) + 8 \ln \left(\frac{6}{\delta}\right)\right) \: .
\end{align*}
This concludes the proof.

\subsection{Proof of Theorem~\ref{thm:classifier_to_parameter_ambiguity}}
\label{sub:proof_of_theorem_ref_thm_classifier_to_parameter_ambiguity}

Let $\delta \in (0,1)$.
Let $\theta^\star \in \Theta$ such that Assumptions~\ref{ass:finite_VC_ambiguity},~\ref{ass:conversion_error_to_parameter_ambiguity} and~\ref{ass:uniform_bernstein_bound} hold true for $d_{\star} \eqdef \mathrm{vc}(\cH_{\star})$ and $\alpha_{\star}, \varepsilon_{\star} > 0$.
Recall that $\mathrm{dist}(u,\cV) \eqdef \inf_{v \in \cV}\|u - v\|$.
Using the notation from Section~\ref{sec:estimator4qualities} and similar to Section~\ref{sub:classifier_to_parameter_identifiability}, the setting is realizable as $h_{\theta^\star} \in \cH _{\star}$, $h_{\theta^\star}  \equiv 1$ and
\begin{align*}
    &\cC_n = \{\theta \mid \forall i \in [n], \: h_{\theta}(X_i) = h_{\theta^\star}(X_i) \} \: , \\
    &p_{\theta^\star}(\cD(\theta^\star,\theta)) = p_{\theta^\star}( \{x \mid h_{\theta}(x) \ne h_{\theta^\star}(x)\} ) \: .
\end{align*}
Using these assumptions,~\citet[Theorem 11]{2946645.3007088} yields, with probability at least $1-\delta/2$,
\begin{align*}
    \alpha_{\star} \min\left\{ \sup_{\theta \in \cC_{n}} \mathrm{dist}(\theta,\cC(\theta^\star)), \varepsilon_{\star} \right\} \le \sup_{\theta \in \cC_{n}} p_{\theta^\star}(\cD(\theta^\star,\theta)) \le \frac{8}{n}\left( d_{\star} \ln \left( \frac{49 e n}{d_{\star}} + 37\right) + 8 \ln \left(\frac{12}{\delta}\right)\right) \: .
\end{align*}

Let $\cC^{-}(\theta^\star) \eqdef \cC(\theta^\star)\setminus\{\theta^\star\}$.
Since $\cC(\theta^\star)$ is finite, choose $\rho_\star>0$ such that, for all $\lambda\in\cC(\theta^\star)$, the balls $B(\lambda,\rho_\star)$ for the norm $\|\cdot\|$ are pairwise disjoint and their union is contained in a compact subset
of $\Theta$. 
Define $K_\star^{-} \eqdef \bigcup_{\lambda\in\cC^{-}(\theta^\star)} B(\lambda,\rho_\star)$.
By continuity of $\theta\mapsto\KL(p_{\theta^\star},p_\theta)$ and uniqueness of its minimizer (see, the standard regularity conditions from Section~\ref{sec:problem_statement}), we have
\[
    \gamma_\star \eqdef \inf_{\theta\in K_\star^{-}} \KL(p_{\theta^\star},p_\theta) > 0 \: .
\]
Let $C_\star\eqdef C_{K_\star^{-}}$ and $D_\star\eqdef D_{K_\star^{-}}$, as given by Assumption~\ref{ass:uniform_bernstein_bound}.
Then, with probability at least $1-\delta/2$,
\[
    \sup_{\theta\in K_\star^{-}} \left| (L_n^{\so}(\theta)-L_n^{\so}(\theta^\star))/n - \KL(p_{\theta^\star},p_\theta)\right| \le C_\star\sqrt{\frac{\log(4/\delta)}{n}} + D_\star\frac{\log(4/\delta)}{n}  \: .
\]
Let us define $q_\star \eqdef \alpha_\star\min\{\varepsilon_\star,\rho_\star\}$ and $n_\star(\delta)\eqdef \max\left\{ n_{\star,1}(\delta), n_{\star,2}(\delta) \right\}$ where
\begin{align*}
    &n_{\star,1}(\delta) \eqdef 1+\Bigg\lceil \max\Bigg\{ d_\star,\, \frac{128}{q_\star}\ln\frac{12}{\delta},\frac{32d_\star}{q_\star} \ln\left(e+\frac{32(49e+37)}{q_\star}\right)\Bigg\} \Bigg\rceil \quad \text{and} \\
    &n_{\star,2}(\delta) \eqdef 1+\left\lceil \max\left\{ \frac{4C_\star^2}{\gamma_\star^2}\ln\frac{4}{\delta}, \frac{2D_\star}{\gamma_\star}\ln\frac{4}{\delta} \right\} \right\rceil \: .
\end{align*}
We have $n_\star(\delta) =_{\delta \to 0} \cO(\log(1/\delta))$ where the constants are hidden for simplicity of exposition. 
For all $n \ge n_{\star}(\delta)$, 
\begin{align*}
    &\alpha_{\star}  \varepsilon_{\star} >  \frac{8}{n}\left( d_{\star} \ln \left( \frac{49 e n}{d_{\star}} + 37\right) + 8 \ln \left(\frac{12}{\delta}\right)\right)\quad \text{and} \quad \gamma_{\star} > C_\star\sqrt{\frac{\log(4/\delta)}{n}} + D_\star\frac{\log(4/\delta)}{n} \: .
\end{align*}
A direct union bound over the two events above (each holding with probability at least $1-\delta/2$), yields, with probability at least $1-\delta$, 
\begin{align*}
    &\mathrm{dist}(\widehat\theta_n^{\dpe},\cC(\theta^\star)) \le \frac{8}{n \alpha_\star}\left( d_{\star} \ln \left( \frac{49 e n}{d_{\star}} + 37\right) + 8 \ln \left(\frac{12}{\delta}\right)\right) \quad \text{and} \quad \inf_{\theta\in K_\star^{-}} \left\{ L_n^{\so}(\theta)-L_n^{\so}(\theta^\star) \right\} > 0  \: .
\end{align*}
Since $\theta^\star\in\cC_n$ and $\widehat\theta_n^{\dpe}\in\argmin_{\theta\in\cC_n}L_n^{\so}(\theta)$, we necessarily have 
\[
    \widehat\theta_n^{\dpe} \notin K_\star^{-} \quad \text{and}\quad \mathrm{dist}(\widehat\theta_n^{\dpe},\cC(\theta^\star)) = \|\widehat\theta_n^{\dpe}- \theta^\star\|\: .
\]
Therefore, for all $n \ge n_{\star}(\delta)$, we have, with probability at least $1-\delta$,
\begin{align*}
     \|\widehat\theta_n^{\dpe}- \theta^\star\| \le \frac{8}{n \alpha_\star}\left( d_{\star} \ln \left( \frac{49 e n}{d_{\star}} + 37\right) + 8 \ln \left(\frac{12}{\delta}\right)\right) \: .
\end{align*}
This concludes the proof.

\subsection{Proof of Theorem~\ref{thm:specialized_classifier_to_parameter_identifiability}}
\label{sub:proof_of_theorem_ref_thm_specialized_classifier_to_parameter_identifiability}

Let $\delta \in (0,1)$, $\eta_0 = \frac{1}{2}\log((2\pi)^d|\Sigma|)$ for $\cF_{\cN, \Sigma}$ and $\eta_0 = \log(2b)$ for $\cF_{\lap, b}$ with $d=1$.
Recall that $\|x\|_{\Sigma} \eqdef \sqrt{x\transpose \Sigma x}$.
Let $\eta \in (\eta_0,+ \infty) $ be a known threshold, and define $\bar \eta \eqdef \eta - \eta_0$.
For thresholded log-likelihood rewards, we have
\[
    r_{\theta}(x) \eqdef \bar \eta - \begin{cases}
        \frac{1}{b}|x - \theta| &\text{for } \cF_{\lap,b}\\
        \frac{1}{2} \| \Sigma^{-1}x - \theta\|_{\Sigma}^2 &\text{for } \cF_{\cN,\Sigma}
    \end{cases} \quad \text{,}\quad c_{\eta} \eqdef \begin{cases}
        b\bar\eta &\text{for } \cF_{\lap,b}\\
        \sqrt{2\bar \eta} &\text{for } \cF_{\cN,\Sigma}
    \end{cases}\quad \text{and}\quad r_{\eta} \eqdef \begin{cases}
        b\bar\eta &\text{for } \cF_{\lap,b}\\
        \sigma\sqrt{2\bar \eta} &\text{for } \cF_{\cN,\sigma^2}
    \end{cases} \: .
\]
To keep the same argument for both families, we define
\[
m_\theta:=
\begin{cases}
\theta,&\mathcal F_{\mathrm{Lap},b},\\
\sigma^2\theta,&\mathcal F_{\mathcal N,\sigma^2}.
\end{cases}
\]

\paragraph{Case $\cF_{\lap,b}$ and $\cF_{\cN,\sigma^2}$.}
For all $x_{1:n} \in \cX^{n}$, let us define
\begin{align*}
    & C_{n}^{+}(x_{1:n}) \eqdef \{ \theta \mid \forall i \in [n] \text{ s.t. }  r_{\theta^\star}(x_i) > 0,\: r_{\theta}(x_i) \ge 0  \}  \quad \text{,} \quad \psi_{n}^{+}(x_{1:n}) \eqdef \bigcup_{\theta \in C_{n}^{+}(x_{1:n})} \cD^{+}(\theta^\star,\theta)\:,\\
    &x_{n,\min}^{+} \eqdef \min \{ x_i \mid r_{\theta^\star}(x_i) > 0 \} \quad \text{and}\quad x_{n,\max}^{+} \eqdef \max \{ x_i \mid r_{\theta^\star}(x_i) > 0 \} \: ,
\end{align*}
with the usual conventions $+ \infty$ and $- \infty$ if $\{ i \mid r_{\theta^\star}(x_i) > 0 \} = \emptyset$.
Up to endpoints of $p_{\theta^\star}$-measure zero, we have
\[
    \psi_{n}^{+}(x_{1:n}) = \left((m_{\theta^\star} - r_{\eta}, x_{n,\min}^{+}) \cup ( x_{n,\max}^{+}, m_{\theta^\star} + r_{\eta})\right) \cap (m_{\theta^\star} - r_{\eta},m_{\theta^\star} + r_{\eta}) \: .
\]
Therefore, this defines a consistent monotone sample compression rules of size at most $2$.
Since $\cC_{n}^{+} \eqdef C_{n}^{+}(X_{1:n})$,~\citep[Theorem 3]{2946645.3007088} yields, with probability at least $1-\delta$,
\begin{align*}
    \sup_{\theta \in \cC_{n}^{+}} p_{\theta^\star}(\cD^{+}(\theta^\star,\theta)) \le p_{\theta^\star}\left(\psi_n^{+}(X_{1:n})\right) \le \frac{1}{n}\left( 42 + 16 \ln \left(\frac{3}{\delta}\right)\right) \: .
\end{align*}
Let $\epsilon_{\cF,\eta}, \alpha_{\cF,\eta} >0$ as in Eq.~\eqref{eq:gaussian_conversion_error_to_parameter_identifiability} of Lemma~\ref{lem:gaussian_conversion_error_to_parameter_identifiability} for $\cF_{\cN,\sigma^2}$ and as in Eq.~\eqref{eq:laplace_conversion_error_to_parameter_identifiability} of Lemma~\ref{lem:laplace_conversion_error_to_parameter_identifiability} for $\cF_{\lap,b}$.
Since
    \[
       \forall \theta \in \R, \quad p_{\theta^\star}\left(\cD^+(\theta^\star,\theta)\right) \geq  \alpha_{\cF,\eta}   \min\left\{ |\theta-\theta^\star|, \epsilon_{\cF,\eta} \right\} \: ,
    \]
we obtain that, with probability at least $1-\delta$, 
\begin{align*}
     \alpha_{\cF,\eta}   \min\left\{\sup_{\theta \in \cC_{n}^{+}} |\theta-\theta^\star|, \epsilon_{\cF,\eta} \right\} \le \frac{1}{n}\left( 42 + 16 \ln \left(\frac{3}{\delta}\right)\right) \: .
\end{align*}
Let us define
\[
    n_{\cF,\eta}(\delta) \eqdef 1 + \left\lceil \frac{42 + 16 \ln \left(\frac{3}{\delta}\right)}{\alpha_{\cF,\eta}  \epsilon_{\cF,\eta}  } \right\rceil =_{\delta \to 0} \cO\left(\frac{\log(1/\delta)}{\alpha_{\cF,\eta}  \epsilon_{\cF,\eta} }\right) \: .
\]
Therefore, for all $n \ge n_{\cF,\eta}(\delta)$, we have, with probability at least $1-\delta$,
\begin{align*}
     \sup_{\theta \in \cC_{n}^{+}}|\theta - \theta^\star| \le \frac{42 + 16 \ln \left(\frac{3}{\delta}\right)}{n \alpha_{\cF,\eta}} \: .
\end{align*}
This concludes the proof of the first result on $\cF_{\lap,b}$ and $\cF_{\cN,\sigma^2}$.

\paragraph{Case $\cF_{\cN,\Sigma}$.}
We have $\bar \eta > 0$, $c_{\eta} = \sqrt{2\bar\eta}$ and $\Sigma$ is positive definite.
Let $E_\theta \eqdef \{x \in \R^d \mid  \| \Sigma^{-1}x - \theta\|_{\Sigma} < c_\eta\}$ for all $\theta\in \R^d$.
Recall that $\overline{X}$ denotes the closure of the set $X$.
Using the notation from Section~\ref{sub:classifier_to_parameter_identifiability}, we have $\cH^{+}_{\star}\eqdef \{h_{\theta}^{+} \mid \theta \in \R^d\}$ where
\[
    \forall x \in \R^d, \: h_{\theta}^{+}(x) \eqdef \begin{cases}
    -1 & \text{if }  x \in E_{\theta^\star}  \setminus \overline{E_\theta}  \\
    +1 &\text{otherwise}
    \end{cases} \: .
\]
Based on an invertible mapping, \(\{E_\theta:\theta\in\mathbb R^d\}\) is a class of Euclidean balls with a fixed radius. 
It is therefore a subclass of the class of all Euclidean balls in \(\mathbb R^d\), whose VC dimension is \(d+1\).
Taking complements and intersecting every set with the same fixed set \(E_{\theta^\star}\) cannot increase the VC dimension. 
Assumption~\ref{ass:finite_VC_identifiability} holds true with explicit constant $d+1$ since
\[
    d_\star^+ = \operatorname{vc}(\mathcal H_\star^+) \le d+1 < \infty \: .
\]

Using Lemma~\ref{lem:gaussian_conversion_error_to_parameter_identifiability}, Assumption~\ref{ass:conversion_error_to_parameter_identifiability} also holds true with explicit constants $\alpha_{\Sigma,\eta}$ and $\epsilon_{\Sigma,\eta}$ as in Eq.~\eqref{eq:gaussian_conversion_error_to_parameter_identifiability} for the norm $\|\cdot\|_{\Sigma}$.

Both Assumptions~\ref{ass:finite_VC_identifiability} and~\ref{ass:conversion_error_to_parameter_identifiability} are shown with constants that are independent of $\theta^\star$.
Therefore, the second result on $\cF_{\cN,\Sigma}$ is a corollary of Theorem~\ref{thm:classifier_to_parameter_identifiability}, applied uniformly over $\theta^\star$.

\subsection{Proof of Theorem~\ref{thm:specialized_classifier_to_parameter_ambiguity}}
\label{sub:proof_of_theorem_ref_thm_specialized_classifier_to_parameter_ambiguity}

We use the technical results for \(\cF_{\Exp}\) established in Appendix~\ref{app:sub_exp_technical_lemmas}, together with the notation introduced therein. In particular, let $\theta_0 \eqdef e^{-\eta}$, $\wt\theta_0 \eqdef e^{1-\eta}$, $x_0 \eqdef e^{\eta-1}$ and the function $f_\eta(y) \eqdef \frac{\eta+\log y}{y}$ for all $y  > 0$.
Let $\theta^\star \in (\theta_0,+\infty)\setminus \{\wt \theta_0\}$ and write $x^\star_\eta\eqdef f_\eta(\theta^\star)\in(0,x_0]$.

For all $x_{1:n} \in \cX^{n}$, let us define
\begin{align*}
    &C_{n}(x_{1:n}) \eqdef \{ \theta \mid \forall i \in [n] \text{ s.t. }  r_{\theta^\star}(x_i) r_{\theta}(x_i)  \ge 0 \}   \quad \text{,} \quad \psi_{n}(x_{1:n}) \eqdef \bigcup_{\theta \in C_{n}(x_{1:n})} \cD(\theta^\star,\theta) \:,\\
    &x_{n,\max}^{+} \eqdef \max\left( \{x_i:r_{\theta^\star}(x_i)>0\} \cup\{0\} \right) \quad \text{and}\quad x_{n,\min}^{-} \eqdef  \min\left( \{x_i:r_{\theta^\star}(x_i)<0\} \cup\{x_0\} \right) \: .
\end{align*}
By Lemma~\ref{lem:init_properties}, $r_\theta(x)>0$ if and only if $x<f_\eta(\theta)$ and \(f_\eta(\theta)\in(0,x_0]\) for every \(\theta>\theta_0\).
Up to endpoints,
\[
    \theta\in C_n(x_{1:n}) \quad\iff\quad f_\eta(\theta) \in  \left[ x_{n,\max}^{+}, x_{n,\min}^{-} \right] \: .
\]
Moreover, Lemma~\ref{lem:inverse_branches_f_eta} shows that every threshold \(x\in(0,x_0)\) is attained by some \(\theta>\theta_0\); indeed, it is attained by the two inverse branches \(\theta_{\eta,0}(x)\) and \(\theta_{\eta,-1}(x)\).
Since $\cD(\theta^\star,\theta) = \left( \min\{x^\star_\eta,f_\eta(\theta)\}, \max\{x^\star_\eta,f_\eta(\theta)\} \right)$, we have, up to endpoints of $p_{\theta^\star}$-measure zero
\[
    \psi_n(x_{1:n}) = (x_{n,\max}^{+}, x_\eta^\star ) \cup (x_\eta^\star, x_{n,\min}^{-} ) \: .
\]
Therefore, this defines a consistent monotone sample compression rules of size at most $2$.
Since $\cC_n \eqdef C_n(X_{1:n})$,~\citep[Theorem 3]{2946645.3007088} yields, with probability at least $1-\delta/2$,
\begin{align*}
    \sup_{\theta \in \cC_{n}} p_{\theta^\star}(\cD(\theta^\star,\theta)) \le p_{\theta^\star}\left(\psi_n(X_{1:n})\right) \le \frac{1}{n}\left( 42 + 16 \ln \left(\frac{6}{\delta}\right)\right) \: .
\end{align*}
Using Lemma~\ref{lem:exponential_conversion_error_to_parameter_ambiguity}, Assumption~\ref{ass:conversion_error_to_parameter_ambiguity} holds true with explicit constants $\alpha_{\star,\eta},\epsilon_{\star,\eta}> 0$ as in Eq.~\eqref{eq:exponential_conversion_error_to_parameter_ambiguity} for the norm $|\cdot|$.

Let $K \subset (0,+\infty)$ be a compact. 
Assumption~\ref{ass:uniform_bernstein_bound} also holds for $\cF_{\Exp}$ with constants
\begin{equation} \label{eq:exp_bernstein_cst}
    C_K \eqdef \frac{\sqrt{2}}{\theta^\star}  \sup_{\theta\in K}|\theta-\theta^\star| \quad \text{and} \quad D_K \eqdef \frac{1}{\theta^\star} \sup_{\theta\in K}|\theta-\theta^\star| \: ,
\end{equation}
because
\[
    \sup_{\theta\in K}
    \left|
        \frac{L_n^{\so}(\theta)-L_n^{\so}(\theta^\star)}{n}
        -
        \KL(p_{\theta^\star},p_\theta)
    \right|
    =
    \sup_{\theta\in K}|\theta-\theta^\star|
    \left|
        \overline X_n-\frac{1}{\theta^\star}
    \right|,
\]
and the centered exponential variable \(X-1/\theta^\star\) is sub-gamma with variance factor \(1/(\theta^\star)^2\) and scale parameter \(1/\theta^\star\).

Combining these results, we can conclude using the same proof as in Appendix~\ref{sub:proof_of_theorem_ref_thm_classifier_to_parameter_ambiguity}.
Let $q_0 ,q_{-1}$ as in Lemma~\ref{lem:exponential_conversion_error_to_parameter_ambiguity}.
Then, as in in Appendix~\ref{sub:proof_of_theorem_ref_thm_classifier_to_parameter_ambiguity}, we can choose $\rho^\star = |q_0 - q_{-1}|/4$ to define $K^{-}_{\star}$ and $\gamma^\star$, then take the associated $C_\star$ and $D_\star$. 
Note that the explicit definition of $n_{\star}(\delta)$ should be modified to account for the improved PAC bound consistent monotone sample compression rules of size at most $2$.
As our statement hides explicit positive constants with the notation $n_{\star}(\delta) =_{\delta\to 0} \cO(\log(1/\delta))$, we do not need to define it explicitly.

\subsection{Local Error-to-parameter Conversion}
\label{sub:local_error_to_parameter_conversion}

In this appendix, we prove the local error-to-parameter conversion for thresholded log-likelihood rewards.

Assumption~\ref{ass:conversion_error_to_parameter_identifiability} holds for $\cF_{\cN, \Sigma}$ (Lemma~\ref{lem:gaussian_conversion_error_to_parameter_identifiability})
\begin{lemma}\label{lem:gaussian_conversion_error_to_parameter_identifiability}
    Let $\eta > \frac{1}{2}\log\left((2\pi)^d|\Sigma|\right)$ and define
    \begin{align} \label{eq:gaussian_conversion_error_to_parameter_identifiability}
       & \alpha_{\Sigma,\eta} \eqdef \begin{cases}
        \sigma^2 e^{-\eta} & \text{if } d=1 \\
        \kappa_{d-1} \beta_d |\Sigma|^{1/2} e^{-\eta} \left( 2\eta-\log\left((2\pi)^d|\Sigma|\right)\right)^{(d-1)/2} & \text{otherwise} 
        \end{cases}> 0 \\
        & \text{and}\quad \epsilon_{\Sigma,\eta}\eqdef \begin{cases}
        \frac{2}{\sigma}\sqrt{2\eta-\log (2\pi \sigma^2))} & \text{if } d=1 \\
        \frac{1}{2d}\sqrt{2\eta - \log((2\pi)^d|\Sigma|)} & \text{otherwise} 
        \end{cases}>0  \: , \nonumber
    \end{align}
    where $\kappa_{d-1} \eqdef \frac{\pi^{(d-1)/2}} {\Gamma\left(\frac{d+1}{2}\right)}$ and $\beta_d \eqdef \left(1-\frac{1}{16d^2}\right)^{(d-1)/2}$.
    Then, for every $(\theta^\star,\theta)\in\R^d \times \R^d$,
    \begin{align*}
        &p_{\theta^\star}\left(\cD^+(\theta^\star,\theta)\right) \geq \begin{cases}
        \alpha_{\Sigma,\eta}   \min\left\{  |\theta-\theta^\star |, \epsilon_{\Sigma,\eta} \right\} & \text{if } d=1 \\
         \alpha_{\Sigma,\eta}   \min\left\{ \left\|\theta-\theta^\star\right\|_{\Sigma}, \epsilon_{\Sigma,\eta} \right\} & \text{otherwise} 
        \end{cases}  \:  .
    \end{align*}
\end{lemma}
\begin{proof}
Let $c_\eta\eqdef \sqrt{2\bar\eta}$ and $E_\theta \eqdef \{x \in \R^d \mid  \| \Sigma^{-1}x - \theta\|_{\Sigma} < c_\eta\}$. Then, $\cD^+(\theta^\star,\theta) =E_{\theta^\star}\setminus\overline{E_\theta}$.

\paragraph{Case \(d=1\).}
Let $Y\eqdef X/\sigma^2$ and define the disagreement region in the transformed coordinates by
\[
    \cD_Y^+(\theta^\star,\theta)
    \eqdef \{y\in\R:\sigma^2 y\in\cD^+(\theta^\star,\theta)\} \: .
\]
Set $r_\eta\eqdef c_\eta/\sigma$ and $\Delta\eqdef\theta-\theta^\star$.
In these coordinates, the two relevant intervals have radius $r_\eta$.
If $0<\Delta<2r_\eta$, then
$\cD_Y^+(\theta^\star,\theta)=(\theta^\star-r_\eta,\theta-r_\eta)$,
whereas, if $-2r_\eta<\Delta<0$, then
$\cD_Y^+(\theta^\star,\theta)=(\theta+r_\eta,\theta^\star+r_\eta)$.
Finally, if $|\Delta|\ge 2r_\eta$, the two intervals are disjoint and
$\cD_Y^+(\theta^\star,\theta)=(\theta^\star-r_\eta,\theta^\star+r_\eta)$.

Under $p_{\theta^\star}$, $Y-\theta^\star\sim\cN(0,\sigma^{-2})$, whose density is
\[
    f_\Sigma(t)
    =\sqrt{\frac{\sigma^2}{2\pi}}
    \exp\left(-\frac{\sigma^2t^2}{2}\right)
    \quad\text{with}\quad
    \inf_{|t|\le r_\eta}f_\Sigma(t)
    =\alpha_{\Sigma,\eta}
    \eqdef\sqrt{\frac{\sigma^2}{2\pi}}
    \exp\left(-\frac{c_\eta^2}{2}\right) \: .
\]
Since
\[
    p_{\theta^\star}\bigl(\cD^+(\theta^\star,\theta)\bigr)
    =\bP\bigl(Y\in\cD_Y^+(\theta^\star,\theta)\bigr),
\]
we obtain, when $|\Delta|<2r_\eta$,
\[
    p_{\theta^\star}\bigl(\cD^+(\theta^\star,\theta)\bigr)
    \ge \alpha_{\Sigma,\eta}|\Delta| \: .
\]
When $|\Delta|\ge 2r_\eta$,
\begin{align*}
    p_{\theta^\star}\bigl(\cD^+(\theta^\star,\theta)\bigr)
    &=\int_{-r_\eta}^{r_\eta}f_\Sigma(t)\,\mathrm dt
    \ge 2r_\eta\alpha_{\Sigma,\eta} \: .
\end{align*}
This concludes the proof.

\paragraph{Case \(d>1\).}
Let $Z \eqdef \Sigma^{-1/2}X-\Sigma^{1/2}\theta^\star \sim\cN(0,I_d)$ and
\[
    v\eqdef\Sigma^{1/2}(\theta-\theta^\star), \qquad  \rho\eqdef\|v\|_2  =\|\theta-\theta^\star\|_\Sigma \: .
\]
After an orthogonal rotation, we may assume that \(v=\rho e_1\). Thus
\[
    p_{\theta^\star}\bigl(\cD^+(\theta^\star,\theta)\bigr) = \mathbb P\left( \|Z\|_2<c_\eta,\, \|Z-\rho e_1\|_2\ge c_\eta \right) \: .
\]
We first consider \(0<\rho\le c_\eta/(2d)\).
Let $\kappa_{d-1} \eqdef \frac{\pi^{(d-1)/2}} {\Gamma\left(\frac{d+1}{2}\right)}$ denote the volume of the unit ball in \(\R^{d-1}\).
Since the standard Gaussian density is bounded below on \(B(0,c_\eta)\) by $(2\pi)^{-d/2} e^{-c_\eta^2/2}$, we obtain
\begin{align*}
    p_{\theta^\star}\left(\cD^+(\theta^\star,\theta)\right)
    &\ge
    \frac{e^{-c_\eta^2/2}}{(2\pi)^{d/2}}
    \operatorname{Leb}\left(
        B(0,c_\eta)\setminus B(\rho e_1,c_\eta)
    \right) \: .
\end{align*}

Slicing the two balls orthogonally to \(e_1\) gives
\begin{align}
    \operatorname{Leb}\left(
        B(0,c_\eta)\setminus B(\rho e_1,c_\eta)
    \right)
    &=
    \kappa_{d-1}
    \int_{-\rho/2}^{\rho/2}
        \left(c_\eta^2-t^2\right)^{(d-1)/2}
    \,\mathrm dt.
    \label{eq:two_ball_difference_volume}
\end{align}
Indeed, the hyperplane \(\{x_1=\rho/2\}\) is the perpendicular bisector of the two centers, and the contributions outside the central slab cancel after translating one of the balls by \(\rho e_1\).

Let us define $\beta_d \eqdef \left(1-\frac{1}{16d^2}\right)^{(d-1)/2}$. Since \(|t|\le\rho/2\) throughout the integral,
\begin{align*}
    \operatorname{Leb}\left(
        B(0,c_\eta)\setminus B(\rho e_1,c_\eta)
    \right)
    &\ge
    \kappa_{d-1}\rho
    \left(c_\eta^2-\frac{\rho^2}{4}\right)^{(d-1)/2}\ge  \kappa_{d-1}c_\eta^{d-1} \beta_d \rho\: ,
\end{align*}
where we used that \(\rho\le c_\eta/(2d)\).
Thus, for $\rho \in (0,c_\eta/2)$, we have 
\begin{align*}
    p_{\theta^\star}\left(\cD^+(\theta^\star,\theta)\right) \ge  \kappa_{d-1} \beta_d \frac{c_\eta^{d-1}e^{-c_\eta^2/2}}{(2\pi)^{d/2}} \|\theta-\theta^\star\|_\Sigma   \: .
\end{align*}
For \(\rho\ge c_\eta/(2d)\), the map $\rho \mapsto \mathbb P\left( Z\in B(0,c_\eta)\setminus B(\rho e_1,c_\eta) \right)$ is nondecreasing. 
Indeed, its complement inside \(B(0,c_\eta)\) has probability
\[
    \int_{\R^d} \phi_d(z) \indi{z\in B(0,c_\eta)} \indi{z-\rho e_1\in B(0,c_\eta)} \,\mathrm dz,
\]
where \(\phi_d\) is the standard Gaussian density. 
This is the convolution of two nonnegative radially symmetric nonincreasing functions, and is therefore radially nonincreasing in \(\rho\).
Consequently, applying the above result at \(\rho=c_\eta/(2d)\) gives
\begin{align}
    p_{\theta^\star}\left(\cD^+(\theta^\star,\theta)\right) \ge \kappa_{d-1} \beta_d \frac{c_\eta^{d-1}e^{-c_\eta^2/2}}{(2\pi)^{d/2}}\frac{c_\eta}{2d}\: . 
\end{align}
This concludes the proof.
\end{proof}

Assumption~\ref{ass:conversion_error_to_parameter_identifiability} holds for $\cF_{\lap, b}$ (Lemma~\ref{lem:laplace_conversion_error_to_parameter_identifiability}).
\begin{lemma}\label{lem:laplace_conversion_error_to_parameter_identifiability}
    Let $\eta > \log(2b)$ and define
    \begin{equation} \label{eq:laplace_conversion_error_to_parameter_identifiability}
        \alpha_{b,\eta} \eqdef e^{-\eta} > 0 \quad \text{and}\quad \epsilon_{b,\eta}\eqdef 2b(\eta-\log(2b)) > 0\: .
    \end{equation}
    Then, for every $(\theta^\star,\theta)\in\R^2$,
    \[
        p_{\theta^\star}\left(\cD^+(\theta^\star,\theta)\right) \geq  \alpha_{b,\eta}   \min\left\{ |\theta-\theta^\star|, \epsilon_{b,\eta} \right\} \: .
    \]
\end{lemma}
\begin{proof}
Let $c_\eta\eqdef b(\eta-\log(2b))$ and $\Delta\eqdef|\theta-\theta^\star|$.
Then,
\[
    \cD^+(\theta^\star,\theta)= \{x\in\R\mid |x - \theta^\star|<c_\eta,\: |x - \theta|> c_\eta\}\: .
\]
By symmetry, it suffices to consider \(\theta\geq\theta^\star\).
If \(0\leq\Delta<2c_\eta\), then $\cD^+(\theta^\star,\theta)= (\theta^\star-c_\eta,\theta-c_\eta)$ and
\begin{align*}
    p_{\theta^\star}\left(
        \cD^+(\theta^\star,\theta)
    \right)
    &=
    \frac{1}{2b}
    \int_{\theta^\star-c_\eta}^{\theta-c_\eta}
        \exp\left(
            -\frac{|x-\theta^\star|}{b}
        \right)
    \,\mathrm dx\ge  e^{-\eta} \Delta .
\end{align*}
If \(\Delta\geq2c_\eta\), the two positive regions are disjoint, so $\cD^+(\theta^\star,\theta)= (\theta^\star-c_\eta,\theta^\star+c_\eta)$.
Similarly, we obtain
\begin{align*}
    p_{\theta^\star}\left( \cD^+(\theta^\star,\theta) \right) \ge 2c_\eta e^{-\eta}.
\end{align*}
This concludes the proof.
\end{proof}

Assumption~\ref{ass:conversion_error_to_parameter_ambiguity} holds for $\cF_{\Exp}$ (Lemma~\ref{lem:exponential_conversion_error_to_parameter_ambiguity}). 
\begin{lemma}\label{lem:exponential_conversion_error_to_parameter_ambiguity}
Let $\theta^\star \in (e^{-\eta},+\infty)\setminus\{e^{1-\eta}\}$ and $x_\eta^\star \eqdef \frac{\eta+\log\theta^\star}{\theta^\star}$.
Let $q_0 \eqdef \theta_{\eta,0}(x_\eta^\star)$ and $q_{-1} \eqdef \theta_{\eta,-1}(x_\eta^\star)$ where the inverse branches are defined in Lemma~\ref{lem:inverse_branches_f_eta}, and $\cC(\theta^\star)=\{q_0,q_{-1}\}$. 
Define
\begin{equation}  \label{eq:exponential_conversion_error_to_parameter_ambiguity}
    \varepsilon_{\star,\eta} \eqdef \frac{1}{2} \min\left\{  1-\frac{e^{-\eta}}{q_0}, \frac{e^{1-\eta}}{q_0}-1, 1-\frac{e^{1-\eta}}{q_{-1}} \right\} > 0 \quad \text{and}\quad \alpha_{\star,\eta} \eqdef  \theta^\star
    \exp\left(-\theta^\star e^{\eta-1}\right) m_{\star,\eta} \min\{1,q_0\} > 0\: ,
\end{equation}
where
\[
    m_{\star,\eta} \eqdef \min\Bigg\{ 
    \frac{ 1-\eta - \log\left( q_0(1+\varepsilon_{\star,\eta}) \right) }{ q_0^2(1+\varepsilon_{\star,\eta})^2 }, 
    \frac{ \log\left( q_{-1}(1-\varepsilon_{\star,\eta})  \right) -1 +\eta}{ q_{-1}^2(1-\varepsilon_{\star,\eta})^2 },
    \frac{ \log\left( q_{-1}(1+\varepsilon_{\star,\eta})  \right) -1 +\eta}{ q_{-1}^2(1+\varepsilon_{\star,\eta})^2 },
    \frac{x_\eta^\star}{\varepsilon_{\star,\eta}} \Bigg\} \: .
\]
Then, for every $\theta\in (0,+\infty)$,
\[
    p_{\theta^\star} \left( \mathcal D(\theta^\star,\theta) \right)\ge \alpha_{\star,\eta} \min\left\{ | \theta - q_0|, |\theta-q_{-1}|, \varepsilon_{\star,\eta} \right\} \: .
\]
\end{lemma}
\begin{proof}
By Lemma~\ref{lem:inverse_branches_f_eta},
\[
    e^{-\eta}<q_0<e^{1-\eta}<q_{-1}.
\]
Hence all three terms in the minimum defining
$\varepsilon_{\star,\eta}$ are strictly positive, and therefore
$\varepsilon_{\star,\eta}>0$. Moreover, its definition ensures that
\[
    q_0(1+\varepsilon_{\star,\eta})<e^{1-\eta}
    <q_{-1}(1-\varepsilon_{\star,\eta}).
\]
Consequently, all three derivative terms in the definition of
$m_{\star,\eta}$ are strictly positive. Since
$x_\eta^\star=f_\eta(\theta^\star)>0$, the last term is also strictly
positive, and thus $m_{\star,\eta}>0$. Finally,
$\theta^\star>0$ and $\exp(-\theta^\star e^{\eta-1})>0$, which yields
$\alpha_{\star,\eta}>0$.

For brevity, write
\[
    \varepsilon\eqdef\varepsilon_{\star,\eta},
    \qquad
    m\eqdef m_{\star,\eta},
    \qquad
    c\eqdef\min\{1,q_0\},
    \qquad
    \rho\eqdef c\varepsilon.
\]
We prove that, for every $\theta>0$,
\[
    p_{\theta^\star}\bigl(\mathcal D(\theta^\star,\theta)\bigr)
    \ge
    \theta^\star e^{-\theta^\star x_0}
    cm
    \min\bigl\{
        |\theta-q_0|,
        |\theta-q_{-1}|,
        \varepsilon
    \bigr\}.
\]

Since $x_\eta^\star=f_\eta(\theta^\star)$ and
$\cC(\theta^\star)=\{q_0,q_{-1}\}$, the reward associated
with $\theta^\star$ changes sign at $x_\eta^\star$.

If $\theta\leq\theta_0$, then $f_\eta(\theta)\leq0$, so
$r_\theta(x)<0$ for every $x\geq0$, whereas
$r_{\theta^\star}(x)>0$ for $x\in[0,x_\eta^\star)$. Hence
\[
    [0,x_\eta^\star)
    \subseteq
    \mathcal D(\theta^\star,\theta).
\]

If $\theta>\theta_0$, both rewards change sign at,
respectively, $x_\eta^\star$ and $f_\eta(\theta)$. Therefore,
up to its endpoints,
\[
    \mathcal D(\theta^\star,\theta)
    \supseteq
    \bigl[
        \min\{f_\eta(\theta),x_\eta^\star\},
        \max\{f_\eta(\theta),x_\eta^\star\}
    \bigr]
\]
whenever $f_\eta(\theta)>0$. If $f_\eta(\theta)\leq0$, the
disagreement set again contains $[0,x_\eta^\star)$.

All these intervals are contained in $[0,x_0]$. Since
$p_{\theta^\star}(x)=\theta^\star e^{-\theta^\star x}$,
we consequently obtain
\begin{equation}
\label{eq:disagreement_lower_f}
    p_{\theta^\star}\bigl(\mathcal D(\theta^\star,\theta)\bigr)
    \ge
    \theta^\star e^{-\theta^\star x_0}
    \begin{cases}
        |f_\eta(\theta)-x_\eta^\star|,
            & \theta>\theta_0,\\
        x_\eta^\star,
            & \theta\leq\theta_0.
    \end{cases}
\end{equation}

It remains to lower-bound the right-hand side in terms of the
distance to $\{q_0,q_{-1}\}$. By the definition of
$\varepsilon$, the intervals
\[
    [q_0(1-\varepsilon),q_0(1+\varepsilon)]
    \quad\text{and}\quad
    [q_{-1}(1-\varepsilon),q_{-1}(1+\varepsilon)]
\]
lie, respectively, inside the increasing and decreasing
branches of $f_\eta$. Moreover, because
$\rho\leq q_0\varepsilon$ and
$\rho\leq q_{-1}\varepsilon$, they contain the absolute
$\rho$-neighbourhoods of $q_0$ and $q_{-1}$.

The definition of $m$ and the mean value theorem therefore give
\[
    |f_\eta(\theta)-x_\eta^\star|
    \geq
    m|\theta-q_k|
    \qquad
    \text{whenever }|\theta-q_k|\leq\rho,
    \quad k\in\{0,-1\}.
\]
Outside these two neighbourhoods, the monotonicity of the two
branches of $f_\eta$ shows that
\[
    |f_\eta(\theta)-x_\eta^\star|\geq m\rho,
\]
except possibly when $f_\eta(\theta)\leq0$, in which case the
corresponding lower bound is $x_\eta^\star\geq m\varepsilon
\geq m\rho$. Thus, in all cases,
\[
    \begin{cases}
        |f_\eta(\theta)-x_\eta^\star|,
            &\theta>\theta_0,\\
        x_\eta^\star,
            &\theta\leq\theta_0
    \end{cases}
    \geq
    m\min\bigl\{
        |\theta-q_0|,
        |\theta-q_{-1}|,
        \rho
    \bigr\}.
\]
Finally, since $\rho=c\varepsilon$ and $c\leq1$,
\[
    \min\{a,b,c\varepsilon\}
    \geq
    c\min\{a,b,\varepsilon\}
    \qquad (a,b\geq0).
\]
Combining this inequality with
\eqref{eq:disagreement_lower_f} proves the claim.
\end{proof}

\subsection{Exponential Distribution: Technical Lemmas}
\label{app:sub_exp_technical_lemmas}

In this appendix, we derive useful technical results for thresholded log-likelihood rewards with $\cF_{\Exp}$.
Throughout this subsection, let
\[
    \theta_0 \eqdef e^{-\eta},
    \qquad
    \wt\theta_0 \eqdef e^{1-\eta},
    \qquad
    x_0 \eqdef e^{\eta-1},
    \qquad
    f_\eta(y) \eqdef \frac{\eta+\log y}{y}
    \quad (y>0).
\]

\begin{lemma}[Basic properties of $f_\eta$]
\label{lem:init_properties}
For every $y>0$ and $x\ge 0$, we have $r_y(x)=y(f_\eta(y)-x)$; in particular, $r_y(x)=0$ if and only if $x=f_\eta(y)$. Moreover, $f_\eta(y)>0$ if and only if $y>\theta_0$, and $f_\eta$ is strictly increasing on $(0,\wt\theta_0)$, strictly decreasing on $(\wt\theta_0,+\infty)$, and satisfies
\[
    \sup_{y>0} f_\eta(y)
    =
    f_\eta(\wt\theta_0)
    =
    x_0 .
\]
\end{lemma}

\begin{proof}
The identity $r_y(x)=y(f_\eta(y)-x)$ follows from $r_y(x)=-yx+\log y+\eta$. Also, $f_\eta(y)>0$ is equivalent to $\eta+\log y>0$, hence to $y>\theta_0$. Finally,
\[
    f_\eta'(y)=\frac{1-\eta-\log y}{y^2},
\]
so $f_\eta$ is increasing before $\wt\theta_0=e^{1-\eta}$ and decreasing after it. Therefore its maximum is attained at $\wt\theta_0$, with value
\[
    f_\eta(\wt\theta_0)
    =
    \frac{\eta+\log(e^{1-\eta})}{e^{1-\eta}}
    =
    e^{\eta-1}
    =
    x_0 .
\]
\end{proof}

\begin{lemma}[Inverse branches of $f_\eta$]
\label{lem:inverse_branches_f_eta}
Let $\overline W_0$ and $\overline W_{-1}$ be defined by Eq.~\eqref{eq:lambert_transforms}. For every $x\in(0,x_0)$, the equation $f_\eta(\theta)=x$ has exactly two solutions on $(\theta_0,+\infty)$, namely
\[
    \theta_{\eta,0}(x)
    \eqdef
    \frac{1}{x}\overline W_0(\eta-\log x),
    \qquad
    \theta_{\eta,-1}(x)
    \eqdef
    \frac{1}{x}\overline W_{-1}(\eta-\log x).
\]
They satisfy
\[
    \theta_0
    <
    \theta_{\eta,0}(x)
    <
    \wt\theta_0
    <
    \theta_{\eta,-1}(x),
\]
with $x\mapsto\theta_{\eta,0}(x)$ strictly increasing and $x\mapsto\theta_{\eta,-1}(x)$ strictly decreasing on $(0,x_0)$. 
Moreover, for $x\in(0,x_0)$,
\begin{align*}
    &\{\theta>\theta_0:r_\theta(x)>0\}
    =
    \bigl(\theta_{\eta,0}(x),\theta_{\eta,-1}(x)\bigr)\: , \\
    &\{\theta>\theta_0:r_\theta(x)<0\}
    =
    (\theta_0,\theta_{\eta,0}(x))
    \cup
    (\theta_{\eta,-1}(x),+\infty) \:.
\end{align*}
Consequently, if $\theta^\star\in(\theta_0,+\infty)\setminus\{\wt\theta_0\}$ and $x^\star_\eta\eqdef f_\eta(\theta^\star)$, then
\[
    \{\theta>\theta_0:f_\eta(\theta)=x^\star_\eta\}
    =
    \{\theta_{\eta,0}(x^\star_\eta),\theta_{\eta,-1}(x^\star_\eta)\},
\]
and one of these two parameters is $\theta^\star$.
\end{lemma}

\begin{proof}
Fix $x\in(0,x_0)$. The equation $f_\eta(\theta)=x$ is equivalent, after setting $u=x\theta$, to
\[
    u-\log u=\eta-\log x.
\]
Since $x<x_0=e^{\eta-1}$, we have $\eta-\log x>1$. By the definition of the transformed Lambert branches in Eq.~\eqref{eq:lambert_transforms}, this equation has exactly two solutions, $\overline W_0(\eta-\log x)\in(0,1)$ and $\overline W_{-1}(\eta-\log x)>1$, which gives the displayed formulas for $\theta_{\eta,0}(x)$ and $\theta_{\eta,-1}(x)$.

By Lemma~\ref{lem:init_properties}, $f_\eta$ is positive on $(\theta_0,+\infty)$, strictly increasing on $(\theta_0,\wt\theta_0)$, strictly decreasing on $(\wt\theta_0,+\infty)$, and has maximum $x_0$ at $\wt\theta_0$. Hence the two solutions lie respectively in $(\theta_0,\wt\theta_0)$ and $(\wt\theta_0,+\infty)$, and the monotonicity of the inverse branches follows from the monotonicity of $f_\eta$ on these two intervals.

Finally, since $r_\theta(x)=\theta(f_\eta(\theta)-x)$ and $\theta>0$, the sign of $r_\theta(x)$ is the sign of $f_\eta(\theta)-x$. The displayed sign descriptions follow immediately from the two-branch shape of $f_\eta$. Taking $x=x^\star_\eta=f_\eta(\theta^\star)$ gives the final claim.
\end{proof}

\paragraph{The Lambert $W$ Function.}
The Lambert $W$ function is defined by $W(z)e^{W(z)}=z$. We use the two real branches $W_0$ and $W_{-1}$ on negative arguments, and define the transformed branches, for $s\ge 1$, by
\begin{equation}
\label{eq:lambert_transforms}
    \overline W_0(s)
    \eqdef
    -W_0(-e^{-s}),
    \qquad
    \overline W_{-1}(s)
    \eqdef
    -W_{-1}(-e^{-s}).
\end{equation}

\begin{lemma}[Transformed Lambert branches]
\label{lem:lambert_branches_properties}
Let $h(u)\eqdef u-\log u$ for $u>0$. For every $s>1$, the equation $h(u)=s$ has exactly two positive solutions, namely $\overline W_0(s)\in(0,1)$ and $\overline W_{-1}(s)>1$. Moreover, $\overline W_0$ is strictly decreasing and $\overline W_{-1}$ is strictly increasing on $(1,+\infty)$.
\end{lemma}

\begin{proof}
If $u=-W_k(-e^{-s})$, then $W_k(-e^{-s})=-u$, and the identity $W_k(z)e^{W_k(z)}=z$ gives $ue^{-u}=e^{-s}$, equivalently $u-\log u=s$. Since $h'(u)=1-1/u$, the function $h$ is strictly decreasing on $(0,1)$, strictly increasing on $(1,+\infty)$, and satisfies $h(1)=1$ and $h(u)\to+\infty$ as $u\to0^+$ or $u\to+\infty$. Hence, for each $s>1$, there is exactly one solution in $(0,1)$ and exactly one solution in $(1,+\infty)$. The monotonicity of $\overline W_0$ and $\overline W_{-1}$ follows by inverting the two monotone branches of $h$.
\end{proof}

\subsection{Pointwise Qualities versus Pairwise Preferences}
\label{sub:pointwise_qualities_versus_pairwise_preferences}
This section gives examples in which directly observed pairwise preferences are more informative than pointwise quality labels. 
The underlying mechanism is particularly transparent in one dimension, i.e., $k=1$. 
Let $\theta^\star \in \Theta\subseteq \R$ be the true parameter to be estimated.
\paragraph{Meta-example.}
Let $m_n \in [n]$ be the number of independent pairs.
Assume there exists $b_{\star} \in (0,1)$ such that $m_n \ge b_{\star}  n $.
Suppose that each deterministic preference feedback can be rewritten as
\[
    \forall k \in [m_n], \quad V_{k} \eqdef \sign(\theta^\star-B_k) \: ,
\]
where the random cut points \(B_k\) have a density bounded below by \(c_{\star}>0\) in a neighborhood of \(\theta^\star\). 
Let
\[
    B_{n,\max}^{-} \eqdef \max\{B_k:B_k<\theta^\star\} \quad \text{and}\quad  B_{n,\min}^{-} \eqdef \min\{B_k:B_k>\theta^\star\}.
\]
Using the positive density lower bound, we have, for every sufficiently small \(t>0\),
\[
    \bP\left( \theta^\star- B_{n,\max}^{-} >t \:\lor\: B_{n,\min}^{-} -\theta^\star>t \right) \leq 2e^{-m_n c_{\star}t} \le 2e^{-b_{\star} c_{\star}  n  t}  \: .
\]
Consequently, every parameter consistent with these preferences lies, with probability at least \(1-\delta\), within distance $\frac{\log(2/\delta)}{b_{\star} c_{\star}  n}$ of \(\theta^\star\). 
Thus, a positive density of preference-induced cut points at the true parameter yields a fast estimation rate, i.e., a high-probability upper bound \(\mathcal O(1/n)\) on $|\wh\theta_n -\theta^\star|$ for every consistent estimator $\wh\theta_n $, when $n$ is large enough.
\paragraph{Example 1: Gaussian Samples with a Positive Likelihood Reward.}
Consider $\cF_{\cN,\sigma^2}$ for $\theta^\star\in\R$ and the reward
\[
    \forall x \in \R,\quad r_\theta(x) \eqdef \sqrt{2\pi\sigma^2}p_{\theta}(x) \: .
\]
This reward is proportional to the likelihood, but is not a thresholded log-likelihood reward. 
Since \(r_\theta(x)>0\) for every \((\theta,x)\), all pointwise quality labels are positive and therefore provide no information beyond the samples. 
The resulting experiment retains the classical \(\Theta(n^{-1/2})\) estimation rate of the Gaussian location model.

Pairwise preferences retain substantially more information. 
To compare the two feedback models at the same budget of \(n\) samples, form \(m_n=\lfloor n/2\rfloor\) disjoint pairs and suppose that, for each \(k\in[m_n]\), the learner directly observes the preference label
\[
    \forall k \in [m_n], \quad W_k \eqdef \sign\left( r_{\theta^\star}(X_{2k-1}) - r_{\theta^\star}(X_{2k}) \right) \: .
\]
Importantly, \(W_k\) is supplied by the pairwise-preference oracle; it cannot be recovered from the pointwise quality labels, which are all positive in this example.
Define the corresponding preference version space by
\[
    \cC_{m_n}^{\mathrm{pref}} \eqdef \left\{ \theta\in\R \mid \forall k \in [m_n], \: W_k \left( r_\theta(X_{2k-1})-r_\theta(X_{2k}) \right) \ge 0\right\}.
\]
For
\[
    B_k
    \eqdef
    \frac{X_{2k-1}+X_{2k}}{2\sigma^2}
    \sim
    \cN\left(\theta^\star,\frac{1}{2\sigma^2}\right),
\]
a direct calculation gives, for every candidate parameter \(\theta\),
\[
    \sign\left(
        r_\theta(X_{2k-1})-r_\theta(X_{2k})
    \right)
    \sign(X_{2k-1}-X_{2k})
    =
    \sign(\theta-B_k).
\]
In particular, $W_k\sign(X_{2k-1}-X_{2k}) = \sign(\theta^\star-B_k)$.
Consequently, the \(k\)-th consistency constraint is equivalent, almost surely, to $(\theta-B_k)(\theta^\star-B_k)\geq 0$, meaning that \(\theta\) and \(\theta^\star\) must lie on the same side of \(B_k\). 
Therefore,
\[
    \cC_{m_n}^{\mathrm{pref}}
    =
    \left[
        \max_{k:B_k<\theta^\star} B_k,
        \;
        \min_{k:B_k>\theta^\star} B_k
    \right],
\]
with the usual conventions when one of the two sets is empty.

The density of $B_k$ is bounded below by
$c_\sigma \eqdef \sigma/(e\sqrt{\pi})$ on
$[\theta^\star-1/\sigma,\theta^\star+1/\sigma]$.
Using the above, we have, for
$n \ge O(\log(1/\delta))$, with probability
at least $1-\delta$,
\[
    \sup_{\theta\in\cC_{m_n}^{\mathrm{pref}}}
    |\theta-\theta^\star|
    \le \frac{\log(2/\delta)}{m_n c_\sigma}
    = O\left(\frac{\log(1/\delta)}{\sigma n}\right) \: .
\]

Thus, although the pointwise signs of the likelihood carry no information, directly observed pairwise preferences yield fast localization.

\paragraph{Example 2: Thresholded Log-likelihood with Nonconstant Qualities.}
The preceding separation might appear to rely on all pointwise labels being identical. 
We next construct a thresholded log-likelihood example in which both labels occur, but remain uninformative because they are already determined by the samples.

Let \(\cX=[-1,1]\), \(\Theta=[-3/2,-1/2]\), and consider the family of densities
\[
    p_\theta(x) \eqdef
    \frac12\left[
        1+\frac14\sign(x)
        +\frac1{20}\left(x^2-\frac13+\theta x\right)
    \right],
    \qquad x\in[-1,1] \: .
\]
This is a valid parametric family. Indeed,
\[
    \int_{-1}^1\sign(x)\,dx
    =
    \int_{-1}^1x\,dx
    =
    \int_{-1}^1\left(x^2-\frac13\right)dx
    =0 \: ,
\]
so that \(p_\theta\) integrates to one. 
Moreover, 
\[
    \forall (\theta,x)\in\Theta\times\cX,\quad \left|x^2-\frac13+\theta x\right| \leq \frac{13}{6}\quad \text{, hence}\quad p_\theta(x) \geq \frac12\left(1-\frac14-\frac{13}{120}\right) = \frac{77}{240}>0 \: .
\]
Consider the thresholded log-likelihood reward with $\eta= \log 2$.
Then,
\[
    \sign(r_\theta(x))= \sign\bigl(\log(2p_\theta(x))\bigr) = \sign\bigl(2p_\theta(x)-1\bigr) \: .
\]
The preceding bound gives
\[
    2p_\theta(x)-1
    \begin{cases}
        \displaystyle
        \geq \frac14-\frac{13}{120}
        =\frac{17}{120}>0,
        & x>0,\\[1ex]
        \displaystyle
        \leq-\frac14+\frac{13}{120}
        =-\frac{17}{120}<0,
        & x<0.
    \end{cases}
\]
Therefore,
\[
    \forall (\theta,x)\in\Theta\times\cX,\quad \sign(r_\theta(x))=\sign(x)
\]
Both quality labels occur with positive probability, but the label is completely determined by the observed sample and is independent of the parameter. 
Pointwise qualities thus provide no information beyond the samples.

The sample-only model is a regular one-dimensional parametric family with fixed support and Fisher information
\[
    \cI(\theta)
    =
    \int_{-1}^1
    \frac{(\partial_\theta p_\theta(x))^2}{p_\theta(x)}
    \,dx
    =
    \int_{-1}^1
    \frac{x^2}{1600\,p_\theta(x)}
    \,dx
    \in(0,+\infty).
\]
It therefore has the classical local \(n^{-1/2}\) estimation scale.

Pairwise preferences, by contrast, provide parameter-dependent constraints.
Form \(m\) independent pairs \((X_k,Y_k)_{k\in[m]}\) and suppose that the
learner directly observes $W_k
    \eqdef
    \sign\left(
        r_{\theta^\star}(X_k)-r_{\theta^\star}(Y_k)
    \right)$.
Define the corresponding preference version space by
\[
    \cC_m^{\mathrm{pref}}
    \eqdef
    \left\{
        \theta\in\Theta\mid \forall k\in[m], \:
        W_k\bigl(r_\theta(X_k)-r_\theta(Y_k)\bigr)\geq 0\right\}\:.
\]

Consider a pair \(x,y>0\). Since the logarithm is strictly increasing,
\begin{align*}
    \sign\bigl(r_\theta(x)-r_\theta(y)\bigr)
    &=
    \sign\bigl(p_\theta(x)-p_\theta(y)\bigr)=\sign\bigl((x-y)(x+y+\theta)\bigr).
\end{align*}
Setting $B(x,y)\eqdef-(x+y)$, we therefore obtain, for every candidate parameter \(\theta\),
\[
    \sign\bigl(r_\theta(x)-r_\theta(y)\bigr)\sign(x-y)
    =
    \sign\bigl(\theta-B(x,y)\bigr).
\]
In particular, for the observed preference,
\[
    W_k\sign(X_k-Y_k)
    =
    \sign(\theta^\star-B_k),
    \qquad
    B_k\eqdef-(X_k+Y_k),
\]
whenever \(X_k,Y_k>0\). Consequently,
\begin{align*}
    &W_k\sign\bigl(r_\theta(X_k)-r_\theta(Y_k)\bigr)=\sign(\theta^\star-B_k)\sign(\theta-B_k).
\end{align*}
Since ties occur with probability zero, the \(k\)-th preference-consistency constraint is thus equivalent to $(\theta^\star-B_k)(\theta-B_k)\geq 0$.
In other words, every pair of positive samples requires \(\theta\) and \(\theta^\star\) to lie on the same side of the random cut \(B_k\).

We next lower-bound the density of these cuts. Recall that
\[
    \forall (\theta,x)\in\Theta\times[-1,1],\quad p_\theta(x)\geq p_{\min}\eqdef\frac{77}{240} \: .
\]
For \(b\in(-2,0)\), set \(s=-b\in(0,2)\). 
The unconditional subdensity of \(B_k=-(X_k+Y_k)\) generated by pairs satisfying \(X_k,Y_k>0\) is
\[
    g_{\theta^\star}(b)
    =
    \int_{\max\{0,s-1\}}^{\min\{1,s\}}
        p_{\theta^\star}(x)
        p_{\theta^\star}(s-x)
    \,dx \: .
\]
For \(b\in\Theta=[-3/2,-1/2]\), we have \(s\in[1/2,3/2]\), and the integration interval has length $\min\{s,2-s\}\geq\frac12$.
It follows that, uniformly over \((\theta^\star,b)\in\Theta^2\),
\[
    g_{\theta^\star}(b)
    \geq
    \frac12p_{\min}^2
    =
    \frac12\left(\frac{77}{240}\right)^2
    \eqdef c_0>0
\]
Assume that \(\theta^\star\in\operatorname{int}(\Theta)\), and let $\rho_\star\eqdef\operatorname{dist}(\theta^\star,\partial\Theta)>0$.
Then, for every \(t\in(0,\rho_\star]\),
\[
    \min \{ \bP_{\theta^\star}\left(
        X_k,Y_k>0,\,
        B_k\in[\theta^\star,\theta^\star+t]
    \right), \bP_{\theta^\star}\left(
        X_k,Y_k>0,\,
        B_k\in[\theta^\star-t,\theta^\star]
    \right)\}
    \geq c_0t \: .
\]
Define the closest preference-induced cuts on either side of \(\theta^\star\) by
\[
    B_-
    \eqdef
    \max\left\{
        B_k:
        X_k,Y_k>0,\ B_k<\theta^\star
    \right\} \quad \text{and} \quad B_+
    \eqdef
    \min\left\{
        B_k:
        X_k,Y_k>0,\ B_k>\theta^\star
    \right\} \: .
\] 
The consistency constraints associated with these positive pairs imply $\cC_m^{\mathrm{pref}}\subseteq[B_-,B_+]$.
Moreover, independence of the pairs gives
\[
    \max\{\bP_{\theta^\star}
    \left(
        \theta^\star-B_->t
    \right), \bP_{\theta^\star}
    \left(
        B_+-\theta^\star>t
    \right)\}
    \leq
    (1-c_0t)^m
    \leq
    e^{-mc_0t}\: .
\]
Therefore, for $n \ge \cO(\log(1/\delta))$, with probability at least \(1-\delta\),
\[
    \sup_{\theta\in\cC_{m}^{\mathrm{pref}}}
    |\theta-\theta^\star|
    \leq
    \frac{\log(2/\delta)}{m c_0}
    =
    \cO\left(
        \frac{\log(1/\delta)}{m}
    \right).
\]
Thus, the pointwise quality labels are nonconstant but redundant, whereas directly observed pairwise preferences yield fast localization.

\paragraph{Example 3: Truncated Gaussian Location Family.}
We finally detail the truncated Gaussian example mentioned in Section~\ref{sec:estimator4qualities}.
Fix \(\sigma>0\), let \(\cX=[-1,1]\) and \(\Theta=[-1/2,1/2]\), and consider the densities
\[
    \forall x\in[-1,1], \quad p_\theta(x)
    \eqdef
    \frac{
        \exp\left(-\frac{(x-\theta)^2}{2\sigma^2}\right)
    }{
        \sigma\sqrt{2\pi}\,Z_\theta
    },
    \quad \text{where}\quad Z_\theta
    \eqdef
    \Phi\left(\frac{1-\theta}{\sigma}\right)
    -
    \Phi\left(\frac{-1-\theta}{\sigma}\right) \:   ,
\]
is the truncation normalizing constant. 
Consider the thresholded log-likelihood reward $r_\theta(x)\eqdef\log p_\theta(x)+\eta$.
Because \(|x-\theta|\leq 3/2\) and \(Z_\theta\leq 1\), we have, uniformly over \((\theta,x)\in\Theta\times\cX\),
\[
    p_\theta(x)
    \geq
    \frac{1}{\sigma\sqrt{2\pi}}
    \exp\left(-\frac{9}{8\sigma^2}\right)
    \eqdef p_{\min}>0 \: .
\]
Therefore, choosing $\eta>-\log p_{\min}$ ensures that $r_\theta(x)>0$ for all $(\theta,x)\in\Theta\times\cX$.
Consequently, all pointwise quality labels are positive and provide no information beyond the samples.

The sample-only model is nevertheless a regular one-dimensional parametric family with fixed support. 
Indeed,
\[
    \partial_\theta\log p_\theta(x)
    =
    \frac{x-\theta}{\sigma^2}
    -
    \partial_\theta\log Z_\theta
    =
    \frac{x-\bE_\theta[X]}{\sigma^2},
\]
and hence its Fisher information is
\[
    \cI(\theta)
    =
    \frac{\mathrm{Var}_\theta(X)}{\sigma^4}
    \in(0,+\infty).
\]
Thus, samples alone, and therefore samples augmented with the uninformative
pointwise labels, retain the classical local \(n^{-1/2}\) estimation scale.

Pairwise preferences, by contrast, yield parameter-dependent constraints.
To compare both feedback models at the same budget of \(n\) samples, form
\(m_n=\lfloor n/2\rfloor\) independent pairs and suppose that the learner
directly observes
\[
    \forall k\in[m_n], \quad W_k
    \eqdef
    \sign\left(
        r_{\theta^\star}(X_{2k-1})
        -
        r_{\theta^\star}(X_{2k})
    \right)\: .
\]
Define the preference version space
\[
    \cC_{m_n}^{\mathrm{pref}}
    \eqdef
    \left\{
        \theta\in\Theta\mid \forall k\in[m_n] , \: 
        W_k
        \left(
            r_\theta(X_{2k-1})
            -
            r_\theta(X_{2k})
        \right)
        \geq 0\right\}.
\]

For any \(x,y\in[-1,1]\), the additive threshold and the parameter-dependent
normalizing constant cancel from the reward difference:
\begin{align*}
    r_\theta(x)-r_\theta(y)
    &=
    -\frac{(x-\theta)^2}{2\sigma^2}
    +\frac{(y-\theta)^2}{2\sigma^2}=\frac{(x-y)(2\theta-x-y)}{2\sigma^2}.
\end{align*}
Setting $B(x,y)\eqdef\frac{x+y}{2}$, we obtain
\[
    \sign\bigl(r_\theta(x)-r_\theta(y)\bigr)\sign(x-y) = \sign\bigl(\theta-B(x,y)\bigr).
\]
In particular, the observed preference satisfies
\[
    W_k\sign(X_{2k-1}-X_{2k}) =\sign(\theta^\star-B_k)\quad \text{with}\quad B_k\eqdef\frac{X_{2k-1}+X_{2k}}{2} \: .
\]
Consequently,
\begin{align*}
    &W_k
    \sign\left(
        r_\theta(X_{2k-1})
        -
        r_\theta(X_{2k})
    \right)=\sign(\theta^\star-B_k)\sign(\theta-B_k).
\end{align*}
Since ties occur with probability zero, the \(k\)-th preference-consistency constraint is equivalent to $(\theta^\star-B_k)(\theta-B_k)\geq 0$.
Thus, each observed preference requires \(\theta\) and \(\theta^\star\) to lie on the same side of the random midpoint \(B_k\).

It remains to verify that these random midpoints have positive density around \(\theta^\star\). 
The density of \(B_k=(X_{2k-1}+X_{2k})/2\) is
\[
    g_{\theta^\star}(b)
    =
    2
    \int_{\max\{-1,\,2b-1\}}^{\min\{1,\,2b+1\}}
        p_{\theta^\star}(x)
        p_{\theta^\star}(2b-x)
    \,dx,
    \qquad b\in[-1,1].
\]
For \(b\in\Theta=[-1/2,1/2]\), the integration interval has length $2-2|b|\geq 1$.
Using \(p_{\theta^\star}\geq p_{\min}\), we therefore obtain the uniform lower bound
\[
    g_{\theta^\star}(b)
    \geq
    2p_{\min}^2
    \eqdef c_{\sigma}>0,
    \qquad
    (\theta^\star,b)\in\Theta^2,
\]
where one may take
\[
    c_{\sigma}
    =
    \frac{1}{\pi\sigma^2}
    \exp\left(-\frac{9}{4\sigma^2}\right).
\]

Assume first that \(\theta^\star\in\operatorname{int}(\Theta)\), and define $\rho_\star
    \eqdef
    \operatorname{dist}(\theta^\star,\partial\Theta)>0$.
For every \(t\in(0,\rho_\star]\), the density lower bound gives
\[
    \min\{\bP_{\theta^\star}
    \left(
        B_k\in[\theta^\star,\theta^\star+t]
    \right), \bP_{\theta^\star}
    \left(
        B_k\in[\theta^\star-t,\theta^\star]
    \right)\}
    \geq c_\sigma t\: .
\]
Let
\[
    B_-\eqdef\max\{B_k:B_k<\theta^\star\},
    \qquad
    B_+\eqdef\min\{B_k:B_k>\theta^\star\}.
\]
The preference-consistency constraints imply $\cC_{m_n}^{\mathrm{pref}}\subseteq[B_-,B_+]$.
Moreover,
\[
    \max\{\bP_{\theta^\star}(\theta^\star-B_->t),\bP_{\theta^\star}(B_+-\theta^\star>t)\}
    \leq
    (1-c_\sigma t)^{m_n}
    \leq
    e^{-m_nc_\sigma t} \: .
\]
Therefore, for $n \ge \cO(\log(1/\delta))$, with probability at least \(1-\delta\),
\[
    \sup_{\theta\in\cC_{m}^{\mathrm{pref}}}
    |\theta-\theta^\star|
    \leq
    \frac{\log(2/\delta)}{m_nc_\sigma}
    =
    \cO\left(
        \frac{\log(1/\delta)}{n}
    \right)\:.
\]
At the endpoints of \(\Theta\), the same conclusion follows from the corresponding one-sided argument. 
Hence, although the pointwise quality labels are completely uninformative, directly observed pairwise preferences yield
fast localization.


\section{Proofs of Section~\ref{sec:reduction_to_preferences}}
\label{app:reduction_to_preferences}
In this appendix, we prove Theorem~\ref{thm:specialized_threshold_misspecification} in Appendix~\ref{sub:proof_of_theorem_ref_thm_specialized_threshold_misspecification}, and its extension, Theorem~\ref{thm:classifier_to_parameter_misspecification} in Appendix~\ref{sub:threshold_misspecification}, under suitable assumptions on \(r_\theta\) and \(p_\theta\) (Assumptions~\ref{ass:finite_VC_misspecification},~\ref{ass:conversion_error_to_parameter_misspecification} and~\ref{ass:closure_strict_consistency_misspecification}).
\subsection{Threshold Misspecification}
\label{sub:threshold_misspecification}

We propose suitable geometric assumptions on \(r_\theta\) and \(p_\theta\) to obtain fast convergence rates for any $\wh \theta_n\in \cC_{n}^{\all}$.
\paragraph{Uniform PAC Bound.}
Let $\cH^{\all} \eqdef \left\{ h_{\theta,\eta}^{\all} \mid (\theta,\eta)\in\Theta\times\R \right\}$ be the class of thresholded classifiers, namely
\[
    \forall x \in \cX,\quad h_{\theta,\eta}^{\all}(x) \eqdef \begin{cases}
        +1 & \text{if }r_\theta(x) \ge \eta \\
        -1 & \text{otherwise}
    \end{cases} \: .
\]
Since the reward model is known only up to an additive constant, its unknown additive constant can equivalently be represented by an unknown threshold $\eta^\star$. 
Without loss of generality, we absorb $\eta^\star$ into the reward and take $\eta^\star=0$. 
The classification problem is therefore realizable, since $h_{\theta^\star,0}^{\all}\in\cH^{\all}$. 
Using $\cI_n^{-} =  \{i\in [n] \mid r_{\theta^\star}(X_i) < 0\}$ yields
\[
    \forall i \in \cI_n^{-}, \: h_{\theta^\star,0}^{\all}(X_i)=-1 \quad \text{and}\quad \forall i \in [n]\setminus \cI_n^{-}, \: h_{\theta^\star,0}^{\all}(X_i)=+ 1  \: .
\]
\begin{assumption} \label{ass:finite_VC_misspecification}
    $d^{\all} \eqdef \mathrm{vc}(\cH^{\all})$ is finite.
\end{assumption}
Let the version space $V_n^{\all}$ and the strictly consistent pairwise set $\cC_{n,>}^{\all}$ be defined by
\begin{align*}
    &V_n^{\all} \eqdef \left\{ (\theta,\eta)\in\Theta\times\R \mid \forall i\in[n], \: h_{\theta,\eta}^{\all}(X_i)=h_{\theta^\star,0}^{\all}(X_i)  \right\} \quad \text{and}\\
    &\cC_{n,>}^{\all} \eqdef \left\{  \theta\in\Theta \mid \forall (i,j)\in([n]\setminus \cI_n^{-})\times\cI_n^- ,\: \ell_\theta(X_i,X_j)>0\right\}\: .
\end{align*}
\begin{lemma}\label{lem:projection_version_space}
    Almost surely, $\cC_{n,>}^{\all}= \mathrm{Proj}_{\Theta}(V_n^{\all})=\left\{ \theta\in\Theta \mid  \exists \eta\in\R,\  (\theta,\eta)\in V_n^{\all} \right\}$.
\end{lemma}
\begin{proof}
Fix $\theta\in\Theta$ and suppose that both labels are observed.
Then, we have
\begin{align*}
    &\exists \eta \in\R,\: (\theta,\eta)\in V_n^{\all} \\
    \iff\quad & \forall i \in [n]\setminus \cI_n^{-},\: r_\theta(X_i)\geq \eta \quad\text{and}\quad \forall i \in \cI_n^{-}, \: r_\theta(X_j)<\eta \\
    \iff\quad & \max_{j\in\cI_n^-}r_\theta(X_j) < \min_{i\in[n]\setminus \cI_n^{-}}r_\theta(X_i)\\
    \iff\quad & \forall(i,j)\in([n]\setminus \cI_n^{-})\times\cI_n^- , \: \ell_\theta(X_i,X_j)>0 \: .
\end{align*}
This is precisely the condition for $\theta \in \cC_{n,>}^{\all} $.

If only positive labels are observed, one may take $\eta=\min_{i\in[n]}r_\theta(X_i)$. If only negative labels are observed, one may choose $\eta>\max_{i\in[n]}r_\theta(X_i)$. 
The conclusion therefore also holds in these two cases.
\end{proof}
Let us define the minimum classification error over thresholds, named \textit{profiled} classification error, as
\begin{equation} \label{eq:profiled_classification_error}
    \forall \theta \in \Theta,\quad \mathrm{er}_{\mathrm{prof}}(\theta^\star,\theta) \eqdef \inf_{\eta \in\R} p_{\theta^\star}\!\left( h_{\theta,\eta}^{\all}(X) \neq h_{\theta^\star,0}^{\all}(X) \right) \: .
\end{equation}

Under Assumption~\ref{ass:finite_VC_misspecification}, combining Lemma~\ref{lem:projection_version_space} with~\citet[Theorem 11]{2946645.3007088} yields, with probability at least $1-\delta$,
\begin{align*}
    \sup_{\theta\in\cC_{n,>}^{\all}} \mathrm{er}_{\mathrm{prof}}(\theta^\star,\theta)  \le  \sup_{(\theta,\eta)\in V_n^{\all}} p_{\theta^\star}\!\left( h_{\theta,\eta}^{\all}(X) \neq h_{\theta^\star,0}^{\all}(X) \right) \le \frac{8}{n} \left(  d^{\all} \ln\left( \frac{49en}{d^{\all}}+37 \right) + 8\ln\left(\frac{6}{\delta}\right) \right) \: .
\end{align*}

\paragraph{Local Error-to-parameter Conversion.}
Assumption~\ref{ass:conversion_error_to_parameter_misspecification} converts control of the profiled classification error into local control of the parameter error.
\begin{assumption} \label{ass:conversion_error_to_parameter_misspecification}
    There exist constants $\alpha_{\star}^{\all},\epsilon_{\star}^{\all}>0$ such that
    \[
        \forall \theta\in \Theta,\quad \mathrm{er}_{\mathrm{prof}}(\theta^\star,\theta) \ge \alpha_{\star}^{\all} \min\left\{ \|\theta-\theta^\star\|, \epsilon_{\star}^{\all} \right\} \: .
    \]
\end{assumption}
Assumption~\ref{ass:closure_strict_consistency_misspecification} ensures that a bound established for strictly consistent parameters extends by continuity to the weakly consistent set $\cC_n^{\all}$.
Using the continuity of $\theta\mapsto \|\theta- \theta^\star\|$, it implies that $\sup_{\theta\in\cC_n^{\all}} \|\theta-\theta^\star\| = \sup_{\theta\in\cC_{n,>}^{\all}} \|\theta-\theta^\star\|$.
\begin{assumption}\label{ass:closure_strict_consistency_misspecification}
    Almost surely, $\cC_n^{\all} = \overline{\cC_{n,>}^{\all}}$ where $\overline{X}$ denotes the closure of the set $X$.
\end{assumption}
\paragraph{Fast Estimation Rate.}
Under these assumptions, Theorem~\ref{thm:classifier_to_parameter_misspecification} gives a \(\widetilde{\cO}(1/n)\) bound on $\sup_{\theta \in \cC_{n}^{\all}}\|\theta - \theta^\star\|$ with high probability; thus fast convergence rates on $\cC_{n}^{\all}$.
\begin{theorem} \label{thm:classifier_to_parameter_misspecification}
    Let $\delta \in (0,1)$.
    Let $\theta^\star \in \Theta$ such that Assumptions~\ref{ass:finite_VC_misspecification},~\ref{ass:conversion_error_to_parameter_misspecification} and~\ref{ass:closure_strict_consistency_misspecification} hold true.
    For all $n \ge  \cO\left(\frac{\log(1/\delta)}{\epsilon^{\all}_{\star} \alpha^{\all}_{\star}}\right)$,  
    \begin{align*}
        \sup_{\theta \in \cC_{n}^{\all}}\|\theta - \theta^\star\| \le \frac{\cO \left( d^{\all}\ln \left( n/d^{\all} \right) + \ln \left(1/\delta\right)\right)}{n \alpha^{\all}_{\star}} \quad \text{with probability at least $1-\delta$}\: .
    \end{align*}
\end{theorem}
\begin{proof}
Let $\delta\in (0,1)$.
Let $\theta^\star \in \Theta$ such that Assumptions~\ref{ass:finite_VC_misspecification}~\ref{ass:conversion_error_to_parameter_misspecification} and~\ref{ass:closure_strict_consistency_misspecification} hold true for $d^{\all}$ and $\alpha^{\all}_{\star}, \varepsilon^{\all}_{\star} > 0$.
Combining the above results, we have shown, with probability at least $1-\delta$,
\begin{align*}
    \alpha^{\all}_{\star} \min\left\{ \sup_{\theta \in \cC_{n}^{\all}} \left\|\theta-\theta^\star\right\|, \varepsilon^{\all}_{\star} \right\} \le\frac{8}{n}\left( d^{\all} \ln \left( \frac{49 e n}{d^{\all}} + 37\right) + 8 \ln \left(\frac{6}{\delta}\right)\right) \: .
\end{align*}
Let us define
\[
    n^{\all}_{\star}(\delta) \eqdef 1+\left\lceil \max\left\{ d^{\all},\, \frac{128}{\varepsilon^{\all}_{\star} \alpha^{\all}_{\star}}\ln\frac{6}{\delta},\, \frac{32d^{\all}}{\varepsilon^{\all}_{\star} \alpha^{\all}_{\star}} \ln\left( e+\frac{32(49e+37)}{\varepsilon^{\all}_{\star} \alpha^{\all}_{\star}} \right) \right\} \right\rceil  =_{\delta \to 0}  \cO\left(\frac{\log(1/\delta)}{\epsilon^{\all}_{\star} \alpha^{\all}_{\star}}\right) \: .
\]
Therefore, for all $n \ge n^{\all}_{\star}(\delta)$, we have, with probability at least $1-\delta$,
\begin{align*}
     \sup_{\theta \in \cC_{n}^{\all}} \left\|\theta-\theta^\star\right\| \le \frac{8}{\alpha^{\all}_{\star} n}\left( d^{\all} \ln \left( \frac{49 e n}{d^{\all}} + 37\right) + 8 \ln \left(\frac{6}{\delta}\right)\right) \: .
\end{align*}
This concludes the proof.
\end{proof}

\subsection{Proof of Theorem~\ref{thm:specialized_threshold_misspecification}}
\label{sub:proof_of_theorem_ref_thm_specialized_threshold_misspecification}

Let $\delta \in (0,1)$, $\eta_0 = \frac{1}{2}\log((2\pi)^d|\Sigma|)$ for $\cF_{\cN, \Sigma}$ and $\eta_0 = \log(2b)$ for $\cF_{\lap, b}$ with $d=1$.
Recall that $\|x\|_{\Sigma} \eqdef \sqrt{x\transpose \Sigma x}$.
Let $\eta \in (\eta_0,+ \infty) $ be a known threshold, and define $\bar \eta \eqdef \eta - \eta_0$.
For thresholded log-likelihood rewards, we have
\[
    r_{\theta}(x) \eqdef \bar \eta - \begin{cases}
        \frac{1}{b}|x - \theta| &\text{for } \cF_{\lap,b}\\
        \frac{1}{2} \| \Sigma^{-1}x - \theta\|_{\Sigma}^2 &\text{for } \cF_{\cN,\Sigma}
    \end{cases} \quad \text{,}\quad c_{\eta} \eqdef \begin{cases}
        b\bar\eta &\text{for } \cF_{\lap,b}\\
        \sqrt{2\bar \eta} &\text{for } \cF_{\cN,\Sigma}
    \end{cases}\quad \text{and}\quad r_{\eta} \eqdef \begin{cases}
        b\bar\eta &\text{for } \cF_{\lap,b}\\
        \sigma\sqrt{2\bar \eta} &\text{for } \cF_{\cN,\sigma^2}
    \end{cases} \: .
\]
To keep the same argument for both families, we define
\[
m_\theta:=
\begin{cases}
\theta,&\mathcal F_{\mathrm{Lap},b},\\
\sigma^2\theta,&\mathcal F_{\mathcal N,\sigma^2}.
\end{cases}
\]

\paragraph{Case $\cF_{\lap,b}$ and $\cF_{\cN,\sigma^2}$.}
For all $x_{1:n}\in\cX^n$, let us define
\begin{align*}
    &C_n^{\all}(x_{1:n}) \eqdef \left\{ \theta\in\Theta \mid \forall i,j\in[n]\text{ s.t. }  r_{\theta^\star}(x_i)r_{\theta^\star}(x_j)<0,\: \ell_{\theta^\star}(x_i,x_j) \ell_\theta(x_i,x_j) \geq 0  \right\} \: , \\
    &\psi_n^{\mathrm{all}}(x_{1:n}) \eqdef (x_{n,\max}^{L},m_{\theta^\star}-r_{\eta}) \cup(m_{\theta^\star}-r_{\eta},x_{n,\min}^{+}) \cup(x_{n,\max}^{+},m_{\theta^\star}+r_{\eta}) \cup(m_{\theta^\star}+r_{\eta},x_{n,\min}^{R}) \: , \\
    &x_{n,\max}^{L} \eqdef\max\{x_i \mid x_i< m_{\theta^\star}-r_{\eta}\},\quad   x_{n,\min}^{R} \eqdef\min\{x_i  \mid  x_i>m_{\theta^\star}+r_{\eta} \},\\
    &x_{n,\max}^{+} \eqdef\max\{x_i \mid m_{\theta^\star}-r_{\eta} <x_i< m_{\theta^\star}+r_{\eta}\},\quad x_{n,\min}^{+}\eqdef\min\{x_i  \mid  m_{\theta^\star}-r_{\eta} <x_i< m_{\theta^\star}+r_{\eta}\}, 
\end{align*}
with respective conventions $-\infty$, $+\infty$, $m_{\theta^\star}-r_{\eta}$, $m_{\theta^\star}+r_{\eta}$.
Therefore, this defines a consistent monotone sample compression rule of size at most $4$.
Using~\citet[Theorem 3]{2946645.3007088} yields, with probability at least $1-\delta$,
\[
    p_{\theta^\star}(\psi_n^{\mathrm{all}}(X_{1:n}) ) \le \frac{1}{n}\left( 84 + 16 \ln \left(\frac{3}{\delta}\right)\right) \: .
\]

We now convert the probability of the compression region into a parameter-estimation error. 
Define $a^\star \eqdef m_{\theta^\star}-r_{\eta}$, $b^\star \eqdef  m_{\theta^\star}+r_{\eta}$, $\rho_\eta\eqdef\frac{r_\eta}{2}$ and $c_{\mathcal F}(\eta)\eqdef \inf_{|x-m_{\theta^\star}|\leq r_\eta+\rho_\eta} p_{\theta^\star}(x)$.
Since both the Gaussian and Laplace families are location families, $c_{\mathcal F}(\eta)$ does not depend on $\theta^\star$. 
More explicitly,  
\[
    c_{\mathcal F}(\eta)
    =
    \begin{cases}
        \displaystyle
        \frac{1}{\sqrt{2\pi\sigma^2}}
        \exp\left(-\frac{9r_\eta^2}{8\sigma^2}\right),
        &\text{for }\cF_{\cN,\sigma^2},\\[3mm]
        \displaystyle
        \frac{1}{2b}
        \exp\left(-\frac{3r_\eta}{2b}\right),
        &\text{for }\cF_{\lap,b}.
    \end{cases}
\]

Suppose that $p_{\theta^\star} \left(\psi_n^{\all}(X_{1:n})\right) < \frac{c_{\mathcal F}(\eta)r_\eta}{2}$.
Then each of the four empirical extrema lies within distance $r_\eta/2$ of its corresponding population boundary. Indeed, otherwise $\psi_n^{\all}(X_{1:n})$ would contain an interval of length $r_\eta/2$ included in $[m_{\theta^\star}-3r_\eta/2,m_{\theta^\star}+3r_\eta/2]$, and would consequently have probability at least $c_{\mathcal F}(\eta)r_\eta/2$. 
In particular, all four empirical extrema are finite, and
\[
    \cC_n^{\all} = C_n^{\all}(X_{1:n}) = 
    \begin{cases}
     \left[ \frac{X_{n,\max}^{L}+X_{n,\max}^{+}}{2\sigma^2}, \frac{X_{n,\min}^{+}+X_{n,\min}^{R}}{2\sigma^2} \right]  &\text{for }\cF_{\cN,\sigma^2},\\
    \left[ \frac{X_{n,\max}^{L}+X_{n,\max}^{+}}{2}, \frac{X_{n,\min}^{+}+X_{n,\min}^{R}}{2} \right]   &\text{for }\cF_{\lap,b}
    \end{cases} \: .
\]

The four intervals defining $\psi_n^{\all}(X_{1:n})$ are pairwise disjoint and contained in $[m_{\theta^\star}-3r_\eta/2,m_{\theta^\star}+3r_\eta/2]$. 
Therefore,
\begin{align*}
    &p_{\theta^\star}
    \left(
        (X_{n,\max}^{L},a^\star)
        \cup
        (X_{n,\max}^{+},b^\star)
    \right) \geq
    c_{\mathcal F}(\eta)
    \left(
        a^\star-X_{n,\max}^{L}
        +
        b^\star-X_{n,\max}^{+}
    \right), \\
    &p_{\theta^\star}
    \left(
        (a^\star,X_{n,\min}^{+})
        \cup
        (b^\star,X_{n,\min}^{R})
    \right) \geq
    c_{\mathcal F}(\eta)
    \left(
        X_{n,\min}^{+}-a^\star
        +
        X_{n,\min}^{R}-b^\star
    \right).
\end{align*}

Using $a^\star+b^\star=2m_{\theta^\star}$, we obtain
\begin{align*}
    \sup_{\theta\in\cC_n^{\all}}  |\theta-\theta^\star| \leq  \begin{cases} 
    \frac{p_{\theta^\star}(\psi_n^{\all}(X_{1:n}))}{2\sigma^2 c_{\mathcal F}(\eta)}  &\text{for }\cF_{\cN,\sigma^2},\\
    \frac{p_{\theta^\star}(\psi_n^{\all}(X_{1:n}))}{2c_{\mathcal F}(\eta)}  &\text{for }\cF_{\lap,b}
    \end{cases} \: .
\end{align*}
    
Let us define
\begin{equation}\label{eq:csts_gaussian_laplace_misspecification}
    \epsilon_{\cF,\eta}^{\all} \eqdef \begin{cases}
        b(\eta - \log(2b)) &\text{for } \cF_{\lap,b}\\
        \sigma^{-1}\sqrt{2 \eta- \log(2\pi\sigma^2)} &\text{for } \cF_{\cN,\sigma^2}
    \end{cases} \quad \text{and}\quad \alpha_{\cF,\eta}^{\all} \eqdef \begin{cases}    
        2\sqrt{2b}e^{-3\eta/2},
        &\text{for }\cF_{\lap,b}\: ,\\
        2\sigma^2(2\pi\sigma^2)^{\frac{5}{8}} e^{-9\eta/4},
        &\text{for }\cF_{\cN,\sigma^2} .
    \end{cases} \: .
\end{equation}
which satisfy $\epsilon_{\cF,\eta}^{\all}>0$ and $\alpha_{\cF,\eta}^{\all}>0$.
Therefore, for $n> \frac{ 4\left(84+16\ln(3/\delta)\right)}{\alpha_{\cF,\eta}^{\all}\epsilon_{\cF,\eta}^{\all}}$, we have, with probability at least $1-\delta$,  then
\[
    \sup_{\theta\in\cC_n^{\all}}  |\theta-\theta^\star| \le \frac{84+16\ln(3/\delta)}{ \alpha_{\cF,\eta}^{\all} n } \: .
\]
This concludes the proof of the first result on $\cF_{\lap,b}$ and $\cF_{\cN,\sigma^2}$.

\paragraph{Case $\cF_{\cN,\Sigma}$.}
This result is a corollary of Theorem~\ref{thm:classifier_to_parameter_misspecification}, hence it remains to show Assumptions~\ref{ass:finite_VC_misspecification},
\ref{ass:conversion_error_to_parameter_misspecification}, and
\ref{ass:closure_strict_consistency_misspecification}. 

For every \((\theta,\eta)\in\R^d\times\R\), the positive region of \(h_{\theta,\eta}^{\all}\) is empty when $\eta < \eta_0\eqdef \frac{1}{2} \log((2\pi)^d|\Sigma|)$, and, when $\eta\ge\eta_0$, of the form $\left\{ x\in\R^d \mid \|\Sigma^{-1}x-\theta\|_\Sigma^2 \le c_\eta \right\}$, where \(c_\eta = \sqrt{2\bar\eta}\) and $\bar \eta = \eta - \eta_0 \ge 0$.
Under the invertible affine transformation \(x\mapsto\Sigma^{-1/2}x\), these sets become Euclidean balls in \(\R^d\).
Since the class of Euclidean balls in \(\R^d\) has VC dimension \(d+1\), $d^{\all} \eqdef \vc(\cH^{\all})\le d+1$.
Thus Assumption~\ref{ass:finite_VC_misspecification} holds true with explicit constant $d+1$.

The function $\theta \mapsto \ell_\theta(x,y)$ is affine for all pair $(x,y) \in \cX^2$ since
\[
    \ell_\theta(x,y) = r_\theta(x)-r_\theta(y) = \langle x-y, \theta - \Sigma^{-1} (x+y)/2\rangle \: .
\]
Moreover, \(\theta^\star\in\cC_{n,>}^{\all}\) since $\min_{(i,j)\in([n]\setminus\cI_n^-)\times\cI_n^-}\ell_{\theta^\star}(X_i,X_j) >0$.
For any \(\theta\in\cC_n^{\all}\) and \(t\in(0,1]\), affine linearity gives
\[
    \ell_{(1-t)\theta+t\theta^\star}(X_i,X_j)
    =
    (1-t)\ell_\theta(X_i,X_j)
    +t\ell_{\theta^\star}(X_i,X_j)>0.
\]
Hence \((1-t)\theta+t\theta^\star\in\cC_{n,>}^{\all}\) and converges to
\(\theta\) as \(t\downarrow0\). Therefore, $\cC_n^{\all}\subseteq\overline{\cC_{n,>}^{\all}}$.
The reverse inclusion is immediate because \(\cC_n^{\all}\) is closed and contains \(\cC_{n,>}^{\all}\).
Thus Assumption~\ref{ass:closure_strict_consistency_misspecification} holds.

Using Lemma~\ref{lem:gaussian_conversion_error_to_parameter_misspecification}, Assumption~\ref{ass:conversion_error_to_parameter_misspecification} also holds true with explicit constants $\epsilon_{\Sigma,\eta}^{\all}$ and $\alpha_{\Sigma,\eta}^{\all}$ as in Eq.~\eqref{eq:gaussian_conversion_error_to_parameter_misspecification} for the norm $\|\cdot\|_{\Sigma}$.

The Assumptions~\ref{ass:finite_VC_misspecification},~\ref{ass:conversion_error_to_parameter_misspecification}, and~\ref{ass:closure_strict_consistency_misspecification} are shown with constants that are independent of $\theta^\star$. Therefore, the second result on $\cF_{\cN,\Sigma}$ is a corollary of Theorem~\ref{thm:classifier_to_parameter_misspecification}, applied uniformly over $\theta^\star$.

\begin{lemma}\label{lem:gaussian_conversion_error_to_parameter_misspecification}
    Let $\eta >\frac{1}{2}\log((2\pi)^d|\Sigma|)$.
    Let $\epsilon_{\Sigma,\eta}$ and $\kappa_{d-1}$ as in Lemma~\ref{lem:gaussian_conversion_error_to_parameter_identifiability}.
    Define $\epsilon_{\Sigma,\eta}^{\all} \eqdef \frac{\sqrt{2\eta-\log((2\pi)^d|\Sigma|)}}{4d} > 0$ and
    \begin{equation}\label{eq:gaussian_conversion_error_to_parameter_misspecification}
        \alpha_{\Sigma,\eta}^{\all} \eqdef 2\kappa_{d-1}\widetilde\beta_d
    |\Sigma|^{1/2}e^{-\eta}(2\eta - \log((2\pi)^d|\Sigma|))^{\frac{d-1}{2}} \exp\left(-\frac{5}{8d}(2\eta - \log((2\pi)^d|\Sigma|))\right) > 0 \: ,
    \end{equation}
where $\wt\beta_d \eqdef \left(1-\frac{1}{4d^2}\right)^{(d-1)/2}$.
Let $\mathrm{er}_{\mathrm{prof}}$ as in Eq.~\eqref{eq:profiled_classification_error}.
Then, for all $(\theta^\star,\theta)\in \R^d \times \R^d$,
\[
    \mathrm{er}_{\mathrm{prof}}(\theta^\star,\theta) \ge \alpha_{\Sigma,\eta}^{\all} \min\left\{ \|\theta-\theta^\star\|_{\Sigma}, \epsilon_{\Sigma,\eta}^{\all} \right\}\: .
\]
\end{lemma}
\begin{proof}
Let 
\[
    Z
    \eqdef
    \Sigma^{-1/2}X-\Sigma^{1/2}\theta^\star
    \sim\cN(0,I_d),
    \qquad
    v
    \eqdef
    \Sigma^{1/2}(\theta-\theta^\star),
    \qquad
    \rho
    \eqdef
    \|v\|_2
    =
    \|\theta-\theta^\star\|_\Sigma.
\]
By rotational invariance, we may assume that \(v=\rho e_1\).
The true positive region is then \(B(0,c_\eta)\), whereas varying the threshold for \(\theta\) amounts to varying the radius of a ball centered at \(\rho e_1\). 
Consequently,
\[
    \mathrm{er}_{\mathrm{prof}}(\theta^\star,\theta)
    =
    \inf_{R\geq0}
    \gamma_d\left(
        B(0,c_\eta)\triangle B(\rho e_1,R)
    \right),
\]
where \(\gamma_d\) denotes standard Gaussian measure.

We use the following elementary interval inequality: for every \(a>0\), \(b\geq0\), \(\rho\geq0\), and \(h\in(0,a]\),
\begin{equation}\label{eq:lebesgue_lb_interval}
    \mathrm{Leb}\left( \bigl([-a,a]\triangle[\rho-b,\rho+b]\bigr) \cap[-a-h,a+h] \right) \geq \min\{2\rho,h\} \: .
\end{equation}
The inequality also holds when the second interval is empty.
Indeed, when the intervals intersect, the sum of the absolute displacements of their left and right endpoints is at least \(2\rho\); truncation to \([-a-h,a+h]\) can reduce the resulting discrepancy only to \(h\). 
If they are disjoint, the left-hand side is at least \(2a\geq h\).

Set $h \eqdef \frac{c_\eta}{2d} $.
Write \(Z=(U,V)\in\R\times\R^{d-1}\). 
For \(\|v\|_2^2\leq(c_\eta)^2-h^2\), define $a_v\eqdef\sqrt{(c_\eta)^2-\|v\|_2^2}$.
Then \(a_v\geq h\). 
The corresponding sections of the true and candidate balls are respectively
\[
    I_v=[-a_v,a_v] \quad\text{and}\quad J_v=[\rho-b_v,\rho+b_v] \: ,
\]
with \(J_v\) possibly empty. 
Applying Eq.~\eqref{eq:lebesgue_lb_interval}, and using
$h/2=\varepsilon_{\Sigma,\eta}^{\mathrm{all}}$,
gives
\[
    \operatorname{Leb}\left(
    (I_v\triangle J_v)
    \cap[-a_v-h,a_v+h]
    \right)
    \ge
    \min\{2\rho,h\}
    =
    2\min\{\rho,\varepsilon_{\Sigma,\eta}^{\mathrm{all}}\}
    \: .
\]
Since the standard Gaussian density \(\phi\) is symmetric and nonincreasing on \(\R_+\),
\[
    \int_{\R} \phi(u)\indi{u\in I_v\triangle J_v}\,\mathrm du \geq 2\phi(a_v+h)\min\{\rho,
\varepsilon_{\Sigma,\eta}^{\mathrm{all}}
\} \: .
\]
Integrating over
\(\|v\|_2^2\leq(c_\eta)^2-h^2\), and using
\(\|v\|_2^2+a_v^2=(c_\eta)^2\), yields, uniformly over \(R\geq0\),
\begin{align*}
    \gamma_d\left(
        B(0,c_\eta)\triangle B(\rho e_1,R)
    \right) &\ge
    \frac{
        2\kappa_{d-1}\bigl((c_\eta)^2-h^2\bigr)^{(d-1)/2}
    }{(2\pi)^{d/2}}
    \exp\left(
        -\frac{(c_\eta)^2}{2}
        -c_\eta h
        -\frac{h^2}{2}
    \right)
    \min\{\rho,
\varepsilon_{\Sigma,\eta}^{\mathrm{all}}
\} \\
    &\ge 2\kappa_{d-1} c_\eta^{d-1}\widetilde\beta_d |\Sigma|^{1/2}e^{-\eta}  \exp\left(-\frac{5}{8d}c_\eta^2\right) \min\{\rho,
\varepsilon_{\Sigma,\eta}^{\mathrm{all}}
\} \: .
\end{align*}
As the right-hand side is independent of \(R\geq0\), taking the infimum concludes the proof.
\end{proof}


\section{Proofs of Section~\ref{sec:lower_bound}}
\label{app:lower_bound}
In this appendix, we prove Theorem~\ref{thm:lower_bound} from Section~\ref{sec:lower_bound} in Appendix~\ref{proof:lower_bound}.
Appendix~\ref{sub:thresholded_log_likelihood_rewards} contains the proof of the hellinger control for thresholded log-likelihood rewards: Assumption~\ref{ass:Hellinger_control} holds for $\cF_{\cN, \Sigma}$ (Lemma~\ref{lem:gaussian_hellinger_control}), $\cF_{\lap, b}$ (Lemma~\ref{lem:laplace_hellinger_control}) and $\cF_{\Exp}$ (Lemma~\ref{lem:exponential_hellinger_control}).
\paragraph{Proof Strategy.}
Let $\mathbb{P}$ and $\mathbb{Q}$ be two distributions, their the total variation ($\TV$) distance is defined as $ \TV(\mathbb{P},\mathbb{Q}) \eqdef \frac{1}{2} \left\|\bP - \bQ \right\|_1$, their squared Hellinger ($\mH$) distance as $\mH(\bP , \bQ ) \eqdef \frac{1}{2} \|\sqrt{\bP} - \sqrt{\bQ} \|_2^2$, and their Bhattacharyya coefficient ($\BC$) is $\BC(\bP , \bQ ) \eqdef \|\sqrt{\bP  \bQ} \|_1$. 
They satisfy
\[
    \TV(\bP^{\otimes n}, \bQ^{\otimes n}) \le \sqrt{2 n \mH(\bP , \bQ)}\quad \text{and} \quad \mH(\bP , \bQ) = 1 - \BC(\bP , \bQ) \: .
\]
Let $\cD_{0}(\theta, \lambda)\eqdef\{ x \in \cX \mid \sign(r_{\theta}(x)) \ne \sign(r_{\lambda}(x))\}$, and $\BC_{0}(\theta, \lambda) \eqdef \bE_{p_{\theta}} [\indi{\cD_{0}(\theta, \lambda)} \sqrt{p_{\lambda}/p_{\theta}}]$.
Then,
\begin{equation} \label{eq:HellingerDistance}
    \forall(\theta, \lambda) \in \Theta^2, \: \mH(q_{\theta},q_{\lambda}) = \BC_{0}(\theta , \lambda)  +  \mH(p_{\theta}, p_{\lambda}) \: .
\end{equation}
\begin{proof}
By definition, we have
\[
    \forall (x,z)\in \cX\times\{\pm 1, 0\}, \quad \sqrt{q_{\theta}(x,z)q_{\lambda}(x,z)} = \left( 1 - \indi{x \in \cD_{0}(\theta, \lambda)} \right) \sqrt{p_{\theta}(x) p_{\lambda}(x) } \: .
\]
By integration and using that $\mH(\bP , \bQ) = 1 - \BC(\bP , \bQ)$, we obtain
\[
    \BC(q_{\theta},q_{\lambda}) = \BC(p_{\theta},p_{\lambda}) - \BC_{0}(\theta, \lambda) \quad \text{and} \quad \mH(q_{\theta},q_{\lambda}) = \BC_{0}(\theta , \lambda)  +  \mH(p_{\theta}, p_{\lambda})\: .
\]
\end{proof}

\subsection{Proof of Theorem~\ref{thm:lower_bound}}
\label{proof:lower_bound}

Let $\Theta_{1}$ and $\gamma_{\cF}(k),\beta_{\cF}(k),c_{1}$ be as in Assumption~\ref{ass:Hellinger_control}. 
Since $\Theta_{1}$ has nonempty interior, let us choose $\theta_{1} \in \mathrm{int}\left(\Theta_{1}\right)$ and $r_{1} > 0$ such that $B_{2}(\theta_{1},r_{1})\subseteq \Theta_{1}$.
Let $\rho_k \eqdef \min\left\{ \gamma_{\cF}(k), \frac{r_1}{\sqrt{k}} \right\}$. Define 
\[
    \delta_{n,k} \eqdef \frac{1}{8(1+c_1)\sqrt n} \min\left\{\frac{1}{\beta_{\cF}(k)\sqrt n},1\right\} \quad \text{and} \quad n_{k} \eqdef  1 + \left\lfloor \min\left\{ \frac{1}{8(1+c_1)\beta_{\cF}(k)\rho_{k}}, \frac{1}{64(1+c_1)^2\rho_{k}^2} \right\} \right\rfloor \: .
\]
For all $n\ge n_{k}$, we have $\delta_{n,k}\in(0,\rho_k)$ and $\delta_{n,k}\sqrt{k}\in(0,r_1)$. 
Let $n \ge n_{k}$, set $\delta=\delta_{n,k}$ and consider the hypercube
\[
    \Theta_1' \eqdef \{\theta_b\}_{b\in\{0,1\}^k}  \quad \text{where} \quad \forall b\in\{0,1\}^k, \: \theta_b\eqdef \theta_{1} + \delta b  \: .
\]
For every $b\in\{0,1\}^k$,
\[
    \|\theta_b-\theta_1\| =\delta\|b\|_2 \leq\delta\sqrt{k} <r_1 \: .
\]
Consequently, $ \Theta_1' \subseteq B_2(\theta_1,r_1) \subseteq\Theta_1$.
Let $d_H$ denote the Hamming distance. 
Then, for any $b,b'\in\{0,1\}^k$,
\[
    \|\theta_b-\theta_{b'}\| = \delta \|b-b'\|_2 = \delta \sqrt{d_H(b,b')} \ge \delta k^{-1/2} d_H(b,b') \:.
\]
Now suppose $d_H(b,b')=1$. 
Then, $\|\theta_b-\theta_{b'}\|=\delta<\gamma_{\cF}(k)$, so Assumption~\ref{ass:Hellinger_control} applies. 
Combining with Eq.~\eqref{eq:HellingerDistance} gives
\[
    \max_{d_H(b,b')=1} \TV\bigl(q_{\theta_b}^{\otimes n},q_{\theta_{b'}}^{\otimes n}\bigr) \le \sqrt{ 2n\left( \beta_{\mathcal F}(k)\delta + c_1\delta^2 \right)} \: .
\]
By the choice $\delta=\delta_{n,k}$, the right-hand side is bounded by $1/2$. 
Therefore, by Assouad's lemma~\citep{tsybakov2009nonparametric},
\[
    \inf_{\widehat \theta_n : (\cX \times \{\pm 1, 0\})^{n} \to \Theta} \sup_{\theta^\star\in\Theta_1} \bE_{q_{\theta^\star}^{\otimes n}} \left[\|\widehat\theta_n-\theta^\star\|\right] \ge \frac{\delta_{n,k}\sqrt k}{8} = \frac{\sqrt k}{64(1+c_1)\sqrt n} \min\left\{\frac{1}{\beta_{\cF}(k)\sqrt n},1\right\}\: .
\]
This concludes the proof.

\subsection{Thresholded Log-likelihood Rewards}
\label{sub:thresholded_log_likelihood_rewards}

Assumption~\ref{ass:Hellinger_control} holds for $\cF_{\cN, \Sigma}$ (Lemma~\ref{lem:gaussian_hellinger_control}).
\begin{lemma}\label{lem:gaussian_hellinger_control}
    Let $\eta>\frac12\log\bigl((2\pi)^d|\Sigma|\bigr)$, $\gamma_{\Sigma,\eta} \eqdef \sqrt{2\eta-\log\bigl((2\pi)^d|\Sigma|\bigr)}$ and $\kappa_{d-1} \eqdef \frac{\pi^{(d-1)/2}}{\Gamma\left(\frac{d+1}{2}\right)}$.
    For \(d\ge3\), define
    \[
        M_{d-1}
        \eqdef
        \left\|f_{\chi^2(d-1)}\right\|_\infty
        =
        \frac{
            \left(\frac{d-3}{2}\right)^{(d-3)/2}
            e^{-(d-3)/2}
        }{
            2\Gamma\left(\frac{d-1}{2}\right)
        },
    \]
    with the convention \(0^0=1\), and define
    \[
        \beta_{\Sigma,\eta}^{\mathrm{G}}
        \eqdef
        \begin{cases}
            \displaystyle
            \sqrt{\frac{2}{\pi}},
            & d=1,
            \\[2ex]
            \displaystyle
            \min\left\{
                \frac{2\gamma_{\Sigma,\eta}}{\pi},
                \sqrt{\frac{2}{\pi}}
            \right\},
            & d=2,
            \\[3ex]
            \displaystyle
            \min\left\{
                \frac{
                    2\kappa_{d-1}\gamma_{\Sigma,\eta}^{d-1}
                }{
                    (2\pi)^{d/2}
                },
                2\sqrt{\frac{2}{\pi}}\,M_{d-1}
            \right\},
            & d\ge3.
        \end{cases}
    \]
    Then, for every \((\theta^\star,\theta)\in\R^d\times\R^d\) such that $0<\|\theta-\theta^\star\|_\Sigma <\gamma_{\Sigma,\eta}$, we have
    \[
        \mH(q_{\theta^\star},q_{\theta}) = \BC_0(\theta^\star,\theta)
        +\mH(p_{\theta^\star},p_\theta)
        \le
        \beta_{\Sigma,\eta}
        \|\theta-\theta^\star\|_\Sigma
        +\frac18\|\theta-\theta^\star\|_\Sigma^2.
    \]
\end{lemma}

\begin{proof}
    Let
    \[
        R\eqdef R_{\Sigma,\eta},
        \qquad
        \delta\eqdef\|\theta-\theta^\star\|_\Sigma,
        \qquad
        m\eqdef\frac{\theta+\theta^\star}{2},
        \qquad
        u\eqdef
        \frac{\Sigma^{1/2}(\theta-\theta^\star)}{\delta}.
    \]
    The set on which either reward vanishes has
    \(p_{\theta^\star}\)-measure zero. Moreover,
    \[
        \sqrt{p_{\theta^\star}(x)p_\theta(x)}
        =
        e^{-\delta^2/8}p_m(x).
    \]
    Hence, using the same change of variables as in the proof of
    Lemma~\ref{lem:gaussian_conversion_error_to_parameter_identifiability},
    \[
        \BC_0(\theta^\star,\theta)
        =
        e^{-\delta^2/8}
        \bP\left(
            Z\in
            B_2\left(\frac{\delta}{2}u,R\right)
            \triangle
            B_2\left(-\frac{\delta}{2}u,R\right)
        \right),
        \qquad
        Z\sim\cN(0,I_d).
        \tag{1}
    \]

    We first recall the volume bound. Since \(\delta<R\), every point
    in the symmetric difference in~\((1)\) has norm at least
    \(R-\delta/2\). Furthermore,
    \[
    \begin{aligned}
        \operatorname{Leb}\left(
            B_2\left(\frac{\delta}{2}u,R\right)
            \triangle
            B_2\left(-\frac{\delta}{2}u,R\right)
        \right)
        &=
        2\kappa_{d-1}
        \int_{-\delta/2}^{\delta/2}
            (R^2-t^2)^{(d-1)/2}\,\mathrm dt\le
        2\kappa_{d-1}R^{d-1}\delta.
    \end{aligned}
    \]
    Consequently,
    \begin{align*}
        \BC_0(\theta^\star,\theta)
        &\le
        \frac{2\kappa_{d-1}R^{d-1}}{(2\pi)^{d/2}}
        \exp\left(
            -\frac{\delta^2}{8}
            -\frac{(R-\delta/2)^2}{2}
        \right)\delta
        \le
        \frac{2\kappa_{d-1}R^{d-1}}{(2\pi)^{d/2}}
        \delta,
    \end{align*}
    where the last inequality follows from \(\delta<R\).

    We next derive dimension-dependent slicing bounds. Write
    \[
        T\eqdef\langle Z,u\rangle,
        \qquad
        Y\eqdef Z-Tu.
    \]
    Then \(T\sim\cN(0,1)\), \(Y\sim\cN(0,I_{d-1})\), and \(T\)
    and \(Y\) are independent.

    Conditionally on \(Y=y\), membership in the two balls in~\((1)\)
    is determined by membership of \(T\) in two intervals of equal
    length whose centers are separated by \(\delta\). Their symmetric
    difference has Lebesgue measure at most \(2\delta\). Therefore,
    using \(\|\phi\|_\infty=(2\pi)^{-1/2}\),
    \[
        \bP\left(
            Z\in
            B_2\left(\frac{\delta}{2}u,R\right)
            \triangle
            B_2\left(-\frac{\delta}{2}u,R\right)
        \right)
        \le
        \sqrt{\frac{2}{\pi}}\,\delta.
        \tag{2}
    \]
    Together with~\((1)\), this proves
    \[
        \BC_0(\theta^\star,\theta)
        \le
        \sqrt{\frac{2}{\pi}}\,\delta.
        \tag{3}
    \]
    For \(d=1\), the volume bound and~\((3)\) coincide. For \(d=2\),
    the volume bound is \(2R\delta/\pi\); taking the minimum with
    \((3)\) gives the claimed constant.

    Suppose now that \(d\ge3\). Let
    \[
        V\eqdef\|Y\|_2^2\sim\chi^2(d-1).
    \]
    Conditionally on \(T=t\), membership in the two balls is determined
    by
    \[
        V\le R^2-\left(t-\frac{\delta}{2}\right)^2
        \quad\text{and}\quad
        V\le R^2-\left(t+\frac{\delta}{2}\right)^2.
    \]
    If these indicators differ, then \(V\) belongs to an interval of
    length at most
    \[
        \left|
            \left(t+\frac{\delta}{2}\right)^2
            -
            \left(t-\frac{\delta}{2}\right)^2
        \right|
        =
        2\delta|t|.
    \]
    Since the density of \(V\) is bounded by \(M_{d-1}\),
    \[
    \begin{aligned}
        \bP\left(
            Z\in
            B_2\left(\frac{\delta}{2}u,R\right)
            \triangle
            B_2\left(-\frac{\delta}{2}u,R\right)
        \right)
        &\le
        2\delta M_{d-1}\bE[|T|]=
        2\sqrt{\frac{2}{\pi}}\,M_{d-1}\delta.
    \end{aligned}
    \]
    Combining this inequality with the volume bound and~\((1)\)
    yields
    \[
        \BC_0(\theta^\star,\theta)
        \le
        \beta_{\Sigma,\eta}^{\mathrm{G}}\delta.
    \]

    Finally,
    \[
        \mH(p_{\theta^\star},p_\theta)
        =
        1-\exp\left(-\frac{\delta^2}{8}\right)
        \le\frac{\delta^2}{8}.
    \]
    Combining the last two inequalities concludes the proof.
\end{proof}

Assumption~\ref{ass:Hellinger_control} holds for $\cF_{\lap, b}$ (Lemma~\ref{lem:laplace_hellinger_control}).
\begin{lemma}\label{lem:laplace_hellinger_control}
    Let $\eta > \log(2b)$.
    Define
    \begin{equation} \label{eq:laplace_hellinger_control}
        \gamma_{b,\eta} \eqdef b \min\{\eta- \log(2b), 2\}\quad ,\quad  \beta_{b,\eta} \eqdef 2 e^{-\eta} \quad \text{and}\quad c_{b,\eta} \eqdef \frac{1}{4b^2}+ \frac{e^{-\eta}}{b} \: .
    \end{equation}
    Then, for every $(\theta^\star,\theta)\in\R \times\R $ such that $0<|\theta-\theta^\star|< \gamma_{b,\eta}$,
\[
    \mH(q_{\theta^\star},q_{\theta}) = \BC_0(\theta^\star,\theta) + \mH(p_{\theta^\star},p_\theta) \le \beta_{b,\eta}|\theta^\star-\theta|  + c_{b,\eta}|\theta^\star-\theta|^2 \: .
\]
\end{lemma}
\begin{proof}
Let \(\Delta\eqdef|\theta-\lambda|\), \(a\eqdef\Delta/b\) and $\bar \eta =\eta- \log(2b)>0$.
The squared Hellinger distance between two Laplace distributions with common scale \(b\) satisfies
\[
    \mH(p_\theta,p_\lambda)
    =
    1-\left(1+\frac{a}{2}\right)e^{-a/2}
    \leq \frac{a^2}{4},
\]
where the inequality follows from \(x^2-1+(1+x)e^{-x}\geq0\) for \(x\geq0\).

Since the zero sets of \(r_\theta\) and \(r_\lambda\) have zero Lebesgue measure, $\BC_0(\theta,\lambda)=\int_{\cD(\theta,\lambda)} \sqrt{p_\theta(x)p_\lambda(x)} \,\diff x$.
By the same interval decomposition used in the proof of Lemma~\ref{lem:laplace_conversion_error_to_parameter_identifiability}, and using translation invariance and symmetry, for
\(0<\Delta<b\bar\eta\) we have
\[
    \cD(\theta,\lambda)
    =
    (\theta-b\bar\eta,\lambda-b\bar\eta)
    \cup
    (\theta+b\bar\eta,\lambda+b\bar\eta),
\]
after assuming without loss of generality that \(\theta<\lambda\).
Integrating the Hellinger affinity over these two intervals gives
\[
    \BC_0(\theta,\lambda) = 4b e^{-\eta}\sinh\left(\frac{\Delta}{2b}\right) \: .
\]
For \(x\in[0,1]\), \(\sinh(x)\leq x+x^2\). 
Hence, whenever \(\Delta<b\min\{\bar \eta,2\}\),
\[
\begin{aligned}
    \BC_0(\theta,\lambda)
    &\leq
    4b e^{-\eta}
    \left(
        \frac{\Delta}{2b}
        +
        \frac{\Delta^2}{4b^2}
    \right)=
    2 e^{-\eta} \Delta   + e^{-\eta}
        \frac{\Delta^2}{b}.
\end{aligned}
\]
Combining this inequality with the Hellinger bound concludes the proof.
\end{proof}

Assumption~\ref{ass:Hellinger_control} holds for $\cF_{\Exp}$ (Lemma~\ref{lem:exponential_hellinger_control}).
\begin{lemma} \label{lem:exponential_hellinger_control}
Let $\eta\in \R$.
Define
\begin{equation}\label{eq:exponential_hellinger_control}
    \gamma_{\eta} \eqdef\frac{e^{-\eta}}2,\qquad
    \beta_{\eta}\eqdef2e^\eta,\qquad
    c_{\eta}\eqdef\frac{4e^{2\eta}}{9} \: .
\end{equation}
Then, for every $(\theta^\star,\theta)\in(e^{-\eta},+\infty)\times(0,+\infty)$ such that $0<|\theta-\theta^\star|<\gamma_{\eta}$, we have
\[
    \mH(q_{\theta^\star},q_{\theta})  = \BC_0(\theta^\star,\theta) +\mH(p_{\theta^\star},p_\theta) \leq \beta_{\eta}|\theta-\theta^\star| +c_{\eta}|\theta-\theta^\star|^2 \: .
\]
\end{lemma}

\begin{proof}
Set
\[
    a\eqdef\theta^\star,
    \qquad
    g_\eta(y)\eqdef\bigl(f_\eta(y)\bigr)_+,
    \qquad
    f_\eta(y)\eqdef\frac{\eta+\log y}{y}.
\]
Since $a>e^{-\eta}$ and
$|\theta-a|<e^{-\eta}/2<a/2$, we have
\[
    \theta\in\left[\frac a2,\frac{3a}{2}\right].
\]
By Lemma~\ref{lem:init_properties}, up to a null set,
\[
    \cD(a,\theta)
    =
    \left(
        \min\{f_\eta(a),g_\eta(\theta)\},
        \max\{f_\eta(a),g_\eta(\theta)\}
    \right).
\]
Consequently,
\[
    \BC_0(a,\theta)
    =
    \frac{2\sqrt{a\theta}}{a+\theta}
    \left|
        A(\theta)-B(\theta)
    \right|,
\]
where
\[
    A(y)\eqdef
    \exp\left(-\frac{a+y}{2}f_\eta(a)\right),
    \qquad
    B(y)\eqdef
    \exp\left(-\frac{a+y}{2}g_\eta(y)\right).
\]
Moreover, $A(a)=B(a)$.

Let $z\eqdef\eta+\log a>0$. For
$y\in[a/2,3a/2]$,
\[
    |A'(y)|
    =
    \frac{z}{2a}
    \exp\left(-\frac{a+y}{2a}z\right)
    =
    \frac{e^\eta z}{2}
    \exp\left(-\frac{3+y/a}{2}z\right)
    \leq \frac{e^\eta}{2}.
\]
The function $B$ is absolutely continuous. If
$q\eqdef\eta+\log y\leq0$, then $B'(y)=0$ almost everywhere.
If $q>0$, writing
\[
    \phi(y)\eqdef\frac{a+y}{2}f_\eta(y)
    =\frac{a+y}{2y}q,
\]
we have, since $a/y\in[2/3,2]$,
\[
    |\phi'(y)|
    =
    \frac{1}{2y}
    \left|1+\frac ay(1-q)\right|
    \leq
    \frac{3/2+q}{y},
    \qquad
    \phi(y)\geq\frac56q.
\]
Hence
\[
    |B'(y)|
    =
    e^{-\phi(y)}|\phi'(y)|
    \leq
    e^\eta\left(\frac32+q\right)e^{-11q/6}
    \leq\frac{3e^\eta}{2}.
\]
Thus $|(A-B)'(y)|\leq2e^\eta$ almost everywhere on
$[a/2,3a/2]$. Since $A(a)=B(a)$,
\[
    |A(\theta)-B(\theta)|
    \leq2e^\eta|\theta-a|.
\]
Using $2\sqrt{a\theta}/(a+\theta)\leq1$, we obtain
\[
    \BC_0(a,\theta)
    \leq2e^\eta|\theta-a|.
\]

Finally,
\[
    \mH(p_a,p_\theta)
    =
    \frac{(\sqrt a-\sqrt\theta)^2}{a+\theta}
    \leq
    \frac{(a-\theta)^2}{(a+\theta)^2}.
\]
Since $\theta\geq a/2$ and $a>e^{-\eta}$,
\[
    a+\theta\geq\frac{3a}{2}>\frac{3e^{-\eta}}2,
\]
and therefore
\[
    \mH(p_a,p_\theta) \leq \frac{4e^{2\eta}}9(a-\theta)^2.
\]
Combining the two bounds concludes the proof.
\end{proof}


\section{Quality-based M-estimators}
\label{app:mestimator}
In this appendix, we discuss the asymptotic behavior of $\sqe$ and $\lle$ from Section~\ref{ssec:QbasedMestimator} in greater depth.
Lemma~\ref{lem:sp_mle_asymptotic_variance} computes the asymptotic variance of $\sqe$; its proof is deferred to Appendix~\ref{proof:sp_mle_asymptotic_variance}.
Lemma~\ref{lem:asymptotic_variance_lle} computes the asymptotic variance of $\lle$; its proof is deferred to Appendix~\ref{proof:asymptotic_variance_lle}.
For thresholded log-likelihood rewards, Appendix~\ref{sub:multivariate_gaussian_distribution_with_known_covariance} discusses the case $\cF_{\cN, \Sigma}$, Appendix~\ref{sub:laplace_distribution_with_known_scale} the case $\cF_{\lap, b}$, and Appendix~\ref{sub:exponential_distribution} the case $\cF_{\Exp}$.
\paragraph{Stochastic Qualities.}
Let $M_{\sto}(\theta) \eqdef \bE_{(X,Z) \sim q_{\theta^\star}^{\sto}}[\log q_{\theta}^{\sto}(X,Z)]$.
Under sufficient regularity, by swapping the integration and the differentiation operators, we obtain
    \[
        \nabla M_{\sto}(\theta^\star)  = \bE_{q_{\theta^\star}^{\sto}}[\nabla \log q_{\theta^\star}^{\sto}] = \nabla 1 = 0_k \: .
    \]
This holds $\cF_{\cN, \Sigma}$, $\cF_{\lap, b}$ and $\cF_{\Exp}$ for thresholded log-likelihood rewards.

Lemma~\ref{lem:sp_mle_asymptotic_variance} compares the efficiency of $\sqe$ and $\so$ in terms of asymptotic variance, namely $V_{\theta^\star,\sto} \eqdef \cI(q_{\theta^\star}^{\sto})^{-1}$ versus $\cI(p_{\theta^\star})^{-1}$.
\begin{lemma} \label{lem:sp_mle_asymptotic_variance}
    Let $\Delta^{\sqe}_{\theta^\star} \eqdef \bE_{p_{\theta^\star}}[\sigma(r_{\theta^\star}) \sigma(-r_{\theta^\star}) \nabla_{\theta^\star} r_{\theta^\star}  \nabla_{\theta^\star} r_{\theta^\star}\transpose]$.
    Then, $\cI(q_{\theta^\star}^{\sto}) = \cI(p_{\theta^\star}) + \Delta^{\sqe}_{\theta^\star}$.
    The p.s.d. matrix $\Delta^{\sqe}_{\theta^\star}$ is definite if $\bP_{p_{\theta^\star}}(|\langle u, \nabla_{\theta^\star} r_{\theta^\star} \rangle| > 0) > 0$ for all $u \in \cS_{k-1}$.
\end{lemma}
Lemma~\ref{lem:sp_mle_asymptotic_variance} shows that $\cI(q_{\theta^\star}^{\sto}) \succeq \cI(p_{\theta^\star})$ and exhibits a mild condition under which $\wh \theta^{\sqe}_{n}$ is asymptotically better than $\wh \theta^{\so}_{n}$, meaning that incorporating quality data improves asymptotic efficiency.
The condition in Lemma~\ref{lem:sp_mle_asymptotic_variance} ensures that $\nabla_{\theta^\star} r_{\theta^\star}$ spans all directions with some probability, making the quality-based estimator asymptotically superior to the sample-only MLE.
For $r_{\theta^\star} = \log p_{\theta^\star} +\eta$, this condition holds for $\cF_{\cN, \Sigma}$ (Appendix~\ref{sub:multivariate_gaussian_distribution_with_known_covariance}), $\cF_{\lap, b}$ (Appendix~\ref{sub:laplace_distribution_with_known_scale}) and $\cF_{\Exp}$ (Appendix~\ref{sub:exponential_distribution}).
\paragraph{Deterministic Qualities.}
Let $M_{\det}(\theta) \eqdef \bE_{X \sim p_{\theta^\star}}[\log q_{\theta}^{\sto}(X,\sign(r_{\theta^\star}(X)))]$.
Under sufficient regularity, we obtain
\begin{equation} \label{eq:consistency_lle}
     \bE_{p_{\theta^\star}}[\sign(r_{\theta^\star}) \sigma(-|r_{\theta^\star}|) \nabla r_{\theta^\star} ] = 0_k \quad \implies \quad \nabla M_{\det}(\theta^\star)  =  0_k \: ,
\end{equation}
For $r_{\theta^\star} = \log p_{\theta^\star} +\eta$, this condition holds for $\cF_{\cN, \Sigma}$ (Appendix~\ref{sub:multivariate_gaussian_distribution_with_known_covariance}) and $\cF_{\lap, b}$ (Appendix~\ref{sub:laplace_distribution_with_known_scale}), but not for $\cF_{\Exp}$ (Appendix~\ref{sub:exponential_distribution}).

Lemma~\ref{lem:asymptotic_variance_lle} computes the asymptotic variance of $\lle$, namely $V_{\theta^\star, \det} $.
\begin{lemma}\label{lem:asymptotic_variance_lle}
    Define
    \begin{align*}
        u_{\theta^\star}
        &\eqdef \sign(r_{\theta^\star}) \sigma(-|r_{\theta^\star}|), \\
        M_{\theta^\star}
        &\eqdef -\nabla_{\theta^\star}\log p_{\theta^\star}
        \bigl(\nabla_{\theta^\star} r_{\theta^\star}\bigr)\transpose .
    \end{align*}
    Let
    \begin{align*}
        H^{\lle}_{\theta^\star}
        &\eqdef
        \bE_{p_{\theta^\star}}
        \left[
            u_{\theta^\star}
            \nabla_{\theta^\star}^2 r_{\theta^\star}
        \right], \\
        \Delta^{\lle}_{\theta^\star}
        &\eqdef
        \bE_{p_{\theta^\star}}
        \left[
            (2\sigma(|r_{\theta^\star}|)-1)
            \sigma(-|r_{\theta^\star}|)
            \nabla_{\theta^\star} r_{\theta^\star}
            \nabla_{\theta^\star} r_{\theta^\star}\transpose
        \right], \\
        R^{\lle}_{\theta^\star}
        &\eqdef
        \bE_{p_{\theta^\star}}
        \left[
            u_{\theta^\star}
            \left(
                M_{\theta^\star}
                +
                M_{\theta^\star}\transpose
            \right)
        \right].
    \end{align*}
    Then
    \[
        V_{\theta^\star,\det}
        \eqdef
        V_{1,\theta^\star}^{-1}
        V_{2,\theta^\star}
        V_{1,\theta^\star}^{-1},
    \]
    where
    \begin{align*}
        V_{1,\theta^\star}
        &=
        \cI(q_{\theta^\star}^{\sto})
        -
        H^{\lle}_{\theta^\star}, \\
        V_{2,\theta^\star}
        &=
        \cI(q_{\theta^\star}^{\sto})
        -
        \Delta^{\lle}_{\theta^\star}
        -
        R^{\lle}_{\theta^\star}.
    \end{align*}
\end{lemma}
The equation~\eqref{eq:consistency_lle} and the conditions $V_{\theta^\star, \det} \prec \cI(p_{\theta^\star})^{-1}$ and $V_{\theta^\star, \det} \prec V_{\theta^\star,\sto}$ depend on the geometry of $r_{\theta^\star}$ and $p_{\theta^\star}$. 
For $r_{\theta^\star} = \log p_{\theta^\star} +\eta$, these conditions are discussed in Appendix~\ref{sub:multivariate_gaussian_distribution_with_known_covariance} for $\cF_{\cN, \Sigma}$ and Appendix~\ref{sub:laplace_distribution_with_known_scale} for $\cF_{\lap, b}$.
\subsection{Proof of Lemma~\ref{lem:sp_mle_asymptotic_variance}}
\label{proof:sp_mle_asymptotic_variance}
Under sufficient regularity, by swapping the integration and the differentiation operators, we have 
    \[
    \bE_{q_{\theta^\star}^{\sto}}[-\nabla^2 \log q_{\theta^\star}^{\sto}] = \bE_{q_{\theta^\star}^{\sto}}[\nabla \log q_{\theta^\star}^{\sto} (\nabla \log q_{\theta^\star}^{\sto})\transpose] \: .
    \]
Direct computation yields that
    \begin{align*} 
        \nabla_{\theta^\star} \log q_{\theta^\star}^{\sto}(x,z) &= \nabla_{\theta^\star} \log  p_{\theta^\star}(x) +  z  \sigma (-z r_{\theta^\star}(x) )  \nabla_{\theta^\star}r_{\theta^\star}(x) \: , \\
        \nabla_{\theta^\star}^2 \log q_{\theta^\star}^{\sto}(x,z) &= \nabla_{\theta^\star}^2 \log p_{\theta^\star}(x)  -  \sigma ( r_{\theta^\star}(x) ) \sigma ( -r_{\theta^\star}(x) )   \nabla_{\theta^\star}r_{\theta^\star}(x) \nabla_{\theta^\star}r_{\theta^\star}(x)\transpose \\
        &\qquad \qquad +  z \sigma (-z r_{\theta^\star}(x) )  \nabla_{\theta^\star}^2r_{\theta^\star}(x) \: .
    \end{align*}
where we used that $g'(x) = \sigma(-x)$ and $g''(x) = -\sigma' (-x) = -\sigma (x)\sigma (-x)$ with $g(x) = \log \sigma(x)$.
Then, 
\begin{align*}
    &\bE_{Z \mid X} \left[ Z \sigma (-Z r_{\theta^\star}(X) )  \nabla_{\theta^\star}^2 r_{\theta^\star}(X) \right] \\
    &= \left(   \sigma (- r_{\theta^\star}(X) )  \sigma ( r_{\theta^\star}(X) )   - \sigma ( r_{\theta^\star}(X) ) \sigma ( -r_{\theta^\star}(X) )  \right)  \nabla_{\theta^\star}^2r_{\theta^\star}(X) =  0_{k \times k} \: .
\end{align*}
Therefore, we have $\cI(q_{\theta^\star}^{\sto}) = \cI(p_{\theta^\star}) + \Delta^{\sqe}_{\theta^\star}$.
For all $x \in \R^k$, we have 
\[
    x\transpose \Delta^{\sqe}_{\theta^\star} x = \|x\|^2 \bE_{p_{\theta^\star}}[\sigma(r_{\theta^\star}) \sigma(-r_{\theta^\star}) \langle x/\|x\|, \nabla_{\theta^\star} r_{\theta^\star} \rangle^2] \ge 0 \: .
\]
It is direct to see that this inequality is strict except if $\bP_{p_{\theta^\star}}(\langle x/\|x\|, \nabla_{\theta^\star} r_{\theta^\star} \rangle^2 = 0) = 1$.
Therefore, $\Delta^{\sqe}_{\theta^\star}$ is a positive definite matrix if $\bP_{p_{\theta^\star}}(|\langle u, \nabla_{\theta^\star}  r_{\theta^\star} \rangle| > 0) > 0$ for all $u \in \cS_{k-1}$.
\subsection{Proof of Lemma~\ref{lem:asymptotic_variance_lle}}
\label{proof:asymptotic_variance_lle}
Under sufficient regularity, by swapping the integration and the differentiation operators, we obtain Eq.~\eqref{eq:consistency_lle} since $\bE_{p_{\theta^\star}}[\nabla_{\theta^\star} \log p_{\theta^\star}] = 0_k$.
    \[
        \bE_{p_{\theta^\star}}[\nabla_{\theta^\star} \log p_{\theta^\star}] = 0_k \quad \text{and} \quad \nabla_{\theta^\star} M_{\det}(\theta^\star) = \bE_{p_{\theta^\star}}[\sign(r_{\theta^\star}(x)) \sigma( - |r_{\theta^\star}(x)|) \nabla_{\theta^\star} r_{\theta^\star}(x)  ] \: .
    \]
Provided Eq.~\eqref{eq:consistency_lle} holds, Under sufficient regularity, M-estimation theory yields that
\[
     \sqrt{n}(\wh \theta^{\lle}_{n}  - \theta^\star ) \rightsquigarrow_{n \to + \infty} \cN(0_{k}, V_{\theta^\star,\det}) \: ,
\]
where $V_{\theta^\star, \det} \eqdef V_{1,\theta^\star}^{-1} V_{2,\theta^\star} V_{1,\theta^\star}^{-1}$ with
\begin{align*}
    V_{1,\theta^\star} &= \bE_{X \sim p_{\theta^\star}}\left[-\nabla_{\theta^\star}^2 \log q_{\theta^\star}^{\sto}(X,\sign(r_{\theta^{\star}}(X))) \right] \: ,\\
    V_{2,\theta^\star} &= \bE_{X \sim p_{\theta^\star}}\left[\nabla_{\theta^\star} \log q_{\theta^\star}^{\sto}(X,\sign(r_{\theta^{\star}}(X))) \nabla_{\theta^\star} \log q_{\theta^\star}^{\sto}(X,\sign(r_{\theta^{\star}}(x)))\transpose \right] \: .
\end{align*}
Combining $z = \sign(r_{\theta^{\star}}(x))$ with the same manipulation as above yields
\begin{align*} 
    &\nabla_{\theta^\star} \log q_{\theta^\star}^{\sto}(x,z) = \nabla_{\theta^\star} \log  p_{\theta^\star}(x) +   \sign(r_{\theta^{\star}}(x))  \sigma (-| r_{\theta^\star}(x) |)  \nabla_{\theta^\star}r_{\theta^\star}(x) \: , \\
    &\nabla_{\theta^\star} \log q_{\theta^\star}^{\sto}(x,z) \nabla_{\theta^\star} \log q_{\theta^\star}^{\sto}(x,z)\transpose = \nabla_{\theta^\star} \log p_{\theta^\star}(x) \nabla_{\theta^\star} \log p_{\theta^\star}(x)\transpose  \\
    &\qquad +  \sigma (-| r_{\theta^\star}(x) |)^2  \nabla_{\theta^\star}r_{\theta^\star}(x)  \nabla_{\theta^\star}r_{\theta^\star}(x) \transpose \\
    &\qquad  +  \sign(r_{\theta^{\star}}(x))  \sigma (-| r_{\theta^\star}(x) |) \left(  \nabla_{\theta^\star}\log p_{\theta^\star}(x) \nabla_{\theta^\star}r_{\theta^\star}(x)\transpose  +   \nabla_{\theta^\star}r_{\theta^\star}(x)  \nabla_{\theta^\star}\log p_{\theta^\star}(x)\transpose  \right) \\
    &\nabla_{\theta^\star}^2 \log q_{\theta^\star}^{\sto}(x,z) = \nabla_{\theta^\star}^2 \log p_{\theta^\star}(x)  -  \sigma ( r_{\theta^\star}(x) ) \sigma ( -r_{\theta^\star}(x) )   \nabla_{\theta^\star}r_{\theta^\star}(x) \nabla_{\theta^\star}r_{\theta^\star}(x)\transpose \\
    &\qquad \qquad +  \sign(r_{\theta^{\star}}(x)) \sigma (-| r_{\theta^\star}(x)| )  \nabla_{\theta^\star}^2r_{\theta^\star}(x) \: .
\end{align*}
where we used that $g'(x) = \sigma(-x)$ and $g''(x) = -\sigma' (-x) = -\sigma (x)\sigma (-x)$ with $g(x) = \log \sigma(x)$.
Using that $ \sigma (-| r_{\theta^\star}(x) |)^2 =  \sigma (-| r_{\theta^\star}(x) |) - \sigma (- r_{\theta^\star}(x) ) \sigma ( r_{\theta^\star}(x) ) $, we have
\[
    V_{1,\theta^\star} = \cI(p_{\theta^\star}) + \Delta^{\sqe}_{\theta^\star} - H^{\lle}_{\theta^\star} \quad \text{and} \quad V_{2,\theta^\star} = \cI(q_{\theta^\star}^{\sto})  - \Delta^{\lle}_{\theta^\star} - R^{\lle}_{\theta^\star} 
\]
where $\Delta^{\sqe}_{\theta} = \bE_{p_{\theta^\star}}[\sigma(r_{\theta}) \sigma(-r_{\theta}) \nabla_{\theta} r_{\theta} ( \nabla_{\theta} r_{\theta})\transpose]$ as in Lemma~\ref{lem:sp_mle_asymptotic_variance}.
This concludes the proof.

\subsection{Multivariate gaussian distribution with known covariance}
\label{sub:multivariate_gaussian_distribution_with_known_covariance}

Let $\bar \eta = \eta - \eta_0$ with $\eta_0 \eqdef \frac{1}{2}\log ((2\pi)^d|\Sigma|)$ where $\eta \in \mathrm{H}_0 = (\eta_0,+\infty)$, hence $\bar \eta > 0$.
\paragraph{Consistency of $\sqe$}
This is a direct consequence of the regularity property of $p_{\theta}$ and $\sigma$.
\paragraph{Asymptotic Variance of $\sqe$}
Let $\Delta^{\sqe}_{\theta^\star}$ as in Lemma~\ref{lem:sp_mle_asymptotic_variance}.
Let $S = \Sigma^{-1/2}(\Sigma \theta^\star - X)$ with $X \sim p_{\theta^\star}$. 
Then, we have $S \sim \cN(0_d,I_{d})$, $r_{\theta^{\star}}(X) = -\frac{1}{2} \| S \|^2 + \bar \eta$ and $\nabla_{\theta^\star} r_{\theta^\star}(X)= - \Sigma^{1/2}S$.
Since $\sigma(t)\sigma(-t)=\sigma'(t)$, we obtain
\begin{align*}
    \Delta^{\sqe}_{\theta^\star} = \Sigma^{1/2} \Delta^{\sqe}_{\cN}(\bar \eta)  \Sigma^{1/2}\quad \text{with} \quad \Delta^{\sqe}_{\cN}(\bar \eta) = \bE_{\cN(0_d,I_{d})}\left[ \sigma'\left(\bar \eta-\frac{\|S\|^2}{2}\right) SS^\top \right] \: .
\end{align*}
Since the law of $S$ is rotationally invariant and the weight $\sigma'\left(\bar \eta-\frac{\|S\|^2}{2}\right)$ depends only on $\|S\|$, the matrix
$\Delta^{\sqe}_{\cN}(\bar \eta)$ commutes with every orthogonal matrix.
Thus it must be a scalar multiple of the identity.
Therefore, we have $\Delta^{\sqe}_{\cN}(\bar \eta)=\alpha_d^{\sqe}(\bar \eta) I_d$ where
\[
    \alpha_d^{\sqe}(\bar \eta) \eqdef \frac{1}{d} \bE_{\cN(0_d,I_{d})}\left[ \sigma'\left(\bar \eta-\frac{\|S\|^2}{2}\right) \|S\|^2 \right]  > 0 \: ,
\]
by taking the trace of the equality.
Since $\cI(q_{\theta^\star}^{\sto})^{-1} = (1 + \alpha_d^{\sqe}(\bar \eta))^{-1} \Sigma^{-1}$, the asymptotic variance of $\wh \theta^{\sqe}_{n}$ is smaller than $\wh \theta^{\so}_{n}$.
\paragraph{Consistency of $\lle$}
Let
\[
    g(r) \eqdef \sign\!\left(-\frac12 r^2+\bar\eta\right) \sigma\!\left(-\left|\frac12 r^2-\bar\eta\right|\right) \: .
\]
Then \(g(\|S\|)S\) is an odd function of \(S\), since \(g(\|-S\|)=g(\|S\|)\). 
As \(S\sim \cN(0_d,I_d)\) is symmetric, \(S\stackrel{d}{=}-S\), and therefore
\[
    \bE[g(\|S\|)S] = \bE[g(\|-S\|)(-S)] = -\bE[g(\|S\|)S] \: .
\]
Since $\bE[g(\|S\|)S]=0_d$, we have $\nabla M_{\det}(\theta^\star) = 0$ is satisfied, hence $\lle$ is consistent.
\paragraph{Asymptotic Variance of $\lle$}
Let $(H^{\lle}_{\theta^\star}, \Delta^{\lle}_{\theta^\star}, R^{\lle}_{\theta^\star})$ as in Lemma~\ref{lem:asymptotic_variance_lle}.
As above, we have
\begin{align*}
    &H^{\lle}_{\theta^\star} = \Sigma H^{\lle}_{\cN}(\bar \eta) \: , \\
    &\Delta^{\lle}_{\theta^\star} = \Sigma^{1/2} \Delta^{\lle}_{\cN}(\bar \eta)  \Sigma^{1/2} \: , \\ 
    &R^{\lle}_{\theta^\star} = \Sigma^{1/2} R^{\lle}_{\cN}(\bar \eta)  \Sigma^{1/2} \: .
\end{align*}
Let \(S\sim\cN(0,I_d)\) and set $Z \eqdef \frac{\|S\|^2}{2}$.
For any measurable \(g:\mathbb R_+\to\mathbb R\) with finite expectation, rotational invariance implies $\bE[g(Z)SS^\top]  =    \bE[g(Z)S_1^2] I_d$
Indeed, the off-diagonal entries vanish by sign symmetry, while the diagonal entries are equal by permutation symmetry.

Let $a_{\bar\eta}(Z) \eqdef \indi{Z\le \bar\eta}-\sigma(\bar\eta-Z)$ and
\[
    b_{\bar\eta}(Z)
    \eqdef
    \left(2\sigma(|Z-\bar\eta|)-1\right)\sigma(-|Z-\bar\eta|)
    =
    \tanh(|Z-\bar\eta|/2)\sigma(-|Z-\bar\eta|).
\]
Then, for \(S\sim\cN(0,I_d)\) and \(Z=\|S\|^2/2\), we have
\begin{align*}
    &H^{\lle}_{\cN}(\bar\eta)   = -\bE[a_{\bar\eta}(Z)] \: , \\
    &\Delta^{\lle}_{\cN}(\bar\eta)  =  \bE[b_{\bar\eta}(Z)SS^\top]   =   \bE[b_{\bar\eta}(Z)S_1^2]I_d \: , \\
    &R^{\lle}_{\cN}(\bar\eta)  =  -2\bE[a_{\bar\eta}(Z)SS^\top]   =    -2\bE[a_{\bar\eta}(Z)S_1^2]I_d \: .
\end{align*}
Defining
\begin{align*}
    &\alpha^{\lle}_{1,d}(\bar\eta)      \eqdef     \bE\!\left[         \indi{Z\le \bar\eta}-\sigma(\bar\eta-Z)     \right] \: , \\
    &\alpha^{\lle}_{2,d}(\bar\eta)     \eqdef     \bE\!\left[         S_1^2         \tanh(|Z-\bar\eta|/2)        \sigma(-|Z-\bar\eta|)    \right] \: , \\
    &\alpha^{\lle}_{3,d}(\bar\eta)     \eqdef     \bE\!\left[         S_1^2        \left(\indi{Z\le \bar\eta}-\sigma(\bar\eta-Z)\right)    \right]\: ,
\end{align*} 
we have shown that
\[
    H^{\lle}_{\theta^\star}
    =
    -\alpha^{\lle}_{1,d}(\bar\eta)\Sigma,
    \qquad
    \Delta^{\lle}_{\theta^\star}
    =
    \alpha^{\lle}_{2,d}(\bar\eta)\Sigma,
    \qquad
    R^{\lle}_{\theta^\star}
    =
    -2\alpha^{\lle}_{3,d}(\bar\eta)\Sigma .
\]
\paragraph{Comparison Between $\lle$ and $\sqe$.}
Consequently, a sufficient condition for the asymptotic variance of \(\widehat\theta^{\lle}_n\) to be smaller than that of \(\widehat\theta^{\sqe}_n\) is
\[
    \left(\alpha^{\lle}_{1,d}(\bar\eta)\right)^2
    +
    \left(1+\alpha_d^{\sqe}(\bar\eta)\right)
    \left(
        2\alpha^{\lle}_{1,d}(\bar\eta)
        +
        \alpha^{\lle}_{2,d}(\bar\eta)
        -
        2\alpha^{\lle}_{3,d}(\bar\eta)
    \right)
    >0 \: .
\]
The sufficient condition above should not be expected to hold uniformly over \(\bar\eta>0\) and \(d\ge 1\). 
In particular, although it provides a useful criterion for comparing the asymptotic variances of \(\widehat\theta^{\lle}_n\) and \(\widehat\theta^{\sqe}_n\), the left-hand side can become negative for some values of \(\bar\eta\) and \(d\). 
In Figure~\ref{fig:GaussianMatrices}, we observes situation where the asymptotic variance of $\wh \theta^{\lle}_{n}$ is larger than $\wh \theta^{\sqe}_{n}$. 
Therefore, the condition identifies a regime in which \(\lle\) improves over \(\sqe\), but it does not establish a uniform dominance result.
\paragraph{Comparison Between $\lle$ and $\so$.}
A sufficient condition for the asymptotic variance of \(\widehat\theta^{\lle}_n\) to be smaller than that of \(\widehat\theta^{\so}_n\) is
\[
    \left(1 + \alpha_d^{\sqe}(\bar\eta) + \alpha^{\lle}_{1,d}(\bar\eta)\right)^2 > 1 + \alpha_d^{\sqe}(\bar\eta) - \alpha^{\lle}_{2,d}(\bar\eta) + 2\alpha^{\lle}_{3,d}(\bar\eta) \: .
\]
The condition is not expected to hold uniformly over \(\bar\eta>0\) and \(d\ge 1\). Indeed, numerical evaluation of the one-dimensional Gamma expectations shows that the left-hand side can be smaller than the right-hand side, e.g. for \(d=1\) and small \(\bar\eta\). 
Thus the condition identifies a regime in which \(\widehat\theta_n^{\lle}\) improves over \(\widehat\theta_n^{\so}\), but it does not yield a uniform dominance result.
\paragraph{Comparison Between $\podpe$ and $\so$.}
We define $\podpe$ as $\dpe$ when using $\cC_{n}^{+}$ instead of $\cC_{n}$, namely
\[
    \wh \theta^{\podpe}_{n} \in \argmin \{ L^{\so}_{n}(\theta) \mid \theta \in \cC_{n}^{+} \}  \: ,
\]
and $\wh \theta^{\podpe}_{n} = \wh \theta^{\so}_{n} $ when $\wh \theta^{\so}_{n} \in \cC_{n}^{+}$.
For thresholded log-likelihood rewards, $\wh \theta^{\podpe}_{n} $ estimates $\theta^\star$ better than $\wh \theta^{\so}_{n}$ for all $n$, i.e., $\podpe$ dominates $\so$ statistically.
\begin{lemma}\label{lem:statistical_dominance_dp_over_so}
    Consider $r_{\theta} = \log p_{\theta} + \eta$ with known $\eta \in \mathrm{H}_0$.
    For all $n \in \N$ and almost surely, we have,
    \begin{align*}
        \text{for } \cF_{\cN, \Sigma}, \quad&\|\wh \theta^{\podpe}_{n} - \theta^\star\|_{\Sigma} \le \|\wh \theta^{\so}_{n} - \theta^\star\|_{\Sigma} \: .
    \end{align*}
\end{lemma}
\begin{proof}
    For Gaussian distributions, this is a direct consequence of the following facts: $\theta^\star \in \cC_n^{+}$, $\wh \theta^{\podpe}_{n} \in \argmin_{\theta \in \cC_n^{+}} \|\theta - \wh \theta^{\so}_{n} \|_{\Sigma}^2$ and $\cC_n^{+}$ is convex.
\end{proof}
\paragraph{Comparison Between $\dpe$ and $\so$.}
For thresholded log-likelihood rewards, $\dpe$ does not dominate $\so$ statistically.
Specifically, we construct a counterexample for $d=1$, $\Sigma=1$, $\theta^\star=0$ and $n=3$.
Let $r = \sqrt{2(\eta -\frac{1}{2}\log(2\pi))} > 0 $ with $\eta > \frac{1}{2}\log(2\pi)$.
Let $X \sim \cN(0_3, I_3)$ and $Z_i \eqdef \sign (r^2-X_i^2)$ for all $i\in [3]$.
Let us define the event 
\[
    \cE \eqdef \left\{
        X_1 \in \left(\frac{23}{20}r,\frac{25}{20}r\right),
        \quad
        X_2,X_3 \in \left(\frac{39}{20}r,\frac{41}{20}r\right)
    \right\} \: ,
\]
which satisfies that $\bP(\cE) > 0$. 
On $\cE$, we have $Z_1=Z_2=Z_3=-1$ and 
\[
    \cC_3 = \left\{ \theta\in\R \mid \:\forall i \in [3], \: |\theta-X_i|\ge r    \right\} =
    \R\setminus \bigcup_{i \in [3]} (X_i-r,X_i+r) \: .
\]
Let $L=\min_{i\in\{1,2,3\}} X_i$ and $M=\max_{i\in\{1,2,3\}} X_i$.
On $\cE$, the intervals $(X_i-r,X_i+r)$ overlap and their union is the single interval $(L-r,M+r)$ since
\[
    M-L    <     \frac{41}{20}r-\frac{23}{20}r     =     \frac{18}{20}r     <     2r \: .
\]
Therefore, we have $\cC_3   =    (-\infty,L-r]\cup [M+r,\infty)$, which contains the true parameter $0$ as $L-r>0$.
Let $\bar X \eqdef \widehat \theta^{\so}_3    =    \frac{1}{3}\sum_{i=1}^3 X_i$.
On $\cE$, we have $X_i>0$ and $L\le X_i\le M$ for all $i \in [3]$, hence $L-r < L \le \bar X \le M < M+r$.
Therefore, we have $\bar X \in (L-r,M+r)$, thus $\bar X \notin \cC_{3}$.

As projection of $\bar X$ on $\cC_{3}$ for the squared distance, $\widehat \theta^{\dpe}_3 \in \{L-r, M+r\}$.
On $\cE$, we have $L=X_1$.
Let $H \eqdef \min\{X_2,X_3\}$.
Therefore, $\bar X=\frac{L+H+M}{3}$ and $L < H \le M$ on $\cE$.
Let $d_{\mathrm{left}}    \eqdef    \bar X-(L-r)$ and $d_{\mathrm{right}}    \eqdef    (M+r)-\bar X $ be the distances from $\bar X$ to the left and right side of $\cC_3$.
Then, $d_{\mathrm{left}}-d_{\mathrm{right}} = 2\bar X-L-M = \frac{2H-L-M}{3}$.
On $\cE$, we have $H>\frac{39}{20}r$, $L<\frac{25}{20}r$ and $M<\frac{41}{20}r$, hence
\begin{align*}
    2H-L-M
    &>
    2\cdot \frac{39}{20}r
    -
    \frac{25}{20}r
    -
    \frac{41}{20}r =
    \frac{78-25-41}{20}r =
    \frac{12}{20}r >0 \: .
\end{align*}
Since $d_{\mathrm{left}}>d_{\mathrm{right}}$, we obtain $\widehat \theta^{\dpe}_3    =    M+r$.
Finally, since $\theta^\star=0$ and $\bar X>0$ on $\cE$, we get
\begin{align*}
    \left|\widehat \theta^{\so}_3-\theta^\star\right|    =    \bar X < M+r = \left|\widehat \theta^{\dpe}_3-\theta^\star\right| \: .
\end{align*}
As $\mathbb P(\cE)>0$, the statistical dominance of $\widehat \theta^{\dpe}_n$ over $\widehat \theta^{\so}_n$ inequality fails with positive probability.

\subsection{Laplace Distribution with Known Scale}
\label{sub:laplace_distribution_with_known_scale}

Let $\bar \eta = \eta - \eta_0$ with $\eta_0 \eqdef \log (2b)$ where $\eta \in \mathrm{H}_0 = (\eta_0,+\infty)$, hence $\bar \eta > 0$.
\paragraph{Consistency of $\sqe$}
This is a direct consequence of the regularity properties of $p_{\theta}$ and $\sigma$.
\paragraph{Asymptotic Variance of $\sqe$}
Let $\Delta^{\sqe}_{\theta^\star}$ be as in Lemma~\ref{lem:sp_mle_asymptotic_variance}.
Let $S = (\theta^\star - X)/b$ with $X \sim p_{\theta^\star}$.
Then, $S \sim \textrm{Lap}(0,1)$, $r_{\theta^{\star}}(X) = - |S| + \bar \eta$ and $\nabla_{\theta^\star} r_{\theta^\star}(X)= -\sign(S)/b$.
Since $\sigma(t)\sigma(-t)=\sigma'(t)$, we obtain
\begin{align*}
    \Delta^{\sqe}_{\theta^\star}
    = \frac{1}{b^2} \alpha^{\sqe}(\bar \eta)
    \quad \text{with} \quad
    \alpha^{\sqe}(\bar \eta)
    \eqdef \bE_{S \sim \textrm{Lap}(0,1)}
    \left[ \sigma'(- |S| + \bar \eta) \right] \: .
\end{align*}
Let $T=|S|$. Since $S\sim \mathrm{Lap}(0,1)$, we have $T\sim \mathrm{Exp}(1)$. Hence
\begin{align*}
    \alpha^{\sqe}(\bar \eta)
    &= e^{-\bar\eta}
    \int_0^\infty
    \frac{1}{(1+e^{t-\bar\eta})^2}\,dt = e^{-\bar\eta}
    \int_{e^{-\bar\eta}}^\infty
    \frac{1}{u(1+u)^2}\,du = e^{-\bar\eta}\log(1+e^{\bar\eta})
    -\frac{1}{1+e^{\bar\eta}} \: ,
\end{align*}
where we used $u=e^{t-\bar\eta}$.
Then, we have shown
\[
    \Delta^{\sqe}_{\theta^\star}
    = \frac{\alpha^{\sqe}(\bar \eta)}{b^2}
    = \frac{1}{b^2}
    \left(
    e^{-\bar\eta}\log(1+e^{\bar\eta})
    -(e^{\bar\eta}+1)^{-1}
    \right)>0 \: .
\]
Since $\cI(q_{\theta^\star}^{\sto})^{-1}
= (1+\alpha^{\sqe}(\bar\eta))^{-1}b^2$, the asymptotic variance of $\wh\theta^{\sqe}_n$ is smaller than that of $\wh\theta^{\so}_n$.
\paragraph{Consistency of $\lle$}
Using the symmetry of the Laplace distribution and the fact that the factor depending on $|S|$ is even in $S$, while $\sign(S)$ is odd, we obtain
\begin{align*}
    \bE_{S \sim \textrm{Lap}(0,1)}
    \left[
    \sign(\bar\eta-|S|)
    \sigma(-||S|-\bar\eta|)
    \sign(S)
    \right]=0 \: .
\end{align*}
Thus, the stationarity condition $\nabla M_{\det}(\theta^\star)=0$ is satisfied, and $\lle$ is consistent under the usual M-estimation consistency conditions.
\paragraph{Asymptotic Variance of $\lle$.}
The Laplace log-likelihood is not twice differentiable at $x=\theta$. Although $\nabla_\theta^2r_\theta(X)=0$ almost surely, differentiating the population objective produces a nonzero contribution from this singularity. Consequently, the pointwise Hessian calculation in Lemma~\ref{lem:asymptotic_variance_lle} cannot be applied directly. We instead compute the curvature of the population objective and the variance of its score.

Recall that
\[
    M_{\det}(\theta)
    =\bE_{p_{\theta^\star}}
    \left[
    \log p_\theta(X)
    +\log\sigma(Zr_\theta(X))
    \right],
    \qquad
    Z=\sign(r_{\theta^\star}(X)) \: .
\]
For $x\neq\theta$, the score satisfies
\[
    \nabla_\theta\log q_\theta^{\sto}(x,z)
    =
    \frac{\sign(x-\theta)}{b}
    \left[
    1+z\sigma(-zr_\theta(x))
    \right] \: .
\]
Writing $a\equiv\bar\eta$, let $T=|S|\sim\Exp(1)$ and define
\[
    u(T)\eqdef
    \sign(a-T)\sigma(-|a-T|) \: .
\]
At $\theta=\theta^\star$, the score has zero expectation and variance
\begin{align*}
    V_{2,\theta^\star}
    &=\frac{1}{b^2}
    \bE\left[(1+u(T))^2\right] =\frac{1}{b^2}
    \left(
    1+\alpha^{\sqe}(a)
    -\alpha_2^{\lle}(a)
    +2\alpha_3^{\lle}(a)
    \right),
\end{align*}
where we define
\begin{align*}
    &\alpha_3^{\lle}(a)
    \eqdef \bE\left[
    \sign(a-T)\sigma(-|a-T|)
    \right] \: ,\\
    &\alpha_2^{\lle}(a)
    \eqdef \bE\left[
    (2\sigma(|a-T|)-1)\sigma(-|a-T|)
    \right] \: .
\end{align*}
The second equality follows from
$\bE[u(T)]=\alpha_3^{\lle}(a)$ and
$\bE[u(T)^2]=\alpha^{\sqe}(a)-\alpha_2^{\lle}(a)$.

To compute the curvature, observe that
\[
    M_{\det}'(\theta)
    =
    \frac{1}{b}\bE_{p_{\theta^\star}}
    \left[
    \sign(X-\theta)
    \left(
    1+Z\sigma(-Zr_\theta(X))
    \right)
    \right] \: .
\]
Differentiating this expectation at $\theta^\star$ yields two contributions. Away from $X=\theta$, differentiation of the logistic term gives
\[
    \frac{1}{b^2}
    \bE\left[
    \sigma(a-T)\sigma(T-a)
    \right]
    =
    \frac{\alpha^{\sqe}(a)}{b^2} \: .
\]
The discontinuity of $\sign(X-\theta)$ at $X=\theta$ contributes
\[
    \frac{2p_{\theta^\star}(\theta^\star)}{b}
    \left(1+\sigma(-a)\right)
    =
    \frac{1+\sigma(-a)}{b^2} \: ,
\]
since $p_{\theta^\star}(\theta^\star)=1/(2b)$ and the true quality label is positive at the center.
Therefore,
\[
    V_{1,\theta^\star}
    \eqdef -M_{\det}''(\theta^\star)
    =
    \frac{1}{b^2}
    \left(
    1+\alpha^{\sqe}(a)+\sigma(-a)
    \right) \: .
\]
Under the standard regularity conditions for nonsmooth M-estimators, the asymptotic variance is consequently
\begin{align*}
    V_{\theta^\star,\det}
    &=V_{1,\theta^\star}^{-2}V_{2,\theta^\star}=b^2
    \frac{
    1+\alpha^{\sqe}(a)-\alpha_2^{\lle}(a)
    +2\alpha_3^{\lle}(a)
    }{
    \left(
    1+\alpha^{\sqe}(a)+\sigma(-a)
    \right)^2
    } \: .
\end{align*}
In particular, the additional curvature term $\sigma(-a)$, which is absent from the pointwise Hessian calculation, reduces the asymptotic variance.
We now give explicit expressions for $\alpha_2^{\lle}$ and $\alpha_3^{\lle}$.

Splitting at $T=a$, we obtain
\begin{align*}
\alpha_{3}^{\lle}(a)
&=
\int_0^a \frac{e^{-t}}{1+e^{a-t}}\,dt
-
\int_a^\infty \frac{e^{-t}}{1+e^{t-a}}\,dt\\
&=
e^{-a}\int_0^a \frac{e^y}{1+e^y}\,dy
-
e^{-a}\int_0^\infty \frac{e^{-s}}{1+e^s}\,ds=
e^{-a}\left[\log(1+e^a)-1\right].
\end{align*}
Using $(2\sigma(z)-1)\sigma(-z)
=\frac{e^z-1}{(1+e^z)^2}$ for $z\ge 0$, and splitting at $T=a$, we get
\begin{align*}
\alpha_{2}^{\lle}(a)
&=
\int_0^a
\frac{e^{a-t}-1}{(1+e^{a-t})^2}
e^{-t}\,dt
+
\int_a^\infty
\frac{e^{t-a}-1}{(1+e^{t-a})^2}
e^{-t}\,dt\\
&=
e^{-a}
\int_0^a
\frac{e^y(e^y-1)}{(1+e^y)^2}\,dy
+
e^{-a}
\int_0^\infty
\frac{e^s-1}{(1+e^s)^2}
e^{-s}\,ds\\
&=
e^{-a}
\left[
\log(1+e^a)
+\frac{2}{1+e^a}
+2\log 2-3
\right] \: .
\end{align*}
Therefore, the quantities defined in Lemma~\ref{lem:asymptotic_variance_lle} satisfy
\begin{align*}
    R^{\lle}_{\theta^\star}
    &=-\frac{2}{b^2}\alpha_3^{\lle}(\bar\eta)
    =-\frac{2e^{-\bar\eta}}{b^2}
    \left[\log(1+e^{\bar\eta})-1\right] \: ,\\
    \Delta^{\lle}_{\theta^\star}
    &=\frac{1}{b^2}\alpha_2^{\lle}(\bar\eta)=\frac{e^{-\bar\eta}}{b^2}
    \left[
    \log(1+e^{\bar\eta})
    +\frac{2}{1+e^{\bar\eta}}
    +2\log 2-3
    \right] \: .
\end{align*}
Combining these expressions with the corrected population curvature gives
\[
    V_{\theta^\star,\det}
    =
    b^2
    \frac{
    1+\alpha^{\sqe}(\bar\eta)
    -\alpha_2^{\lle}(\bar\eta)
    +2\alpha_3^{\lle}(\bar\eta)
    }{
    \left(
    1+\alpha^{\sqe}(\bar\eta)
    +\sigma(-\bar\eta)
    \right)^2
    }\: .
\]
\paragraph{Comparison Between $\lle$ and $\sqe$.}
Recall that $ V_{\theta^\star,\sto}
    =\frac{b^2}{1+\alpha^{\sqe}(\bar\eta)}$.
Consequently, a necessary and sufficient condition for the asymptotic variance of $\wh\theta_n^{\lle}$ to be smaller than that of $\wh\theta_n^{\sqe}$ is
\begin{align*}
    &\left(1+\alpha^{\sqe}(\bar\eta)\right)
    \left(
    \alpha_2^{\lle}(\bar\eta)
    -2\alpha_3^{\lle}(\bar\eta)
    +2\sigma(-\bar\eta)
    \right)+\sigma(-\bar\eta)^2>0 \: .
\end{align*}
This condition does not hold uniformly over $\bar\eta>0$.
For instance, when $\bar\eta=1$, the normalized asymptotic variances satisfy
\[
    \frac{V_{\theta^\star,\det}}{b^2}
    \approx 0.617
    <0.824
    \approx\frac{V_{\theta^\star,\sto}}{b^2} \: .
\]
However, as $\bar\eta\to+\infty$, the explicit expressions yield
\begin{align*}
    \frac{V_{\theta^\star,\det}}{b^2}
    &=
    1-2(\log 2)e^{-\bar\eta}
    +O\left(\bar\eta^2e^{-2\bar\eta}\right),\\
    \frac{V_{\theta^\star,\sto}}{b^2}
    &=
    1-(\bar\eta-1)e^{-\bar\eta}
    +O\left(\bar\eta^2e^{-2\bar\eta}\right).
\end{align*}
Thus, $\wh\theta_n^{\sqe}$ has a smaller asymptotic variance than $\wh\theta_n^{\lle}$ for sufficiently large $\bar\eta$. Neither estimator uniformly dominates the other.
\paragraph{Comparison Between $\lle$ and $\so$.}
Since the asymptotic variance of $\wh\theta_n^{\so}$ is $b^2$, a necessary and sufficient condition for $\wh\theta_n^{\lle}$ to have smaller asymptotic variance is
\begin{align*}
    &\alpha^{\sqe}(\bar\eta)^2
    +\alpha^{\sqe}(\bar\eta)
    +\alpha_2^{\lle}(\bar\eta)
    -2\alpha_3^{\lle}(\bar\eta)\\
    &\quad
    +2\sigma(-\bar\eta)
    \left(1+\alpha^{\sqe}(\bar\eta)\right)
    +\sigma(-\bar\eta)^2>0 \: .
\end{align*}
Unlike the condition obtained by omitting the nonsmooth curvature contribution, this inequality holds for every $\bar\eta>0$.

Indeed, let $a=\bar\eta$, $q=e^{-a}$ and $L=\log(1+e^a)$. From the explicit expressions above,
\[
    1+\alpha^{\sqe}(a)+\sigma(-a)
    =1+qL
\]
and
\begin{align*}
    &1+\alpha^{\sqe}(a)
    -\alpha_2^{\lle}(a)
    +2\alpha_3^{\lle}(a) =
    1+2qL-2q\log 2
    -\frac{q}{1+e^a} \: .
\end{align*}
Consequently,
\begin{align*}
    &\left(1+qL\right)^2
    -\left(
    1+2qL-2q\log 2
    -\frac{q}{1+e^a}
    \right)=
    q^2L^2+2q\log 2
    +\frac{q}{1+e^a}>0 \: .
\end{align*}
Therefore, for all $\bar\eta>0$, $V_{\theta^\star,\det}  <b^2   =V_{\theta^\star,\so}$.
Thus, under the stated asymptotic regularity conditions, $\wh\theta_n^{\lle}$ is uniformly more efficient than the sample-only MLE for the Laplace location family with known scale and thresholded log-likelihood rewards.

\subsection{Exponential Distribution}
\label{sub:exponential_distribution}

Let $\eta \in \R$ and $\theta^\star > e^{-\eta}$.
\paragraph{Consistency of $\sqe$}
This is a direct consequence of the regularity property of $p_{\theta}$ and $\sigma$.
\paragraph{Asymptotic Variance of $\sqe$}
Let $S = \theta^\star X$ with $X \sim p_{\theta^\star}$.  
Then, $S \sim \Exp(1)$, $r_{\theta^{\star}}(X) = - S + \log \theta^\star + \eta$ and $\nabla_{\theta^\star} r_{\theta^\star}(X)= (1-S)/\theta^\star$.
Since $\sigma(t)\sigma(-t)=\sigma'(t)$, we obtain
\begin{align*}
    \Delta^{\sqe}_{\theta^\star} = \frac{1}{(\theta^\star)^2}  \alpha^{\sqe}( \eta) \quad \text{with} \quad \alpha^{\sqe}( \eta) \eqdef \bE_{\Exp(1)}\left[ \sigma'(-  S + \log \theta^\star + \eta) (1-S)^2 \right] > 0 \: .
\end{align*}
Since $\cI(q_{\theta^\star}^{\sto})^{-1} = (1 + \alpha^{\sqe}(\eta))^{-1} (\theta^\star)^2$, the asymptotic variance of $\wh \theta^{\sqe}_{n}$ is smaller than $\wh \theta^{\so}_{n}$.
Therefore, the asymptotic variance of $\wh \theta^{\sqe}_{n}$ is smaller than $\wh \theta^{\so}_{n}$.
\paragraph{Lack of Consistency of $\lle$}
The sign of $\nabla_{\theta^\star}M_{\det}(\theta^\star)$ equals the one of
\begin{align*}
     \bE_{S \sim \Exp(1)}[\sign(- S + \log \theta^\star + \eta) \sigma(-|- S + \log \theta^\star + \eta|) (1-S) ] \: .
\end{align*}
Let us define $a \eqdef \log \theta^\star+\eta$ and $I(a)\eqdef \mathbb E\!\left[\sign(a-S)\sigma(-|a-S|)(1-S)\right]$.
Let $\operatorname{Li}_2(z) \eqdef -\int_0^z \frac{\log(1-t)}{t}\,dt$ and $F(a)\eqdef \log(1+e^a)+\operatorname{Li}_2(-e^a)+a\mathbf 1_{\{a>0\}}$.
Then, we have $I(a)=e^{-a}F(a)$, hence $\sign(\nabla_{\theta^\star}M_{\det}(\theta^\star) )  = \sign (F(a) )$.

We can show that $\lim_{a\to-\infty}F(a)=0$, $\lim_{a\to+\infty}F(a)=-\infty$, $F(0)=\log 2-\frac{\pi^2}{12}<0$ and $F(1)\approx 0.50698>0$.
Moreover, we have 
\begin{align*}
    F'(a)= \begin{cases}
        \frac{e^a}{1+e^a}-\log(1+e^a) < 0 &\text{if } a\le 0 \\
        1+\frac{e^a}{1+e^a}-\log(1+e^a) &\text{if } a> 0
    \end{cases}
    \quad \text{and} \quad \forall a > 0, \quad F''(a)=-\left(\frac{e^a}{1+e^a}\right)^2<0 \: .
\end{align*}
This implies that \(F(a)<0\) for all $a\le 0$ and \(F\) is strictly concave on \((0,\infty)\).
Therefore, \(F\) has exactly two positive zeros $(\alpha, \beta)$, and we have $(\alpha, \beta) = (0.165, 2.956)$ numerically.

In summary, we have $\nabla_{\theta^\star}M_{\det}(\theta^\star) = 0$ if and only if $\theta^\star \in \{e^{\alpha-\eta}, e^{\beta-\eta}\}$.


\section{Supplementary Experiments}
\label{app:experiments}
Using the same empirical setup as in Section~\ref{sec:experiments}, we conduct additional experiments to support our theoretical claims for other distributions (Appendix~\ref{app:ssec_other_distributions}), other estimators and thresholds for Gaussian distributions (Appendix~\ref{app:ssec_other_estimators}), other convex surrogates of the $0$-$1$ loss (Appendix~\ref{app:ssec_other_losses}), normalized/regularized versions of the logistic loss (Appendix~\ref{app:ssec_norm_regu}) and quality-based M-estimators (Appendix~\ref{app:ssec_qualityMestimator}).
Appendix~\ref{app:ssec_implementation_details} contains additional implementation details.
\subsection{Implementation Details}
\label{app:ssec_implementation_details}
\paragraph{Feasibility and Optimization.}
Although empirical $0$--$1$ minimization is generally NP-hard under stochastic feedback, deterministic feedback often yields tractable interpolation constraints.
If $\theta\mapsto r_\theta(x)$ is quasi-concave for all $x$, then the positive-only set $\cC_n^{+}$ is convex and has at most $|\cI_n^{+}|$ constraints.
Thus, any estimator $\wh\theta_n \in \cC_n^{+}$ can be computed by convex feasibility methods for convex sets, given efficient separation or projection oracles. 
Moreover, whenever $L_n^{\so}$ is convex, $\wh\theta_n \in \argmin \{ L^{\so}_{n}(\theta) \mid \theta \in \cC_{n}^{+} \}$ can be computed by convex optimization over a convex set.
As we will detail below, this covers thresholded log-likelihood rewards for Gaussian distributions with ball constraints, and Laplace and exponential distributions with interval constraints.

The full quality set $\cC_n$ is generally less tractable because bad samples impose sublevel constraints that need not be convex.
Thus, feasibility and constrained likelihood minimization over $\cC_n$ are generally non-convex.
They become tractable under a stronger structure, such as quasi-affinity of $\theta\mapsto r_\theta(x)$, which reduces both positive and negative constraints to half-spaces.

The geometry of $\cC_n^{\all}$ is more attractive than $\cC_n$ or $\cC_n^{+}$.
If $\theta\mapsto \ell_\theta(x,y)$ is quasi-affine for all $(x,y)\in\cX^2$, then $\cC_n^{\all}$ is a convex polytope with at most $|\cI_{n}^{+}||\cI_{n}^{-}| $ half-space constraints, as in Gaussian or Laplace distributions with shifted log-likelihood rewards.
Hence, feasibility reduces to linear feasibility, and likelihood minimization over $\cC_n^{\all}$ is convex whenever $L_n^{\so}$ is convex.
Due to the at most quadratic number of constraints, preference-based estimators using $\cC_n^{\all}$ can become computationally more intensive than those based on $\cC_n^{+}$. 

\paragraph{Thresholded Log-likelihood Rewards.}
For thresholded log-likelihood rewards, we provide more details depending on the considered distributions.
\begin{itemize}
    \item $\cF_{\cN,I_d}$ and $\cC_{n}^{+}$: the feasible set is an intersection of balls and is convex. 
    $\anye$ ($\po$) is therefore a convex feasibility problem, while $\dpe$ ($\po$) is the projection of the sample mean onto this set. 
    These globally tractable estimators already attain our multivariate fast-rate guarantee.
    \item $\cF_{\cN,1}$, $\cF_{\lap,1}$ or $\cF_{\Exp}$ and $\cC_{n}^{+}$: the feasible set is an interval and the negative log-likelihood is convex on its natural domain. 
    $\anye$ ($\po$) and $\dpe$ ($\po$) are therefore globally computable. 
    For $\cF_{\Exp}$, these $\po$ estimators are computationally tractable, yet do not identify the true component.
    \item Univariate full-quality constraints: in all our Gaussian, Laplace, exponential, Rayleigh, and Pareto examples, the feasible set is a finite union of intervals. 
    A global $\anye$ or $\dpe$ solution can be obtained by sorting the constraint endpoints and minimizing the one-dimensional likelihood on each feasible component. 
    \item $\cF_{\cN,I_d}$ and $\cC_{n}$: bad labels generally introduce nonconvex sublevel constraints. 
    We do not claim a general polynomial-time algorithm or a global certificate in this setting. 
    In the experiments, we used Ipopt with one initialization and warm-started successive sample sizes. 
    Thus, the reported DQ solution is a local solution. 
    Importantly, none of our multivariate fast-rate guarantees requires solving this nonconvex problem exactly.
    \item Preference-based estimators: for Gaussian and Laplace shifted log-likelihood rewards, comparison constraints are half-spaces, so both cross and all-pairs feasible sets are convex polytopes and likelihood minimization remains convex. 
    The all-pairs formulation contains $\cO(n^2)$ written constraints, although in one dimension only the extremal observations adjacent to the quality boundaries determine the feasible interval.
\end{itemize}
\paragraph{Reproducibility.}
Code for reproducing our empirical results is available as supplementary material.
Our code is implemented in \texttt{Julia}~\citep{Julia-2017}, version \texttt{1.12.5}.
The plots are generated with \texttt{StatsPlots}. 
The optimization problems defining some of our estimators are solved numerically with \texttt{JuMP}~\citep{lubin2023jump}, by using the \texttt{Ipopt}~\citep{wachter2006implementation} and \texttt{Optim}~\citep{Mogensen2018} solvers. 
Other dependencies are listed in the \texttt{Readme.md} that provides detailed julia instructions to reproduce our experiments, as well as a \texttt{script.sh} to run them all at once. 
Our experiments are conducted on 4 Intel(R) Core(TM) Ultra 7 165U 4.9GHz CPU with 8G memory.
The duration of individual experiment ranges from a couple of minutes to a couple of hours for the most expensive ones.
\begin{figure}[ht]
    \centering
    \includegraphics[width=0.48\linewidth]{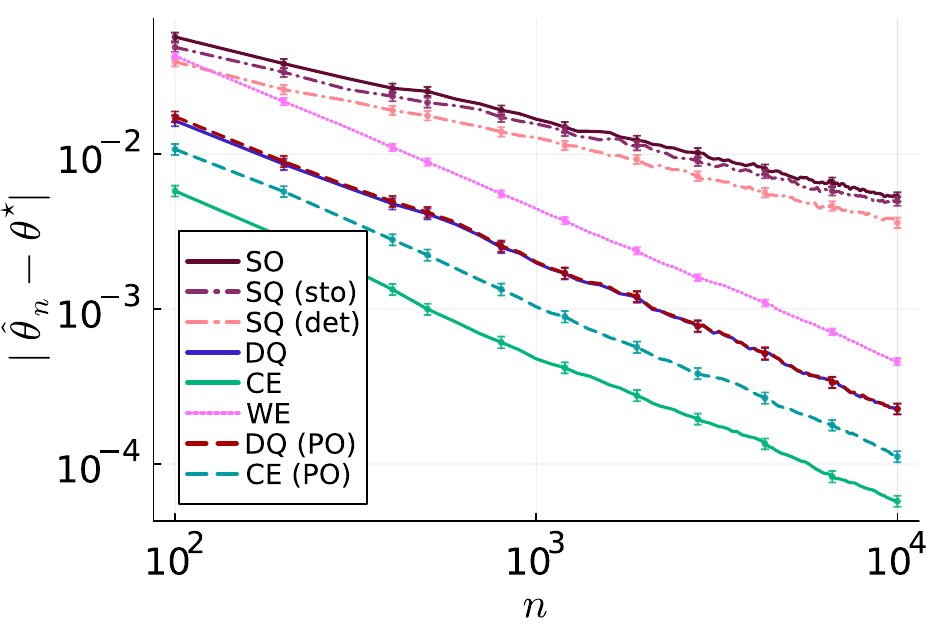}  
    \includegraphics[width=0.48\linewidth]{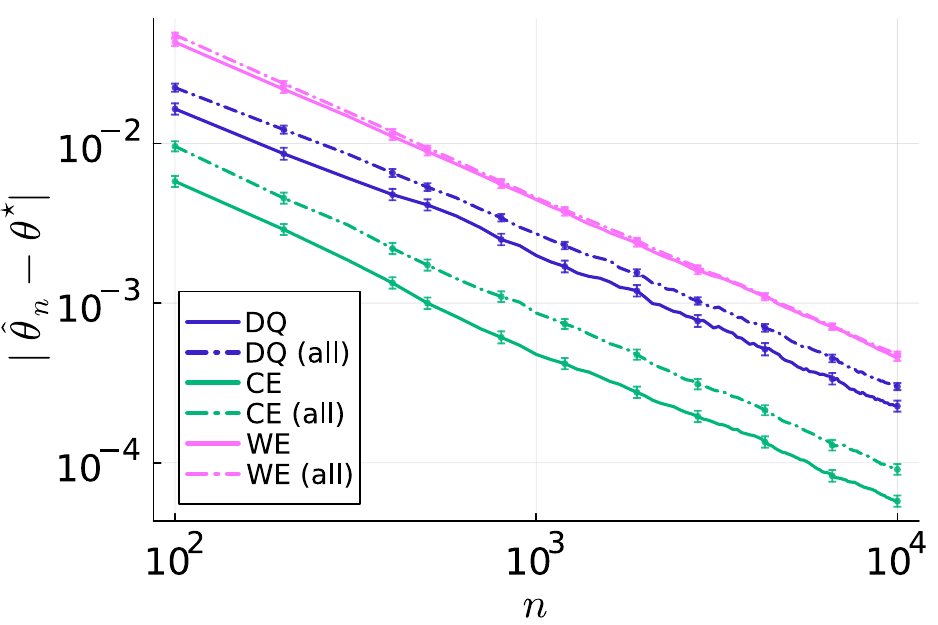}  \\
    \includegraphics[width=0.7\linewidth]{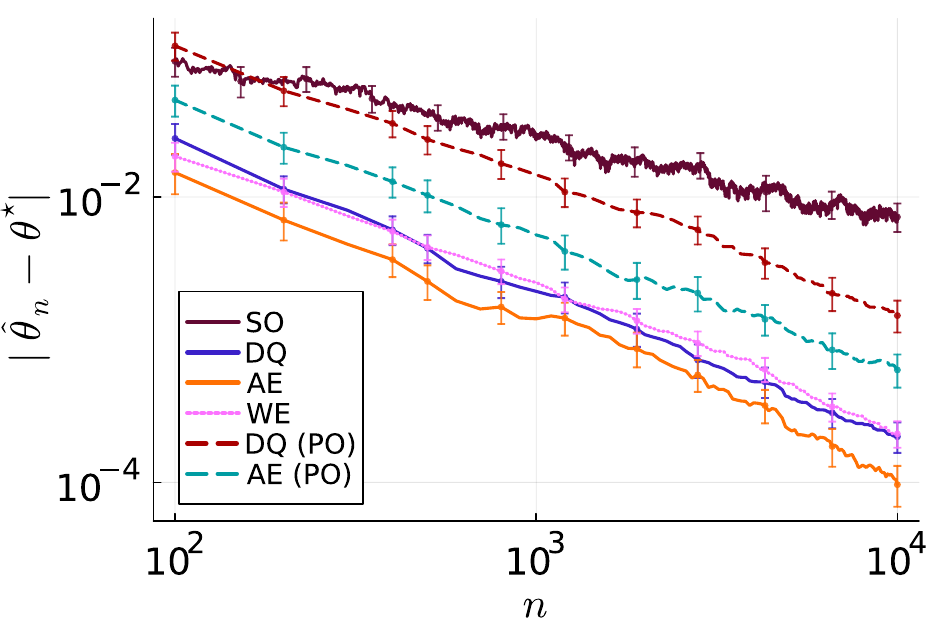} 
    \caption{Estimation errors with $\theta^\star \sim \cU([1,2])$ of (a-b) Lap$(\theta^\star,1)$ ($N_{\text{runs}} = 10^3$) and (c) Rayleigh$(\sqrt{\theta^\star})$ ($N_{\text{runs}} = 10^2$).
    The thresholds $\eta(\theta^\star)$ are chosen such that $p_{\theta^\star}(r_{\theta^\star} > 0)=1/2$.
    Error bars are $\pm 2$ standard errors.}
    \label{fig:OtherDistFctnErrorSupp}
\end{figure}
\subsection{Accelerated Rates for Other Distributions}
\label{app:ssec_other_distributions}
\subsubsection{Laplace Distribution with Known Scale}
\paragraph{Estimators.}
For $\cF_{\lap, 1}$, we have $\wh \theta^{\so}_{n} = \textrm{median}(\{X_i\}_{i \in [n]})$ and
\[
     \cC_n = \{\theta \mid \forall i \in [n], \: Z_i(\eta  - \log(2) - |X_i - \theta|) \ge 0\} \: .
\]
As for $\cF_{\cN, 1}$, we define the estimators $\sqe$, $\lle$, $\dpe$, $\cee$, $\we$, $\dpe$ ($\po$), and $\cee$ ($\po$).
\paragraph{Experiments.}
Similarly as Figures~\ref{fig:Gaussian1DVariousError}(a-b), Figures~\ref{fig:OtherDistFctnErrorSupp}(a-b) confirms empirically the difference in estimation rate between M-estimators ($\so$, $\sqe$ and $\lle$)---obtaining $\cO(1/\sqrt{n})$---and our estimators based on quality constraints---achieving $\cO(1/n)$.
As expected, $\sqe$ and $\lle$ slightly outperform $\so$, and estimators using $\cC_n$ are consistently better than their counterparts using $\cC_n^{+}$ or $\cC_{n}^{\all}$.
\subsubsection{Rayleigh Distribution}
Let $\sigma > 0$ be the scale parameter characterizing a Rayleigh distribution, and $\theta = \frac{1}{\sigma^2} > 0$.
We have $\Theta\subseteq \R_{+}$, $\cX =\R_{+}$ and $k=d=1$.
Then, for all $x \in \cX$,
\begin{align*}
    &r_{\theta}(x) = \log p_{\theta}(x) + \eta =  -x^2 \theta /2 + \log(x) + \log(\theta) + \eta \: .
\end{align*}
\paragraph{Estimators.}
We have $\wh \theta^{\so}_{n} = \frac{2n}{\sum_{i \in [n]} X_i^2}$ and
\[
     \cC_n = \{\theta \mid \forall i \in [n], \: Z_i(\eta  -X_i^2 \theta /2 + \log(X_i) + \log(\theta)) \ge 0\} \: .
\]
As for $\cF_{\cN, 1}$, we define the estimators $\dpe$, $\anye$, $\we$, $\dpe$ ($\po$), and $\anye$ ($\po$).
\paragraph{Experiments.}
Similarly as Figures~\ref{fig:Gaussian1DVariousError}(a) and Figure~\ref{fig:OtherDistFctnErrorSupp}(a), Figure~\ref{fig:OtherDistFctnErrorSupp}(c) confirms empirically the difference in estimation rate between $\so$---obtaining $\cO(1/\sqrt{n})$---and our estimators based on quality constraints---achieving $\cO(1/n)$.
\begin{figure}[ht]
    \centering
    \includegraphics[width=0.32\linewidth]{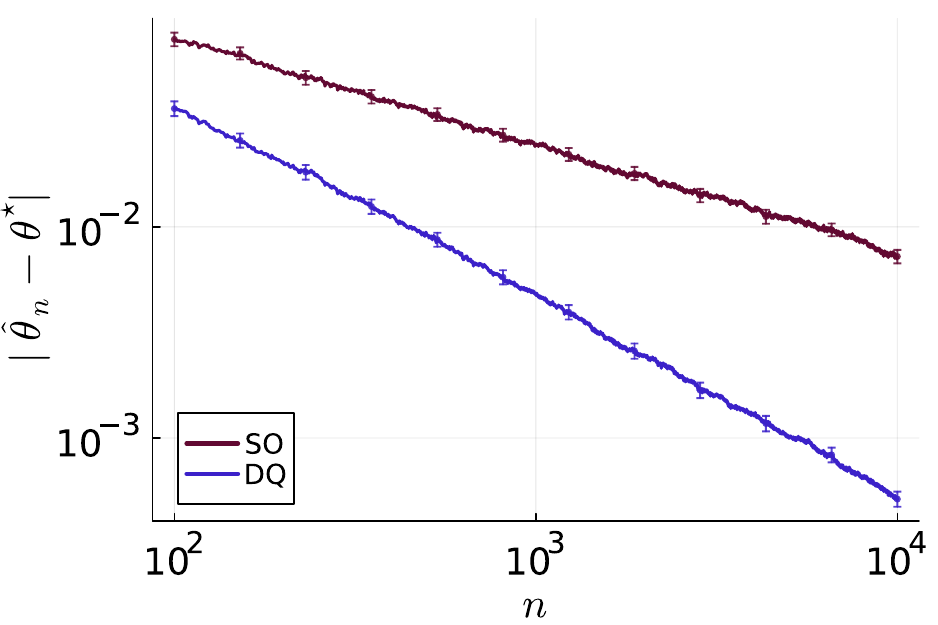}  
    \includegraphics[width=0.32\linewidth]{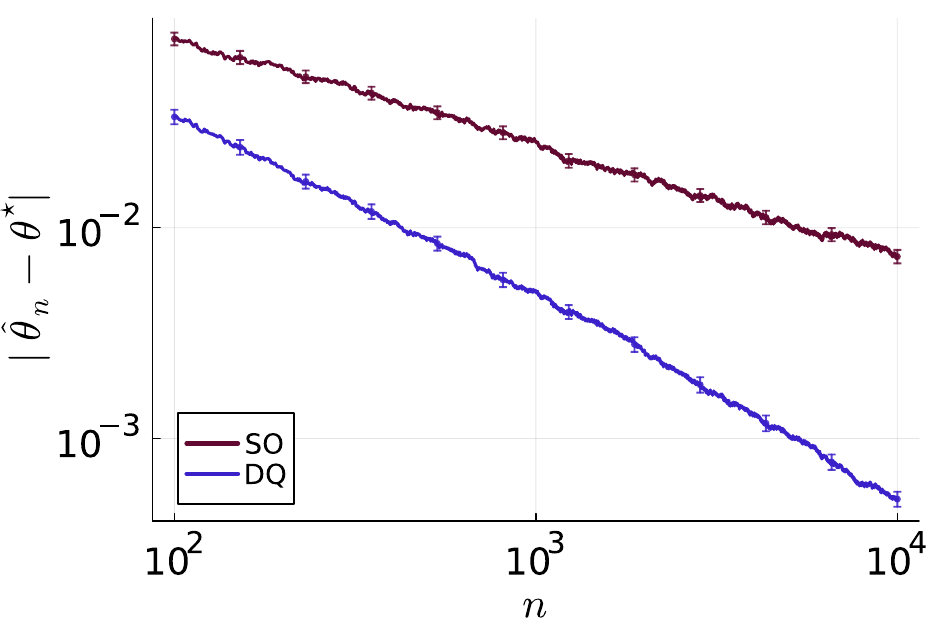}  
    \includegraphics[width=0.32\linewidth]{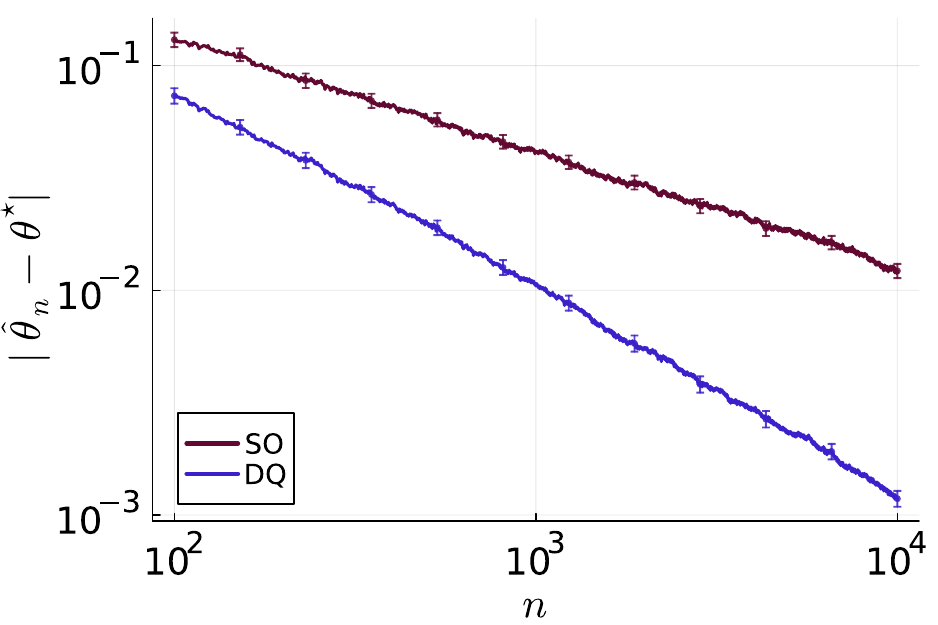} 
    \caption{Estimation errors of (a) Exp$(\theta^\star)$ with $\theta^\star \sim \cU([1,2])$, (b) Lap$(0,\theta^\star)$ with $\theta^\star \sim \cU([1,2])$, and (c) Pareto$(\theta^\star,1)$ with $\theta^\star \sim \cU([2,3])$ ($N_{\text{runs}} = 10^3$).
    The thresholds $\eta(\theta^\star)$ are chosen such that $p_{\theta^\star}(r_{\theta^\star} > 0)=1/2$.
    Error bars are $\pm 2$ standard errors.}
    \label{fig:OtherDistFctnErrorBisSupp}
\end{figure}
\subsubsection{Exponential Distribution}
\paragraph{Estimators.}
For $\cF_{\Exp}$, we have $\wh \theta^{\so}_{n} = n/(\sum_{i \in [n]}X_i)$ and
\[
     \wh \theta^{\dpe}_{n} = \argmin_{\theta \in \cC_{n}} \{\theta / \wh \theta^{\so}_{n} - \log \theta \} \quad \text{with}\quad \cC_n = \{\theta \mid \forall i \in [n], \: Z_i(\eta  - \theta X_i + \log \theta) \ge 0\} \: .
\]
\paragraph{Experiments.}
Figure~\ref{fig:OtherDistFctnErrorBisSupp}(a) confirms empirically the difference in estimation rate between $\so$ and $\dpe$, i.e., $\cO(1/\sqrt{n})$ versus $\cO(1/n)$.
\subsubsection{Laplace Distribution with Known Center}
Let $\theta$ denote the positive part of the natural parameter of a centered laplace distribution, i.e., $\theta = 1/\sigma$.
We have $\Theta = \R_{+}^{\star}$, $\cX = \R$ and $k = d = 1$.
Then, for all $x\in \cX$,
\begin{align*}
    &r_{\theta}(x) = \log p_{\theta}(x) + \eta = -\theta |x| + \log (\theta/2) + \eta \: .
\end{align*}
\paragraph{Estimators.}
We have $\wh \theta^{\so}_{n} = n/(\sum_{i \in [n]} | X_i |)$ and
\[
     \wh \theta^{\dpe}_{n} = \argmin_{\theta \in \cC_{n}} \{ \theta / \wh \theta^{\so}_{n} - \log \theta \} \quad \text{with} \quad \cC_n = \{\theta \mid \forall i \in [n], \: Z_i(\eta  - \theta |X_i| + \log (\theta/2)) \ge 0\} \: .
\]
\paragraph{Experiments.}
Figure~\ref{fig:OtherDistFctnErrorBisSupp}(b) confirms empirically the difference in estimation rate between $\so$ and $\dpe$, i.e., $\cO(1/\sqrt{n})$ versus $\cO(1/n)$.
\subsubsection{Pareto Distribution with Known Minimum}
Let $\theta > 1$ be the shape parameter of a Pareto distribution with known minimum $x_m \eqdef 1$.
We have $\Theta = (1,+ \infty)$, $\cX = [1, + \infty)$ and $k = d = 1$.
Then, for all $x\in \cX$,
\begin{align*}
    &r_{\theta}(x) = \log p_{\theta}(x) + \eta = - (\theta+1) \ln (x)  + \ln \theta +  \eta \: .
\end{align*}
\paragraph{Estimators.}
We have $\wh \theta^{\so}_{n} = n/(\sum_{i \in [n]} \log X_i)$ and
\[
     \wh \theta^{\dpe}_{n} = \argmin_{\theta \in \cC_{n}} \{ \theta / \wh \theta^{\so}_{n} - \log \theta \} \quad \text{with} \quad \cC_n = \{\theta \mid \forall i \in [n], \: Z_i(\eta  - (\theta+1) \log (X_i) + \log \theta) \ge 0\} \: .
\]
\paragraph{Experiments.}
Figure~\ref{fig:OtherDistFctnErrorBisSupp}(c) confirms empirically the difference in estimation rate between $\so$ and $\dpe$, i.e., $\cO(1/\sqrt{n})$ versus $\cO(1/n)$.
\begin{figure}[ht]
    \centering
    \includegraphics[width=0.7\linewidth]{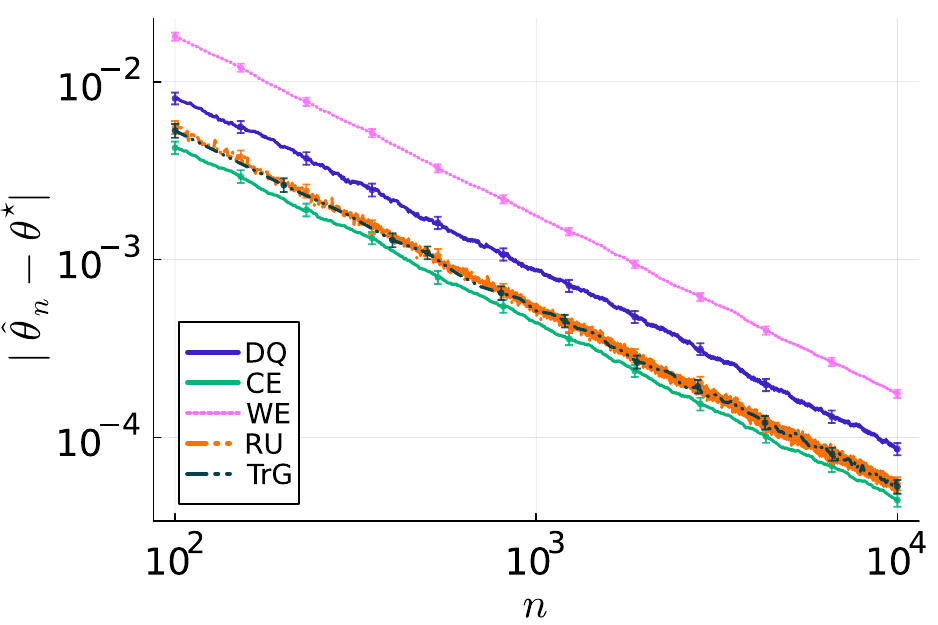}   
    \caption{Estimation errors of $\cN(\theta^\star,1)$ with $\theta^\star \sim \cU([1,2])$ ($N_{\text{runs}} = 10^3$) for $\eta(1/2) \approx 1.15$ so that $c_{\eta(1/2)}^2 \eqdef F^{-1}_{\chi^2(1)}(1/2)$.
    Error bars are $\pm 2$ standard errors.}
    \label{fig:GaussianFctnErrorSupp}
\end{figure}
\subsection{Other Estimators and Thresholds for Gaussian Distributions}
\label{app:ssec_other_estimators}
\paragraph{Other Estimators for Univariate Gaussian.}
To better understand the performance of $\cee$, we consider other estimators.
The randomized uniform ($\rue$) estimator returns $\wh \theta^{\rue}_{n} \sim \cU([L_n,U_n])$ since $\cC_{n} = [L_n,U_n]$. 
The truncated Gaussian estimator ($\mathrm{TrG}$) returns a realization from a Gaussian distribution with mean $\wh \theta^{\cee}_{n}$ and variance $4/n$, which is truncated to $\mathcal C_n$. 

Figure~\ref{fig:GaussianFctnErrorSupp} reveals that $\mathrm{TrG}$ performs on par with $\rue$, yet $\cee$ slightly outperform both $\mathrm{TrG}$ and $\rue$. 
This suggests that being far away from the boundary of $\mathcal C_n$ improves performance compared to $\dpe$ that lies on the boundary of $\mathcal C_n$ (as observed empirically). 
Moreover, randomization on $\mathcal C_n$ worsens performance compared to $\cee$.
Using the derivation in the introduction on univariate Gaussian, it is coherent that $\cee$ improves on $\dpe$ by a multiplicative constant: the average of those two (non-independent) random variables decreases faster. 
\begin{figure}[ht]
    \centering
    \includegraphics[width=0.48\linewidth]{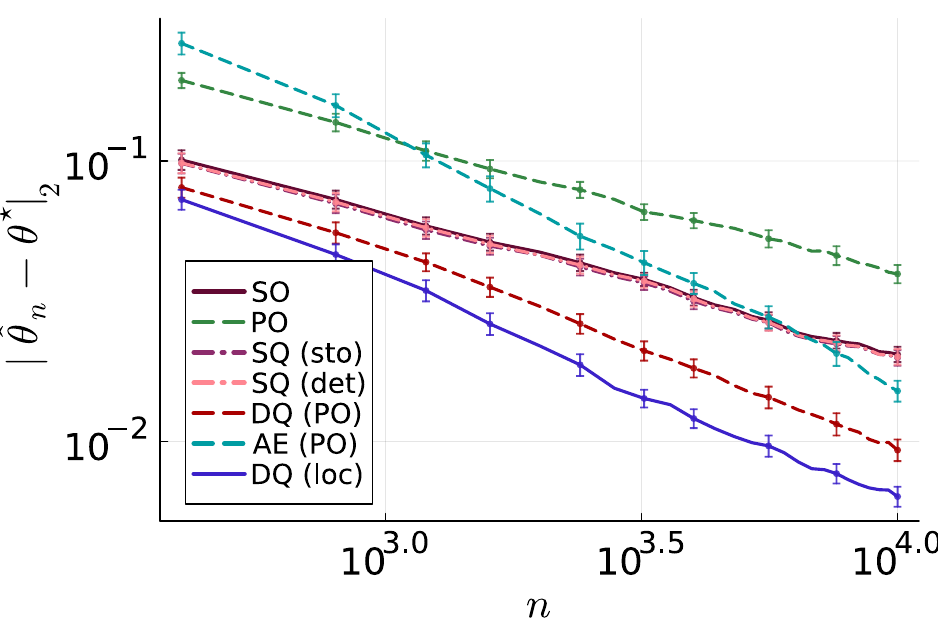}
    \includegraphics[width=0.48\linewidth]{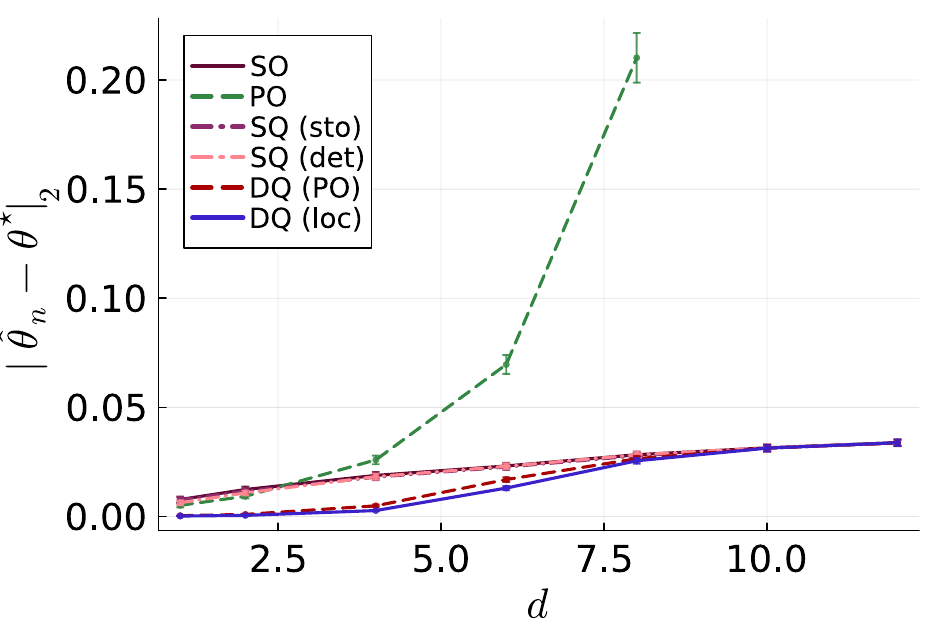} \\
    \includegraphics[width=0.48\linewidth]{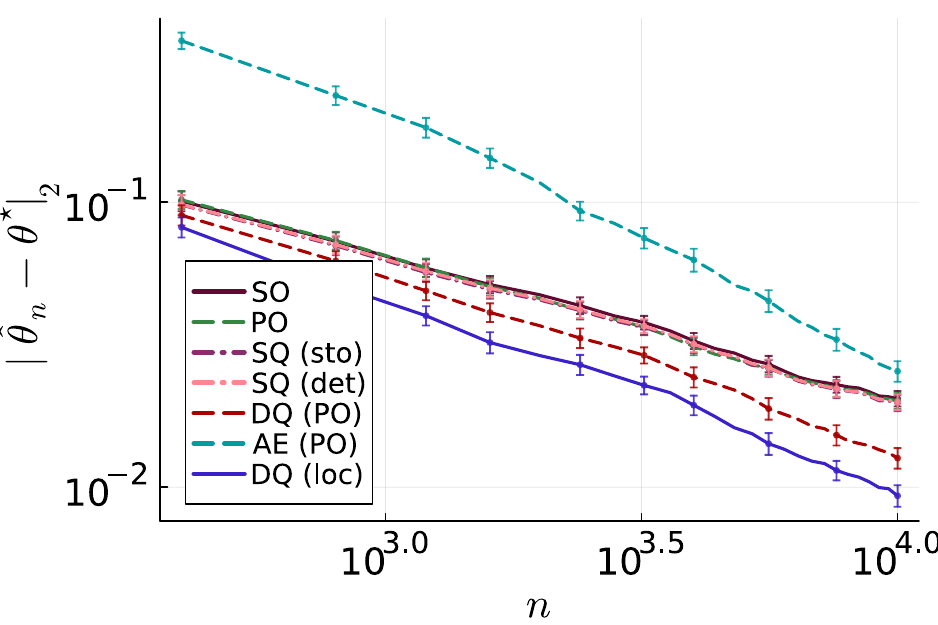}
    \includegraphics[width=0.48\linewidth]{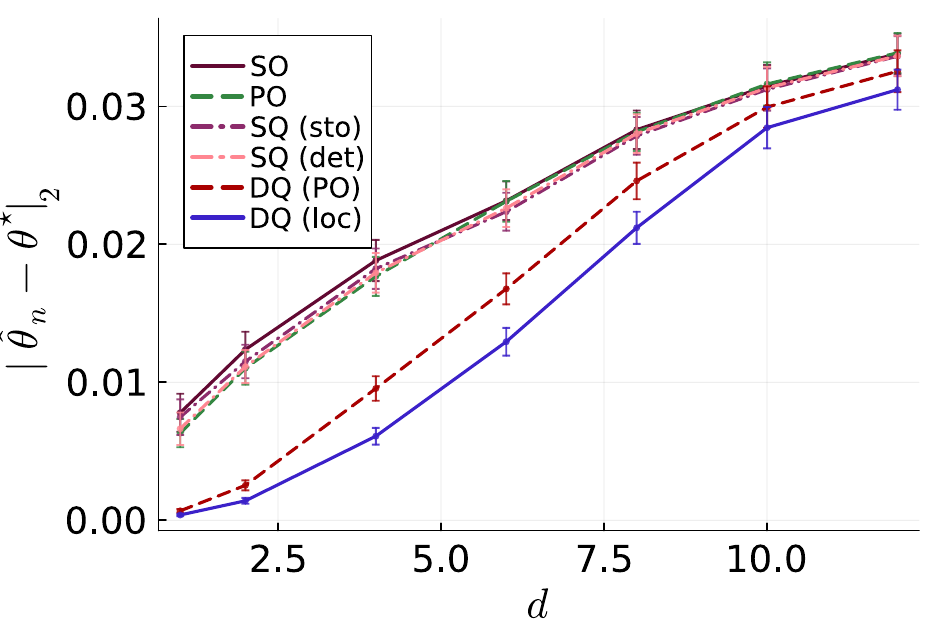}
    \caption{Estimation errors of $\cN(\theta^\star,I_d)$ with $\theta^\star \sim \cU([1,2]^{d})$ ($N_{\text{runs}} = 10^2$) for (top) $\eta(d) = \frac{1}{2}(1+d\log (2\pi))$ at (a) $d=5$ and (b) $n=10^4$ and (bottom) $\eta(d) = d\left(\frac{6}{5}+\frac{1}{2}\log (2\pi)\right)$ at (c) $d=5$ and (d) $n=10^4$. Error bars are $\pm 2$ standard errors.}
    \label{fig:GaussianLargeDErrorSupp}
\end{figure}
\paragraph{Other Thresholds for Multivariate Gaussian.}
To better understand the impact of class imbalance for larger $d$, we consider alternative threshold choices to $\eta(d)$ such that $c_{\eta(d)}^2 \eqdef F^{-1}_{\chi^2(d)}(1/2)$ used in Figure~\ref{fig:GaussianImpactDonError}.
A direct adaptation of the experiments in Figure~\ref{fig:Gaussian1DVariousError}(c) will be to consider fixed quantiles, i.e., $\eta(d,q) $ such that $c_{\eta(d,q)}^2 \eqdef F^{-1}_{\chi^2(d)}(q)$.
Instead, we choose to effect of the attrition of each class with increasing $d$.
For vanishing good samples, we consider $\eta(d) = \frac{1}{2}(1+d\log (2\pi))$ since $\bP_{\cN(\theta^\star,I_d)}(r_{\theta^\star} > 0)\to_{d \to +\infty} 0$.
For vanishing bad samples, we consider $\eta(d) = d\left(\frac{6}{5}+\frac{1}{2}\log (2\pi)\right)$ since $\bP_{\cN(\theta^\star,I_d)}(r_{\theta^\star} > 0)\to_{d \to +\infty} 1$.

Overall, Figure~\ref{fig:GaussianLargeDErrorSupp} confirms the findings of Figures~\ref{fig:GaussianImpactDonError} and~\ref{fig:Gaussian1DVariousError}(c).
For different choices of thresholds, fixed sample size or increasing dimension, $\dpe$ is the only estimator obtaining the best-of-both-worlds estimation error rate.
Moreover, our findings remain robust under class imbalance, except when the imbalance becomes too stringent, in which case performance degrades.
\begin{figure}[ht]
    \centering
    \includegraphics[width=0.58\linewidth]{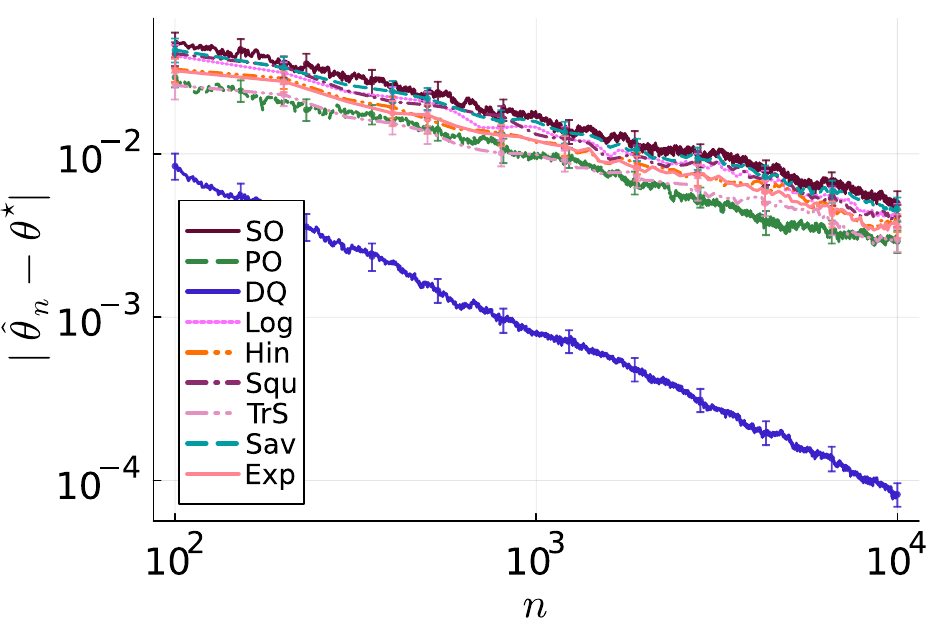}
    \includegraphics[width=0.38\linewidth]{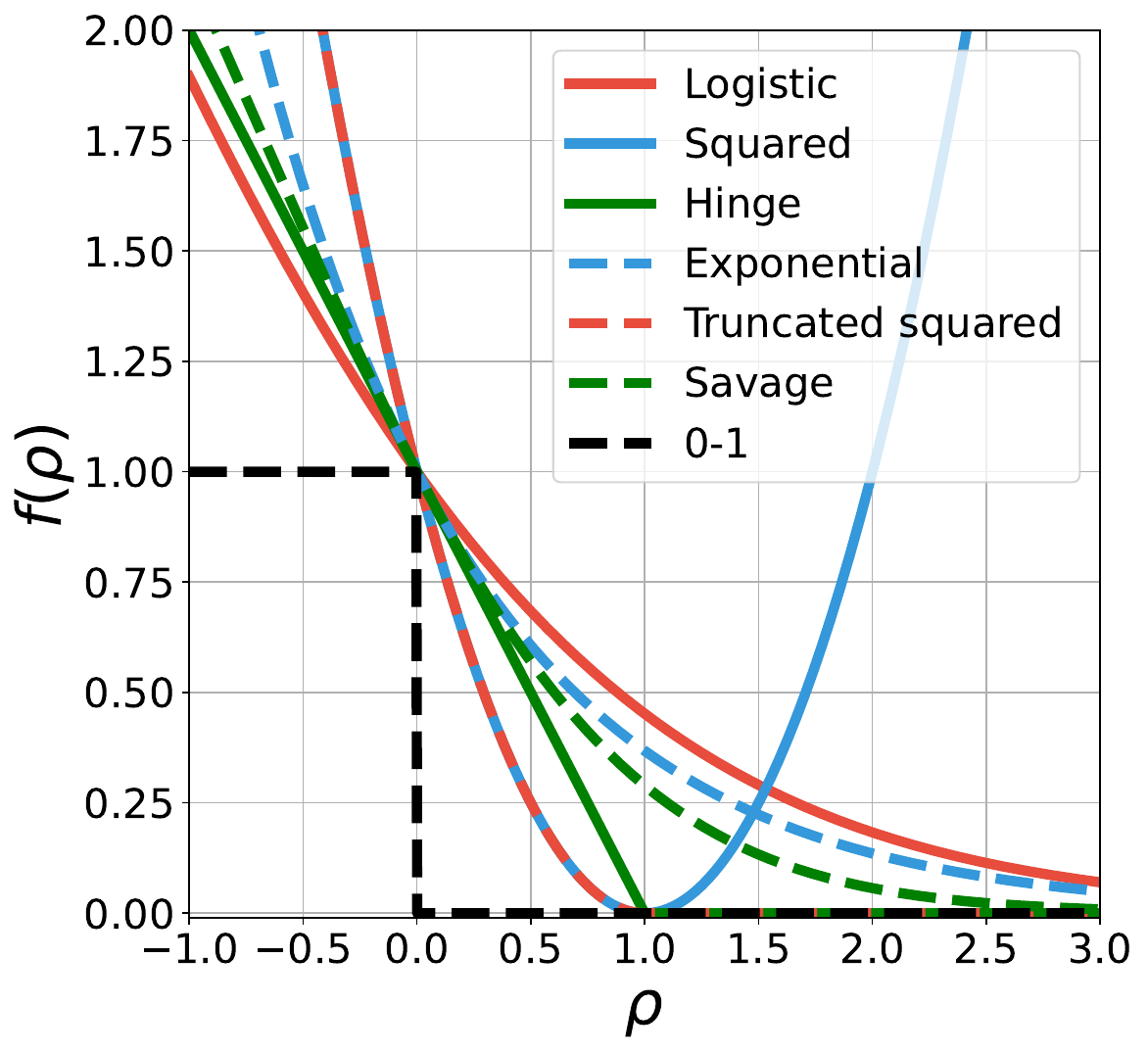}
    \caption{(a) Estimation errors of $\cN(\theta^\star,1)$ with $\theta^\star \sim \cU([1,2])$ ($N_{\text{runs}} = 200$) for $\eta(1/2) \approx 1.15$ so that $c_{\eta(1/2)}^2 \eqdef F^{-1}_{\chi^2(1)}(1/2)$.
    Error bars are $\pm 2$ standard errors.
    (b) Notable examples of binary classification loss functions~\citep[Figure 2]{tang2024generalized}. }
    \label{fig:OtherLossesDErrorSupp}
\end{figure}
\subsection{Estimators based on Convex Surrogate of the $0$--$1$ Loss}
\label{app:ssec_other_losses}
While \ref{eq:dqle} minimizes an objective based on the $0$-$1$ loss, \ref{eq:sqmle} minimizes an objective involving the logistic loss $f_{\textrm{Log}}(x) \eqdef \log(1+\exp(-x))$.
As in~\citet{tang2024generalized}, we can generalize this approach to other surrogate losses for the $0$-$1$ loss, including both convex losses and the non-convex Savage loss; see Figure~\ref{fig:OtherLossesDErrorSupp}(b).
Specifically, we consider the Hinge loss (Hin), i.e., $f_{\textrm{Hin}}(x) \eqdef \max\{0,1-x\}$, the square loss (Squ), i.e., $f_{\textrm{Squ}}(x) \eqdef (1-x)^2$, the truncated square loss (TrS), i.e., $f_{\textrm{TrS}}(x) \eqdef \max\{0,1-x\}^2$, the Savage loss (Sav), i.e., $f_{\textrm{Sav}}(x) \eqdef (1+\exp(x))^{-2}$, and the exponential loss (Exp), i.e., $f_{\textrm{Exp}}(x) \eqdef \exp(-x)$.

Given $(X_i,Z_i)_{i \in [n]} \sim q_{\theta^\star}^{\otimes [n]}$ and a loss $f$, we consider the $f$-estimator
\begin{equation} \label{eq:fsurrogateEstimator}
    \wh \theta^{f}_{n} \in \argmin_{\theta \in \Theta} \{ L^{\so}_{n}(\theta) +  \sum_{i \in [n]}  f (Z_i r_{\theta}(X_i)) \} \: .
\end{equation}
All those estimators are computed with the \texttt{Ipopt} solver.
Figure~\ref{fig:OtherLossesDErrorSupp}(a) shows that all estimators perform on par with \ref{eq:sqmle}, i.e., the one based on the logistic loss.
\begin{figure}[ht]
    \centering
    \includegraphics[width=0.48\linewidth]{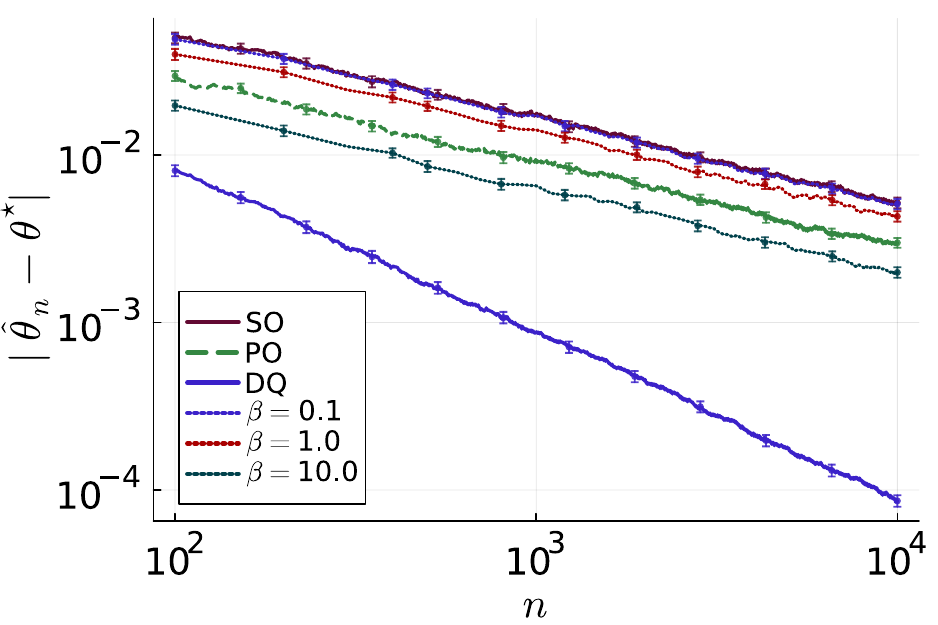}
    \includegraphics[width=0.48\linewidth]{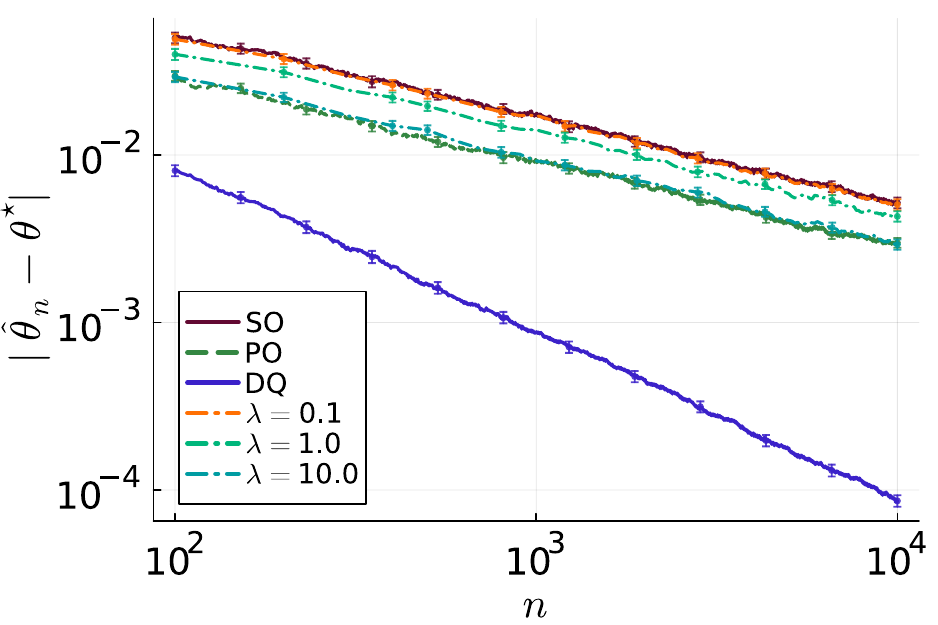}
    \caption{Estimation errors of $\cN(\theta^\star,1)$ with $\theta^\star \sim \cU([1,2])$ when (a) normalizing by $\beta$ for $\lambda=1$ and (b) regularizing by $\lambda$ for $\beta=1$ ($N_{\text{runs}} = 10^3$) for $\eta(1/2) \approx 1.15$ so that $c_{\eta(1/2)}^2 \eqdef F^{-1}_{\chi^2(1)}(1/2)$. Error bars are $\pm 2$ standard errors.}
    \label{fig:NormReguErrorSupp}
\end{figure}
\subsection{Impact of Normalization and Regularization}
\label{app:ssec_norm_regu}
The estimator defined in Eq.~\eqref{eq:fsurrogateEstimator} can be further generalized by introducing a regularization parameter $\lambda \ge 0$ and a normalization parameter $\beta > 0$.
Given $(X_i,Z_i)_{i \in [n]} \sim q_{\theta^\star}^{\otimes [n]}$, a loss $f$ and regularization/normalization $(\lambda,\beta)$, we consider the $(f,\lambda,\beta)$-estimator
\begin{align*}
    \wh \theta^{f,\lambda,\beta}_{n} \in \argmin_{\theta \in \Theta} \{ L^{\so}_{n}(\theta) + \lambda \sum_{i \in [n]}  f (\beta Z_i r_{\theta}(X_i)) \} \: .
\end{align*}
While similar modifications could be made for other losses, we focus on the logistic loss $f_{\textrm{Log}}(x) \eqdef \log (1+\exp(-x))$.
In particular, we recover \ref{eq:sqmle} by taking $\lambda = \beta = 1$.
Figures~\ref{fig:NormReguErrorSupp}(a) and (b) showcase the ``mild'' impact of normalization and regularization.
\begin{table}[ht]
\begin{center}
\begin{tabular}{l l l l l l l}
\hline
 & & & $\sqe$ &  & $\lle$ &  \\
Distribution & Known & $\eta$ & $\nabla_{\theta} M_{\sto}(\theta^\star)$ & $V_{\theta^\star,\sto}$ & $\nabla_{\theta} M_{\det}(\theta^\star)$ & $V_{\theta^\star,\det}$ \\
\hline
$\cN(0,1)$ & scale & $1.15$ &  $(1 \pm 14) \cdot 10^{-4}$  &  $0.85$  & $(1\pm 6)  \cdot 10^{-4}$  &  $0.46$  \\
 &  & $1.42$ &  $(7 \pm 13) \cdot 10^{-4}$  &  $0.84$  & $(-3\pm 7)  \cdot 10^{-4}$  &  $0.47$  \\
 &  & $2.12$ &  $(9 \pm 12) \cdot 10^{-4}$  &  $0.83$  & $(-5\pm 7)  \cdot 10^{-4}$  &  $0.60$  \\
Lap$(0,1)$ & scale & $1.39$  &  $(7 \pm 12) \cdot 10^{-4}$  &  $0.82$  & $(-5 \pm 8) \cdot 10^{-4}$  &  $0.52$  \\
 &  & $1.69$  &  $(2 \pm 12) \cdot 10^{-4}$  &  $0.82$  & $(7 \pm 7) \cdot 10^{-4}$  &  $0.62$  \\
 & center & $1.39$  &  $(-3 \pm 6) \cdot 10^{-4}$  & $0.92$   & $0.1316 \pm 3 \cdot 10^{-4} $ &    \\
 &  & $1.69$  &   $(-3 \pm 6) \cdot 10^{-4}$  & $0.92$  & $0.1095 \pm 2 \cdot 10^{-4} $ &    \\
Rayleigh$(1)$ & & $0.70$ &  $(-4 \pm 8)\cdot 10^{-4} $   &   $0.87$  & $0.1310 \pm 5 \cdot 10^{-4} $  &     \\
 & & $1$ &  $(5 \pm 8)\cdot 10^{-4} $   &  $0.86$   & $0.1578 \pm 5 \cdot 10^{-4} $  &     \\
Exp$(1)$ &  & $0.69$ &  $(0\pm 7) \cdot 10^{-4} $   &   $0.90$  & $0.1773 \pm 3 \cdot 10^{-4} $  &   \\
 &  & $2$ &  $(-1 \pm 7) \cdot 10^{-4} $   &  $0.89$   & $0.0830 \pm 4 \cdot 10^{-4} $  &   \\
Pareto$(2,1)$ & minimum  & $0.35$ &  $(1 \pm 3)  \cdot 10^{-4} $     &  $0.98$   & $0.0695 \pm 1 \cdot 10^{-4} $  &     \\
 &   & $2$ &  $(-2 \pm 3) \cdot 10^{-4} $     &  $0.98$   & $0.0308 \pm 2 \cdot 10^{-4} $  &     \\
\hline
\end{tabular}
\end{center}
\caption{Numerical estimation of the consistency of $\sqe$ and $\lle$, and asymptotic variance when consistent ($N_{\text{runs}} = 10^6$). The Student’s \(t\)-confidence intervals are defined at confidence level $95\%$.}
\label{tab:consistency_computation}
\end{table}
\subsection{Quality-based M-estimator}
\label{app:ssec_qualityMestimator}
\paragraph{Asymptotic Behavior of $\sqe$ and $\lle$.}
Table~\ref{tab:consistency_computation} provides numerical simulation to confirm that $\theta^\star$ maximize $M_{\sto}$, but not necessarily $M_{\det}$.
For $r_{\theta} = \log p_{\theta} + \eta$, only $\cF_{\cN, \Sigma}$ and $\cF_{\lap, b}$ ensure that $\nabla M_{\det}(\theta^\star) = 0$.
Therefore, $\sqe$ and $\lle$ are consistent estimators of $\theta^\star$, and we report their asymptotic variance as $V_{\theta^\star,\sto}$ and $V_{\theta^\star,\det}$.
For all the other classes of distributions, $\sqe$ is consistent but not $\lle$, hence its asymptotic variance is not reported.
\begin{figure}[ht]
    \centering
    \includegraphics[width=0.6\linewidth]{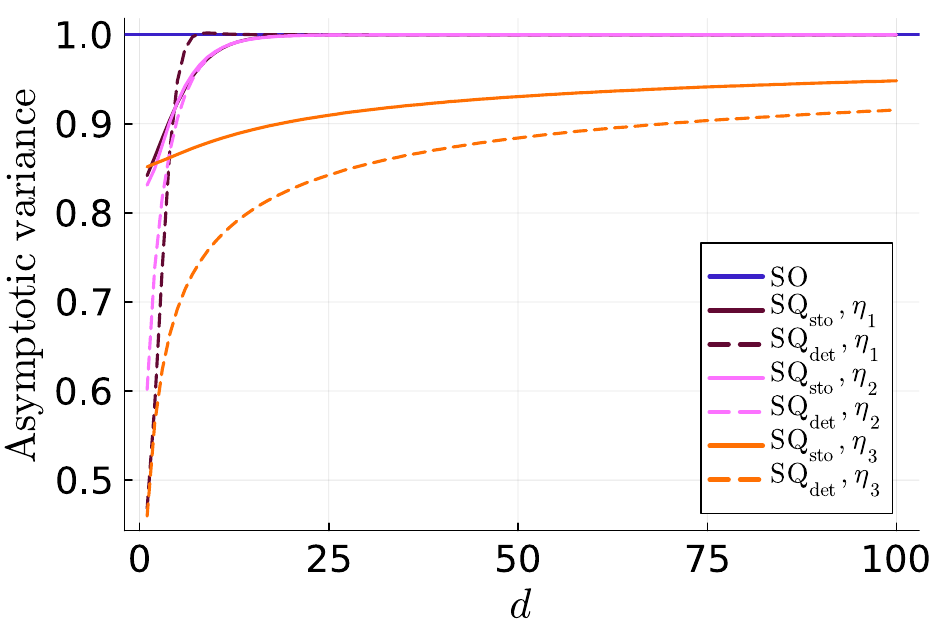}
    \caption{Numerical estimation of the asymptotic variance $V_{\theta^\star,\sto} = \alpha_d I_d$ for $\sqe$ and $V_{\theta^\star,\det} = \alpha_d I_d$ for $\lle$ with $\cN(0_d,I_d)$ ($N_{\text{runs}} = 10^2$). The thresholds are $\eta_1(d) = \frac{1}{2}(1+d\log (2\pi))$, $\eta_2(d) = d\left(\frac{6}{5}+\frac{1}{2}\log (2\pi)\right)$ and $\eta_3(d)$ such that $c_{\eta_3(d)}^2 = F^{-1}_{\chi^2(d)}(1/2)$. }
    \label{fig:GaussianMatrices}
\end{figure}
\paragraph{Asymptotic Variance of $\sqe$ and $\lle$ for Multivariate Gaussian.}
We study the impact of the threshold on the asymptotic covariance of $\sqe$ and $\lle$ for multivariate Gaussian as a function of $d$.
Figure~\ref{fig:GaussianMatrices} shows that they quickly converge towards the asymptotic variance of $\so$ (i.e., $I_{d}$) when the class imbalance becomes severe, namely for $\eta_1$ and $\eta_2$.
When the two classes are balanced (i.e., $\eta_3$), $\sqe$ and $\lle$ have consistently lower asymptotic variance than $\so$, and $\lle$ outperforms $\sqe$.

\end{document}